\documentclass[journal,twoside]{IEEEtran}
\usepackage[utf8]{inputenc}
\usepackage{amsmath,amssymb,graphicx}
\usepackage{nicefrac}
\usepackage[font=footnotesize]{caption}
\usepackage[font=footnotesize]{subcaption}
\usepackage{soul}
\usepackage{booktabs}
\usepackage{dblfloatfix} 
\usepackage{color}
\usepackage{setspace}
\usepackage[inline]{enumitem}
\usepackage{hyperref}
\usepackage{cleveref}
\usepackage{multirow}
\definecolor{darkgreen}{rgb}{0,0.6,0.2}

\usepackage[table]{xcolor}
\definecolor{lightgray}{rgb}{0.63, 0.79, 0.95}

\usepackage{algorithm}
\usepackage{algorithmic}
\usepackage{array}

\usepackage{alphalph}

\usepackage{url}

\title{Deep Image Translation with an Affinity-Based Change Prior for Unsupervised Multimodal Change Detection}
\author{Luigi~Tommaso~Luppino,~%
        Michael~Kampffmeyer,%~\IEEEmembership{Member,~IEEE,}
        Filippo~Maria~Bianchi,\\
        Gabriele~Moser,%~\IEEEmembership{Senior Member,~IEEE,}
        Sebastiano~Bruno~Serpico,%~\IEEEmembership{Fellow,~IEEE,}\\
        Robert~Jenssen,%~\IEEEmembership{Senior Member,~IEEE,}
        and~Stian~Normann~Anfinsen,%~\IEEEmembership{Member,~IEEE}% <-this % stops a space
\thanks{Manuscript received November 8, 2020; revised January 8, 2021; accepted January 23, 2021. The work of Luigi Tommaso Luppino was supported by the Research Council of Norway under Grant 251327. This work was supported in part by the Research Council of Norway and in part by NVIDIA Corporation by the donation of the GPU used for this research. \textit{(Corresponding author: Luigi Tommaso Luppino.)}}
\thanks{Luigi Tommaso Luppino, Michael Kampffmeyer, Robert Jenssen and Stian Normann Anfinsen are with the Machine Learning Group, Department of Physics and Technology, UiT The Arctic University of Norway, 9037 Troms\o{}, Norway (e-mail: \href{mailto:luigi.t.luppino@uit.no}{luigi.t.luppino@uit.no}).}% <-this % stops a space
\thanks{Filippo Maria Bianchi is with the Department of Mathematics and Statistics, UiT The Arctic University of Norway, 9037 Troms\o{}, Norway and
NORCE (the Norwegian Research Centre), 5008 Bergen, Norway.}%
\thanks{Gabriele Moser and Sebastiano Bruno Serpico are with DITEN Department, University of Genoa, 16145 Genoa, Italy.}% <-this % stops a space
}

\begin{document}

\maketitle

\begin{abstract}
Image translation with convolutional neural networks has recently been used as an approach to multimodal change detection. 
Existing approaches train the networks by exploiting supervised information of the change areas, which, however, is not always available.
A main challenge in the unsupervised problem setting is to avoid that change pixels affect the learning of the translation function. 
We propose two new network architectures trained with loss functions weighted by priors that reduce the impact of change pixels on the learning objective. 
The change prior is derived in an unsupervised fashion from relational pixel information captured by domain-specific affinity matrices. 
Specifically, we use the vertex degrees associated with an absolute affinity difference matrix and demonstrate their utility in combination with cycle consistency and adversarial training. The proposed neural networks are compared with state-of-the-art algorithms. Experiments conducted on three real datasets show the effectiveness of our methodology.
\end{abstract}


\iffalse
Change detection (CD) methods in remote sensing aim at identifying changes that happen on the Earth by comparing two or more satellite or aerial images acquired at different times.
Traditional CD methods rely on homogeneous data, however for many practical examples and applications homogeneity does not hold true, especially when different sensors are involved.
Heterogeneous data imply different domains, diverse statistical distributions and class signatures.
A common solution is to apply highly nonlinear transformations to transfer the data to a common domain where they can be compared.
Nonetheless, this crucial step often requires the iterative fine-tune of the transformation functions starting from preliminary results such as random initialisation, manual sample selection, post classification comparison or clustering.
In this work we propose two unsupervised frameworks for bi-temporal heterogeneous change detection based on data transformation and domain mapping.
The main features of the proposed approach are affinity matrices comparison, cyclic frameworks, and adversarial architectures.
The first takes place as the preliminary step which allows to highlight areas affected by changes, whose contributions to the training of the networks are penalised.
Experiments were conducted on two different real datasets, and the two proposed architectures were compared with two other recently published methods.
\fi

\section{Introduction}

\subsection{Background}

\IEEEPARstart{T}{he} goal of change detection (CD) methods based on earth observation data is to recognise changes on Earth by comparing two or more satellite or aerial images covering the same area at different times~\cite{singh1989review}.
Multitemporal applications include the monitoring of long term trends, such as deforestation, urban planning, and earth resources surveys, whereas bi-temporal applications mainly regard the assessment of natural disasters, for example earthquakes, oil spills, floods, and forest fires~\cite{luppino2017clustering}.
This paper will focus on the latter case, and more specifically on the scenario where the changes must be detected from two satellite images with high to medium spatial resolution (10 to 30 meters).
These resolutions allow to detect changes in ground coverage (forest, grass, bare soil, water etc.) below hectare scale, but are not suitable to deal with changes affecting small objects on meter scale (buildings, trees, cars etc.).
At these resolutions it is common to assume that co-registration can be achieved by applying simple image transformations such as translation, rotation, and re-sampling~\cite{zhan2018iterative,liu2016deep,liu2018change,zhu2017deep}.
This means that each pixel in the first image and its corresponding one in the second image represent the same point on the Earth.
Consequently, even a simple pixel-wise operation (e.g.\ a difference or a ratio) would highlight changes when working with homogeneous data~\cite{liu2016deep,niu2018conditional,khan2017forest},
i.e.\ data collected by the same sensor, under the same geometries and seasonal or weather conditions, and using the same configurations and settings.
More robust and efficient approaches consider complex algorithms rather than simple mathematical operations to detect changes, and many examples of homogeneous CD methods can be found in the literature~\cite{khan2017forest,gong2016change,gong2017feature,lyu2016learning,mou2018learning}.

\subsection{Motivation}

To rely on only one data acquisition modality represents a limitation, both in terms of response time to sudden events and in terms of temporal resolution when monitoring long-term trends.
To exemplify, heterogeneous change detection algorithms facilitate rapid change analyses by being able to utilise the first available image, regardless of modality~\cite{dalla2015challenges,ghamisi2019multisource}.
They also allow to increase the number of samples in a time series of acquisitions by inserting images from multiple sensors.
On one hand, this allows to exploit the images acquired by all the available sensors, but on the other hand raises additional challenges.
Heterogeneous sensors usually measure different physical quantities, meaning that one terrain type might be represented by dissimilar statistical models from sensor to sensor, while surface signatures and their internal relations may change completely across different instruments~\cite{liu2016deep,niu2018conditional,gong2019coupling}.
For example, optical and synthetic aperture radar (SAR) payloads are dominantly used for CD in remote sensing~\cite{zhao2017discriminative,zhan2018log} and they are often seen as complementary: the use of optical instruments is affected by solar illumination and limited to low cloud coverage, whilst SAR can operate at any time and under almost any weather conditions, because clouds are transparent to electromagnetic waves at SAR frequencies.
On the other hand, optical data take real values affected by a modest additive Gaussian noise (mainly due to atmospheric disturbance, thermal and shot noise inside the sensor), whose effect can be easily accounted for~\cite{landgrebe2005signal}, whereas SAR feature vectors take complex values representing the coherent sum of the backscattered echoes, which can present high fluctuations from one pixel to the next both in amplitude and phase, resulting in the so-called speckle, a multiplicative effect which is more challenging to mitigate~\cite{moreira2013tutorial}.
In few words, it is not guaranteed that the data acquired by heterogeneous sources lie in a common domain, and a direct comparison is meaningless without processing and co-calibrating the data first~\cite{luppino2017clustering}.

Heterogeneous CD methods are meant to cope with these issues, and as discussed in~\cite{gong2016coupled,luppino2019unsupervised}, there is not a unique way to categorize them.
However, two general criteria to group them are the following:
\begin{enumerate*}
  \item unsupervised methods or supervised methods;
  \item deep learning methods or traditional signal processing methods.
\end{enumerate*}
The analysis in this paper will exclusively cover unsupervised frameworks. Since they do not require any supervised information about the change, they are usually more appealing than the supervised counterparts.
Indeed, collecting labelled data is often costly and nontrivial, both in terms of the time and competence required~\cite{zhan2018iterative,zhan2018log}.
Concerning the second distinction, deep learning has become the state-of-the-art in many image analysis tasks, including in the field of remote sensing~\cite{liu2016deep,zhu2017deep}.
Deep learning methods can achieve high performance thanks to the flexibility of neural networks, which are able to apply highly nonlinear transformations to any kind of input data.
For these reasons, the analysis of the literature will mainly focus on deep learning, although many important methods, based on minimum energy~\cite{touati2018energy}, nonlinear regression~\cite{luppino2019unsupervised}, dictionary learning~\cite{gong2016coupled},~\cite{ferraris2019coupled}, manifold learning~\cite{prendes2015new}, fractal projections~\cite{mignotte2020fractal}, or copula theory~\cite{mercier2008conditional} are worth mentioning.
We refer the interested readers to \cite{luppino2019unsupervised} for a state-of-the-art analysis on heterogeneous CD based on more classical methods.

We point out that heterogeneous CD can be framed within the general context of multimodal data fusion, which broadly encompasses all processing, learning, and analysis methodologies aimed at jointly exploiting different data modalities. In remote sensing, these modalities most typically correspond to different sensors, missions, spatial resolutions, or acquisition properties (e.g., incidence angle, radar polarization, and spectral channels)~\cite{gomezchova2015multimodal}. Note that heterogeneous CD methods are effective also to deal with the simpler case in which the heterogeneity between the images is merely due to different environmental conditions at the moment of the acquisitions (weather, time of the day, season, and so forth). We refer the reader to the review paper in~\cite{gomezchova2015multimodal} for a general taxonomy of multimodal fusion in remote sensing, with examples of multiresolution, multiangular, multisensor, multitemporal, and spatial-spectral fusion using a variety of methodological approaches, including deep learning and also discussing a CD case study. In the case of image classification, recent examples of multimodal approaches based on deep neural networks include the multimodal deep learning framework in~\cite{hong2020morediverse}, the multisensor and multiscale method in~\cite{audebert2018beyond} for semantic labeling in urban areas, and the technique in~\cite{benedetti2018m3} for land cover mapping from multimodal satellite image time series. The role of shallow and deep learning approaches in the area of feature extraction -- with focus on hyperspectral imagery and involving various data fusion concepts -- has recently been reviewed in~\cite{rasti2020feature}. The scientific outcome of a recent international contest in the area of multimodal fusion with open satellite and ancillary/geospatial data has been presented in~\cite{yokoya2018opendata}.

\subsection{Proposed method}

We propose to combine traditional machine learning and pattern recognition techniques with deep image translation architectures to perform unsupervised CD based on heterogeneous remote sensing data.
More specifically, a comparison of domain-specific affinity matrices allows us to retrieve in a self-supervised manner the \emph{a priori} change indicator, referred to as the prior, driving the training process of our deep learning methods.
In particular, our aim is to provide a reliable and informative prior, representative of the whole feature space, which is an alternative with respect to other priors previously used for heterogeneous CD, such as randomly initialised change maps, clustering/post-classification-comparison outputs, or supervised sample selection.
The proposed prior computation method is an efficient approach that provides more useful information than randomly-initialised change maps, which are associated with convergence problems and inconsistent overall performance. 
It is directly and automatically obtainable from the input data without need of any tuning and, as opposed to clustering methods, it does not require to select sensible hyperparameters such as the number of clusters, which strictly depends on the area under investigation and the number of land covers present in the scene.
The advantage with respect to post-classification and supervised sample selection is that the latter make use of prior information which can be difficult to obtain, or user prompt or in-situ measurements, which are time-consuming and/or expensive. Instead, none of the aforementioned information are required by the proposed approach.

Two architectures are proposed: The X-Net is composed of two fully convolutional networks, each dedicated to mapping the data from one domain to the other; The ACE-Net consists of two autoencoders whose code spaces are aligned by adversarial training.
Their performance and consistency are tested against two recent state-of-the-art methods on three benchmark datasets, illustrating how the proposed networks perform favourably
as compared to them.
Summing up, the main contributions of this work are:
\begin{itemize}
    \item A novel procedure to obtain a priori information on structural changes between the images based on a comparison of intramodal information on pixel relations.
    \item Two neural network architectures designed to perform unsupervised change detection, which explicitly incorporate this prior.
\end{itemize}
Moreover, this work represents a valuable contribution to the field of study as the proposed framework for heterogeneous change detection is made publicly available at this link: \url{https://github.com/llu025/Heterogeneous_CD}, together with the re-implementation of the two reference methods, as well as the three datasets used in this paper.

The remainder of this article is structured as follows: \Cref{sec:related_work} describes the theoretical background and the related work.
\Cref{sec:method} introduces the reader to the notation, the proposed procedure and the architectures.
Results on three datasets are presented in \Cref{sec:results}.
\Cref{sec:limits} includes a discussion of the main features and drawbacks of each method used in this work.
\Cref{sec:concl} concludes the paper and summarises the proposed method and obtained results.

\section{Related work}\label{sec:related_work}
The most common solution to compare heterogeneous data is to transform them and make them compatible.
This is the main reason why many of the heterogeneous CD methods are related to the topics of domain adaptation and feature learning.
In the following we list the main deep learning architectures that are found in the heterogeneous CD literature, along with some examples of methods implementing them.

\subsection{Stacked Denoising Autoencoders}
\subsubsection{Background}
The autoencoder (AE) is a powerful deep learning architecture which has proven capable of solving problems like feature extraction, dimensionality reduction, and clustering~\cite{hinton2006reducing}.
A denoising AE (DAE) is a particular type of AE trained to reconstruct an input signal that has been artificially corrupted by noise.
The stacked denoising autoencoder (SDAE) is probably the most used model to infer spatial information from data and learn new representations and features.
SDAEs are trained following the same procedure as DAEs, but their ability of denoising is learned in a layerwise manner by injecting noise into one layer at the time, starting from the outermost layer and moving on towards the innermost one~\cite{vincent2010stacked}.
In the following, some examples from the heterogeneous change detection literature are presented.

\subsubsection{Applications}\label{sec:sccn}
Su \textit{et al.}~\cite{su2017deep} used change vector analysis to distinguish between three classes: unchanged areas, positive changes and negative changes, as defined in~\cite{bovolo2007theoretical}.
They exploit two SDAEs to extract relevant features and transfer the data into a code space, where code differences from co-located patches are clustered to achieve a preliminary distinction between samples from the three classes.
These samples are then used to train three distinct mapping networks, each of which learns to take the features extracted from one image as input and transform them into plausible code features related to another image.
The goal of the first network is to reproduce the expected code from the latter image in case of a positive change, the second aims to do the same in case of a negative change, and the last takes care of the \textit{no-change} case.
A pixel is eventually assigned to the class corresponding to the reproduced code showing the smallest difference with the original code from the second image.

In a very similar fashion, Zhang \textit{et al.}~\cite{zhang2016cd} first use a spatial details recovery network trained on a manually selected set to coregister the two images, but then extract relevant features from them with two SDAEs trained in an unsupervised fashion.
Starting from these transformed images, manual inspection, post-classification comparison or clustering provides a coarse change map.
This is used to select examples of unchanged pairs of pixels, which are used to train a mapping network.
Once the data are mapped into a common domain, feature similarity analysis highlights change pixels, which are isolated from the rest by segmentation;

In a paper by Zhan \textit{et al.}~\cite{zhan2018log}, SAR data are log-transformed and stacked together with the corresponding optical data.
Next, a SDAE is used to extract two relevant feature maps from the stack, one for each of the input modalities.
These are then clustered separately and the results are compared to obtain a difference image.
The latter is segmented into three clusters: pixels certain to belong to changed areas, pixels certain to belong to unchanged areas, and uncertain pixels.
Finally, the pixels labelled with certainty are used to train a classification network, which is then able to discriminate the uncertain pixels into the \textit{change} and \textit{no-change} clusters, providing the final binary change map.

Zhan \textit{et al.\ }~\cite{zhan2018iterative} proposed to learn new representative features for the two images by the use of two distinct SDAEs.
A mapping network is then trained to transform these extracted features into a common domain, where the pixels are forced to be similar (dissimilar) according to their probability of belonging to the unchanged (changed) areas.
The probability map is initialised randomly and the training alternates between two phases: updating the parameters of the mapping network according to the probabilities, and updating the map according to the output of the network.
Once the training reaches its stopping criterion, the difference between the two feature maps is obtained.
Instead of producing a binary change map, this method introduces a hierarchical clustering strategy that highlights different types of change as separate clusters.

The symmetric convolutional coupling network (SCCN) was proposed by Liu \textit{et al.}~\cite{liu2016deep}: %and it represents the first \textit{state-of-the-art} method against which our approach is compared.
After two SDAEs are pretrained separately on each image, their decoders are removed, one of the encoders is frozen, and the other is fine-tuned by forcing the codes of the pixels most likely to not represent changes to be similar.
The pixel probability of \textit{no-change} is initialised randomly, and is updated iteratively and alternately together with the parameters of the encoders.
A stable output of the objective function is eventually reached and the probability map is finally segmented into the usual binary change map.
This method was later improved in~\cite{zhao2017discriminative} by modifying slightly the objective function and the probability map update procedure.

\subsection{Generative Adversarial Networks}
\subsubsection{Background}
Among the most important methods in the literature of domain adaptation and data transformation are the generative adversarial networks (GANs).
Proposed by Goodfellow \textit{et al.\ }in~\cite{goodfellow2014generative}, these architectures consist of two main components competing against each other.
Drawing samples from a random distribution, a generator aims at reproducing samples from a specific target distribution as output.
On the other hand, a discriminator has the goal to distinguish between \textit{real} data drawn from the target distribution and \textit{fake} data produced by the generator.
Through an adversarial training phase, the generator becomes better at producing fake samples and it is rewarded when it fools the discriminator, whereas the latter improves its discerning skills and is rewarded when it is able to detect fake data.
Both the two parts try to overcome their opponent and become better, benefiting from this competition.

A drawback of this method is the difficulty in balancing the strength of the two components.
Their efforts have to be equal, otherwise one will start to dominate the other, hindering the simultaneous improvement of both.
Conditional GANs~\cite{Isola_2017_CVPR} are a particular case, where fake data is generated from a distribution conditioned on the input data.
This architecture is suitable for the task of \textit{image-to-image translation}: images from one domain are mapped into another (e.g.\ drawings or paintings into real pictures, winter landscapes into summer ones, maps of cities into aerial images).

\subsubsection{Applications}\label{sec:cgan}
The potential of this method to transform data acquired from one satellite sensor into another is striking, and it was first explored in~\cite{merkle2018exploring} to match optical and SAR images.
The dataset used consists of pairs of co-located optical and SAR images acquired at the same time.
The generator learns during training to produce a plausible SAR image starting from the optical one, without knowing what the corresponding real SAR data look like.
The same optical image and one of the two SAR images, either the generated or the original, are provided to the discriminator, which has to infer whether the images are a \textit{real} or \textit{fake} pair.
For testing, the generator takes the optical images as input and provides the synthetic SAR data, whereas the original SAR data become the ground truth.

In~\cite{niu2018conditional}, the same concept is applied to perform heterogeneous CD.
%Our method is compared also against it as benchmark, and thus it deserves a detailed explanation.
The scheme is always the same: a generator tries to reproduce SAR patches starting from the corresponding optical ones, and a discriminator aims at detecting these \textit{fake} patches.
%However, although the authors claim that the generator is able to translate the optical patches into similar SAR ones, they also admit that the pixel difference between the original SAR data with the generated ones may still be large.
In order to facilitate a direct comparison, they \iffalse therefore \fi introduce an approximation network which learns to transform the original SAR patches into the generated ones.
Note that the training of all these networks must be carried out on patches not containing change pixels, and any other patch must be flagged and excluded from this process.
At first, all the flags are set to \textit{no-change}. Then these steps are iterated: the conditional GAN is updated, the approximation network is tuned accordingly, and finally the generated and approximated patches are compared to flag the ones containing changes.
Once the training phase is over, the generated image and the approximated image are pixel-wise subtracted and segmented binarily.

\subsection{Cyclic Generative Adversarial Networks}
\subsubsection{Background}
A more complex framework than the conditional GAN is the cycle GAN~\cite{zhu2017unpaired}.
The idea is simple: instead of using just one generator-discriminator couple dealing with the transformation from domain $\mathcal{X}$ to domain $\mathcal{Y}$, another tandem generator-discriminator is added to do the vice versa.
This means that the framework can be tested for so-called \textit{cycle consistency}: It should be possible to perform a composite translation of data from domain $\mathcal{X}$ to domain $\mathcal{Y}$, and then onwards to domain $\mathcal{X}$ (denoted $\mathcal{X}\!\to\!\mathcal{Y}\!\to\!\mathcal{X}$), and the full translation cycle should reproduce the original input.
Equivalently, the cycle $\mathcal{Y}\!\to\!\mathcal{X}\!\to\!\mathcal{Y}$ should reproduce the original input in domain $\mathcal{Y}$.

In~\cite{murez2018image}, this framework is applied and extended further: Along with the two input domains $\mathcal{X}$ and $\mathcal{Y}$, a latent space $\mathcal{Z}$ is introduced in between them. Data from the original domains are transformed to $\mathcal{Z}$, where they should ideally not be discernible.
Thus, four generators are used to map data across domains: from $\mathcal{X}$ to $\mathcal{Z}$, from $\mathcal{Z}$ to $\mathcal{Y}$, from $\mathcal{Y}$ to $\mathcal{Z}$, and from $\mathcal{Z}$ to $\mathcal{X}$.
The accurate reconstruction of the images is the first enforced principle: Data mapped from domain $\mathcal{X}$ ($\mathcal{Y}$) to $\mathcal{Z}$ must be mapped back correctly to $\mathcal{X}$ ($\mathcal{Y}$).
The next requirement is cycle-consistency: Starting from $\mathcal{X}$ ($\mathcal{Y}$) and going first to $\mathcal{Z}$ and then to $\mathcal{Y}$ ($\mathcal{X}$), the images must go back to $\mathcal{X}$ ($\mathcal{Y}$) passing through $\mathcal{Z}$ again and match exactly with the original input.
Concerning the discriminators, there are three: one should distinguish whether data mapped into $\mathcal{Z}$ come originally from $\mathcal{X}$ or $\mathcal{Y}$; another discriminates between original images from $\mathcal{X}$ and images which started from $\mathcal{Y}$ and performed half a cycle; the third does the same in domain $\mathcal{Y}$.

\subsubsection{Applications}
Inspired by these concepts, Gong \textit{et al.\ } proposed the coupling translation networks to perform heterogeneous CD~\cite{gong2019coupling}.
However, their architecture is simpler.
Two variational AEs are combined so that their encoders separately take as input optical and SAR patches, respectively, and the two codes produced are stacked together.
The stacked code is then decoded by both decoders and each of those yields two output patches: one is the reconstruction of the input patch from the same domain, the other is the transformation of the input patch from the opposite domain.
The latter must be detected by a discriminator which is taught to discern reconstructed data from \textit{fake} transformed data.
This framework has only two discriminators, one after each decoder, whereas the code spaces of the two AEs are aligned throughout the training, eventually becoming the common latent domain, namely $\mathcal{Z}$.
Together with the adversarial loss, the reconstruction and the cycle-consistency drive the learning process, which enables the two networks to translate data across domains, such that a direct comparison is feasible.

In the following section we explain how our methodology fits in this picture, framed in-between cycle-consistency and adversarial training.

\section{Methodology}\label{sec:method}

The same geographical region is scanned by two sensors whose pixel measurements lie in domains $\mathcal{X}$ and $\mathcal{Y}$, respectively. The first sensor captures an image $\mathcal{I_X} \in \mathcal{X}^{H \times W}$ at time $t_1$, and the other sensor an image $\mathcal{I_Y}\in\mathcal{Y}^{H \times W}$ at time $t_2$. $H$ and $W$ denote the common height and width of the images, that are obtained through coregistration and resampling. The feature spaces $\mathcal{X}$ and $\mathcal{Y}$ have dimensions $|\mathcal{X}|$ and $|\mathcal{Y}|$.

We further assume that a limited part of the image has changed between time $t_1$ and $t_2$.
The final goal of the presented method is to transform data consistently from one domain to the other.
To do so, it is crucial to learn a one-to-one mapping between the land cover signatures of one domain and the corresponding signatures in the other.
Since no prior information is available, a reasonable option is to learn a mapping from every pixel in $\mathcal{I_X}$ to the corresponding pixel in $\mathcal{I_Y}$ and vice versa. 

%Assume that we have two functions, $F(\boldsymbol{X}):\mathcal{X}^{h\times w}\to\mathcal{Y}^{h\times w}$ and $G(\boldsymbol{Y}):\mathcal{Y}^{h\times w}\to\mathcal{X}^{h\times w}$, that can transform images patches $\boldsymbol{X}\in\mathcal{X}^{h\times w}$ and $\boldsymbol{X}\in\mathcal{X}^{h\times w}$. 

A possibility would be to train two regression functions
\begin{align*}
\hat{\boldsymbol{Y}} &= F(\boldsymbol{X}):\mathcal{X}^{h\times w}\to\mathcal{Y}^{h\times w} \\ \hat{\boldsymbol{X}} &= G(\boldsymbol{Y}):\mathcal{Y}^{h\times w}\to\mathcal{X}^{h\times w}
\end{align*}
to map image patches $\boldsymbol{X}\!\in\!\mathcal{X}^{h\times w}\!\subseteq\!\mathcal{I_X}$ and $\boldsymbol{Y}\!\in\!\mathcal{Y}^{h\times w}\!\subseteq\!\mathcal{I_Y}$ between the image domains by using the entire images $\mathcal{I_X}$ and $\mathcal{I_Y}$ as training data.
However, the presence of areas affected by changes would distort the learning process, because they would promote a transformation from one land cover in one domain to a different land cover in the other domain.
For example, forests and fire scars may be erroneously connected, as may land and flooded land.
To reduce the impact of these areas on training, we first perform a preliminary analysis to highlight changes.
Then, the contribution of each pixel to the learning process is inversely weighted with a score expressing the chance of it being affected by a change.
In this section, we first describe the algorithm providing the preliminary change analysis. We then propose two deep learning architectures and, finally, explain how they can exploit the prior computed in the change analysis.

\subsection{Prior computation}\label{subsec:affinity}
To compute a measure of similarity between multimodal samples based on affinity matrices, we adopt an improved version of the original method proposed in our previous work~\cite{luppino2019unsupervised}. Please notice that the following procedure is totally unsupervised and does not require any ancillary information or knowledge about the data nor about the acquiring sensors.

A $k \times k$ sliding window covers an area $p$ of both $\mathcal{I_X}$ and $\mathcal{I_Y}$, from which a pair of corresponding patches $\boldsymbol{X}$ and $\boldsymbol{Y}$ are extracted.
$\boldsymbol{X}_i$ ($\boldsymbol{Y}_i$) and $\boldsymbol{X}_j$ ($\boldsymbol{Y}_j$) stand for feature vector $i$ and $j$ of patch $\boldsymbol{X}$ ($\boldsymbol{Y}$), with $i,j\in\{1,\dots,k^2\}$.
The distance between a pixel pair $(i,j)$ is defined as $d^{m}_{i,j}$, where the modality $m\in\{\mathcal{X},\mathcal{Y}\}$ depends on whether the samples are taken from $\boldsymbol{X}$ or $\boldsymbol{Y}$.
The appropriate choice of distance measure depends on the domain and the underlying data distribution.
The hypothesis of Gaussianity for imagery acquired by optical sensors is commonly assumed~\cite{bovolo2007theoretical,bovolo2015time}.
Concerning SAR intensity data, a logarithmic transformation is sufficient to bring it to near-Gaussianity~\cite{luppino2017clustering,zhan2018log}.
We use the computationally efficient Euclidean distance, as it is suitable for (nearly) Gaussian data.

Once computed, the distances between all pixel pairs can be converted to affinities, intended as values describing how close two points are in some feature space according to a metric~\cite{koutroumbas2008pattern}, for instance by the Gaussian kernel:
\begin{equation}
A^{m}_{i,j} =  \exp\left\{-\frac{\left(d^{m}_{i,j}\right)^2}{h_m^2}\right\}\, \in \left(0,1\right], \quad i,j\in\{1,\dots,k^2\}\,.
\label{eq:affm}
\end{equation}
$A^{m}_{i,j}$ are the entries of the affinity matrix $A^{m} \in \mathbb{R}^{k^2 \times k^2}$ for the given patch and modality $m$.
Here, the term affinity is used as synonym for similarity, as it has been widely used in the machine learning literature, especially with methods based on graph theory, such as spectral clustering and Laplacian eigenmaps~\cite{von2007tutorial,shi2000normalized,liu2017learning,cheng2018depth,maire2016affinity,chung1997spectral}.
The kernel width $h_m$ is domain-specific and can be determined automatically.
Our choice is to set it equal to the average distance to the $K^{th}$ nearest neighbour for all data points in the relevant patch ($\boldsymbol{X}$ or $\boldsymbol{Y}$), with $K=\frac{3}{4}k^2$.
In this way, a characteristic distance within the patch is captured by this heuristic, which is robust with respect to outliers
%. 
%In particular, the close neighbourhood of pixel $i$ presents values of $A^{m}_{i,j}$ within a reasonable interval, whereas the other affinities gradually tend to $0$~
\cite{myhre2012mixture}.
Silverman's rule of thumb~\cite{wand1995kernel} and other common approaches to determine the kernel width have not proven themselves effective in our experimental evaluation, so they were discarded.
Once the two affinity matrices are computed, a matrix $D$ holding the element-wise absolute differences $D_{i,j} = \lvert A_{i,j}^{\mathcal{X}} - A_{i,j}^{\mathcal{Y}} \rvert$ can be obtained.

Our previous algorithm \cite{luppino2019unsupervised} would at this point evaluate the Frobenius norm of $D$ and assign its value to all the pixels belonging to $p$.
Then, the $k \times k$ window is shifted one pixel and the procedure is iterated for the set $\mathcal{P}$ of all overlapping patches $p$ that can be extracted from the image.
The final result for each pixel is derived by averaging the set $\mathcal{S}^F$ of Frobenius norms obtained with all the patches covering that pixel.
Clearly, the loop over the patches in $\mathcal{P}$ is computationally heavy, although when shifting a patch one pixel, most of the already computed pixel distances can be reused. 
If $N= H\cdot W$ is the total number of pixels in the images, the cardinality of $\mathcal{P}$ is
\begin{equation}
\begin{split}
    \lvert\mathcal{P}\rvert & = (H-k+1)\cdot(W-k+1) \\
    & = N -(H + W)(k-1) + (k-1)^2\,.
\end{split}
\end{equation}
Shifting the sliding window by a factor larger than one will speed up the algorithm, but with the result that the final map of averaged Frobenius norms exhibits an unnatural tile pattern.

To address this issue, we propose to compute the following mean over the rows of $D$ (or columns, since $A^{\mathcal{X}}$ and $A^{\mathcal{Y}}$ are symmetrical, hence so is $D$):
\begin{equation}
    \alpha_i = \frac{1}{k^2} \sum_{j=1}^{k^2} \lvert A^{\mathcal{X}}_{i,j} - A^{\mathcal{Y}}_{i,j} \rvert, \quad i\in\{1,\dots,k^2\}
\label{eq:alpha}
\end{equation}
The main rationale for this operation is that pixels affected by changes are the ones perturbing the structural information captured by the affinity matrices, and so, on average, their corresponding rows in $D$ should present larger values.

We can also choose to look at $D$ as the affinity matrix of a change graph, with change affinities $D_{i,j}$ that indicate whether the relation between pixel $i$ and $j$ has changed. The row sums of $D$ become vertex degrees of the graph that sum the change affinities of individual pixels. A high vertex degree suggests that many pixel relations have changed, and that the pixel itself is subject to a change. The scaling of the vertex degree by $1/k^2$ normalises and fixes the range of $\alpha_i$ to $[0,1]$, which simplifies both thresholding and probabilistic interpretation. Another advantage of the vertex degree is that it isolates evidence about change for a single pixel, whereas the Frobenius norm of $D$ accumulates indications of change for an entire patch and provides change evidence that is less localised. In conclusion, $\alpha_i$ contains more reliable information and, most importantly, relates only to a single pixel $i$. It is therefore possible to introduce a shift factor $\Delta > 1$, which on one hand means that the final result becomes an average over a smaller set $\mathcal{S}^\alpha$, but on the other hand speeds up the computations considerably.
Potentially, this shift can be as large as the patch size, reducing the amount of patches by a factor of $k^2$. However, this is not desirable, since each pixel will be covered only once, leaving us with a set $S^\alpha$ of one element and no room for averaging.

\begin{figure*}[ht!]
\hfill
\begin{minipage}{.48\columnwidth}
  \begin{subfigure}{\linewidth}
    \centering
    \includegraphics[width=0.8\linewidth,keepaspectratio]{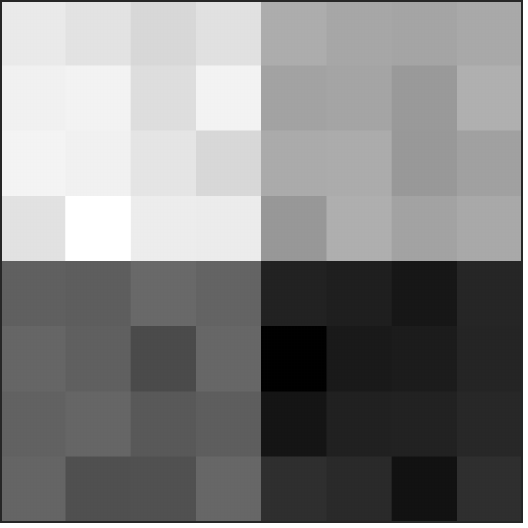}
    \caption{$\boldsymbol{X}$ (SAR) at $t_1$}
    \label{fig1:sar}
  \end{subfigure}\\[1ex]
  \begin{subfigure}{\linewidth}
    \centering
    \includegraphics[width=0.8\linewidth,keepaspectratio]{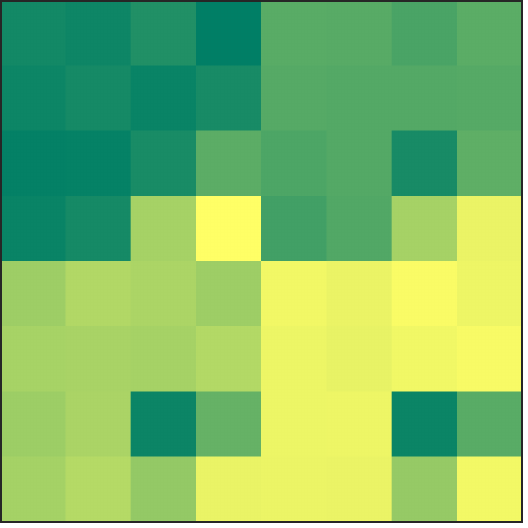}
    \caption{$\boldsymbol{Y}$ (optical) at $t_2$}
    \label{fig1:opt}
  \end{subfigure}
\end{minipage}%
\begin{minipage}{.48\columnwidth}
  \begin{subfigure}{\linewidth}
    \centering
    \includegraphics[width=0.8\linewidth,keepaspectratio]{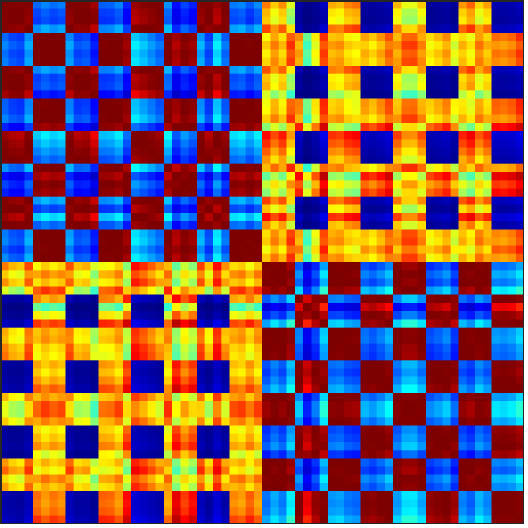}
    \caption{$A^{\mathcal{X}}$}
    \label{fig1:aff1}
  \end{subfigure}\\[1ex]
  \begin{subfigure}{\linewidth}
    \centering
    \includegraphics[width=0.8\linewidth,keepaspectratio]{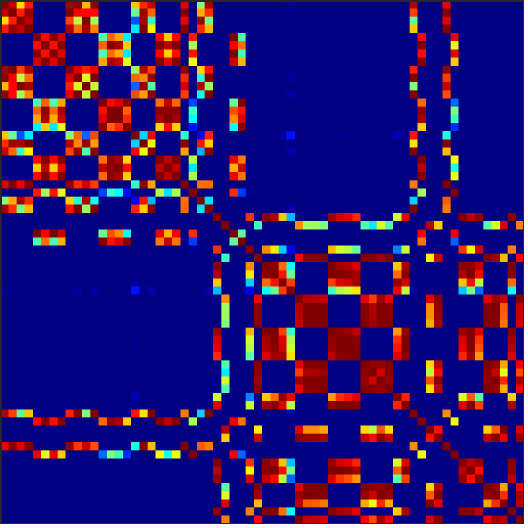}
    \caption{$A^{\mathcal{Y}}$}
    \label{fig1:aff2}
  \end{subfigure}
\end{minipage}%
\begin{minipage}{.48\columnwidth}
  \begin{subfigure}{\linewidth}
    \centering
    \includegraphics[width=0.8\linewidth,keepaspectratio]{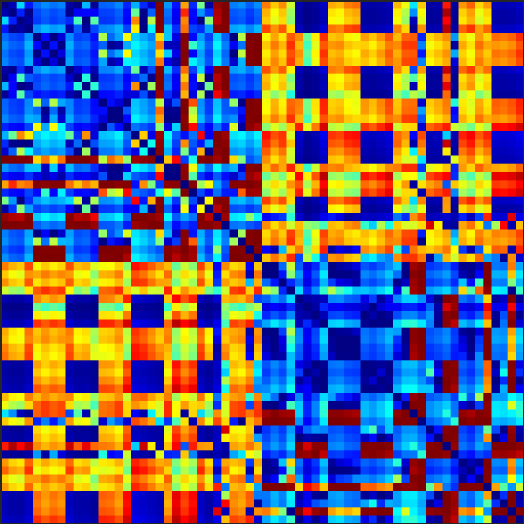}
    \caption{$D=\lvert A^{\mathcal{X}} - A^{\mathcal{Y}} \rvert$}
    \label{fig1:diff}
  \end{subfigure}
\end{minipage}%
\begin{minipage}{.48\columnwidth}
  \begin{subfigure}{\linewidth}
    \centering
    \includegraphics[width=0.8\linewidth,keepaspectratio]{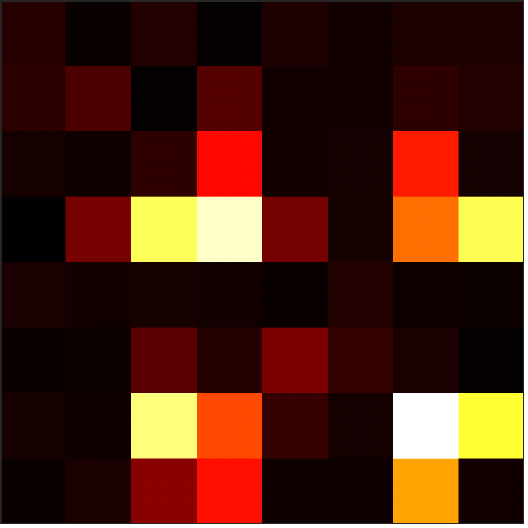}
    \caption{Prior image $\boldsymbol{\alpha}$}
    \label{fig1:test}
  \end{subfigure}\\[1ex]
  \begin{subfigure}{\linewidth}
    \centering
    \includegraphics[width=0.8\linewidth,keepaspectratio]{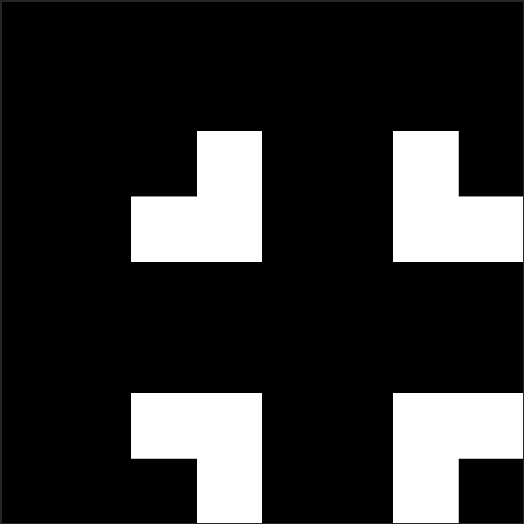}
    \caption{Confusion map (thresholded $\boldsymbol{\alpha}$)}
    \label{fig1:cd}
  \end{subfigure}
\end{minipage}
\hfill

\caption{Toy example. \subref{fig1:sar}) Patch from the SAR image at time $t_1$; \subref{fig1:opt}) Corresponding patch in the optical image at time $t_2$; \subref{fig1:aff1}-\subref{fig1:diff}) Affinity matrices and their absolute difference; \subref{fig1:test}) Prior image $\boldsymbol{\alpha}$ obtained from $D$ by applying \eqref{eq:alpha}; \subref{fig1:cd}) Confusion map obtained by thresholding $\boldsymbol{\alpha}$, with true positives (white) and true negatives (black). Best viewed in colour.}
\label{fig:toy_example}
\end{figure*}

The toy example in Fig.\ \ref{fig:toy_example} helps to explain the effectiveness of the proposed approach.
To make this case easier to explain, $\Delta$ is set equal to $k$: each pixel in the image is covered only once.
Fig.\ \ref{fig1:sar} simulates a patch $\boldsymbol{X}$ of $8 \times 8$ pixels extracted from a SAR image captured at $t_1$.
It consists of four blocks representing four different classes, whose pixel intensities are affected by speckle (large variability associated with the multiplicative signal model of SAR images).
The corresponding patch $\boldsymbol{Y}$ extracted from an optical image at $t_2$ is depicted in Fig.\ \ref{fig1:opt}; The same classes are disposed in the same way and the pixel intensities are affected by additive Gaussian noise.
%Note that the intensities of one class across the two images are not correlated, so they can be similar, as for the classes in the top-right and bottom-left blocks of $p$, or completely different, as for the other two classes. 
Changes are introduced by placing 4 pixels representing each class in the bottom right of each block of $\boldsymbol{Y}$.
In this way, all the possible transitions between one class and the others occur between $t_1$ and $t_2$.
Clearly, a transition from one class to another represents a change, whereas no change occurs when the same class is present at the two dates.
The $64 \times 64$ affinity matrices $A^{\mathcal{X}}$ and $A^{\mathcal{Y}}$ computed from $\boldsymbol{X}$ and $\boldsymbol{Y}$ are depicted in Fig.\ \ref{fig1:aff1} and \ref{fig1:aff2}.
They both show a regular squared pattern, with high affinities in red and low affinities in blue, which corresponds to the the block structure of $\boldsymbol{X}$ and $\boldsymbol{Y}$. Moreover, the latter presents the expected irregularities and perturbations due to the introduced changed pixels that are breaking the block pattern in Fig.\ \ref{fig1:opt}.
Once the change affinity matrix $D$ is evaluated (Fig.\ \ref{fig1:diff}), it can be transformed by \eqref{eq:alpha} into the $8 \times 8$ image of the prior $\alpha_i$ shown in Fig.\ \ref{fig1:test}, where dark (bright) pixels indicate small (large) values of $\alpha_i$. This prior image is denoted $\boldsymbol{\alpha}$. Finally, one may retrieve a CD map by thresholding $\boldsymbol{\alpha}$, which in this case matches the ground truth with $100$\% accuracy, as shown in Fig.\ \ref{fig1:cd} by the confusion map where only true positives (white) and true negatives (black) are present.
\begin{algorithm}
\begin{spacing}{1.2}
\caption{Evaluation of $\boldsymbol{\alpha}$:}\label{alg:aff}
\begin{algorithmic}
\FORALL{patches $p_\ell, \ell \in \{1,\dots,|\mathcal{P}|\}$}
\STATE Compute $d_{i,j}^{m} \> \forall i,j \in p_\ell^{m},  \, m = \boldsymbol{X},\boldsymbol{Y}$
\STATE Determine $h^{\boldsymbol{X}}_\ell$ and $h^{\boldsymbol{Y}}_\ell$
\vspace{0.8pt}
\STATE Compute $A_{i,j}^{m}=\exp\left\{-\left(\frac{d_{i,j}^{m}}{h^{m}_\ell}\right)^2\right\}, \, m = \boldsymbol{X},\boldsymbol{Y}$
\vspace{1pt}
%A_{i,j}^{\boldsymbol{Y}}=\exp\left\{-\left(\frac{d_{i,j}^{\boldsymbol{Y}}}{h^{\boldsymbol{Y}}_p}\right)^2\right\}$
\STATE Compute $\alpha_{i,\ell} = \frac{1}{k^2} \sum_j \lvert A^{\boldsymbol{X}}_{i,j} - A^{\boldsymbol{Y}}_{i,j} \rvert \, \forall i \in p_\ell$ 
\STATE Add $\alpha_{i,\ell}$ to the set $\mathcal{S}^\alpha_i \, \forall i \in p_\ell$
\ENDFOR
\FORALL{pixels $i \in \{1,\dots,N\}$}
\STATE Compute $\alpha_{i} = \frac{1}{\lvert\mathcal{S}_i^\alpha\rvert} \underset{\{\ell\,|\,\alpha_{i,\ell} \in \mathcal{S}_i^\alpha\}}{\sum}\alpha_{i,\ell}$
\ENDFOR
\end{algorithmic}
\end{spacing}
\end{algorithm}

Given the set $\mathcal{P}$ of all the image patches of size $k \times k$ spaced by a step size $\Delta$, Algorithm \ref{alg:aff} summarises the procedure to obtain a set of priors $\{\alpha_i\}_{i=1}^N$ for the whole dataset, which can be rearranged into the image $\boldsymbol{\alpha}\in \mathbb{R}^{H \times W}$.
For each pixel $i\in\{1,\dots,N\}$ in the image, the mean over $\mathcal{S}_i^\alpha$ is computed, where $\mathcal{S}_i^\alpha$ is the set of the $\alpha_{i,\ell}$ obtained with all the patches $p_\ell\in\mathcal{P}$ covering pixel $i$.
If $\Delta$ is a factor of $k$, this average is calculated over $\left(\nicefrac{k}{\Delta}\right)^2$ values.

The size $k$ has an important role in the effectiveness of this methodology, because the patches $p$ could be too small or too big to capture the shapes and the patterns within them.
To reduce the sensitivity to this parameter, one may suggest to use different values of $k$ for Algorithm \ref{alg:aff} and combine the results in an ensemble manner.
For example, once $k$ is defined, the method can be applied also for $k_\textit{small} = \nicefrac{k}{2}$ and $k_\textit{big} = 2 \cdot k$.
However, the size of the matrices containing first $d_{i,j}^{m}$ and then $A_{i,j}^{m}$ exhibits a quadratic growth with respect to $k$, thus becoming quickly unfeasible in terms of memory usage and computational time.
Hence, instead of applying the method to the original images with $k_\textit{big}$, we suggest to down-sample the images by a factor of $2$, apply the algorithm with $k$, and re-scale the output to the original size.
This procedure might introduce artifacts and distortions, but their effects are mitigated when combined with the results obtained with $k_\textit{small}$ and $k$.

In the following subsections, we explain how to exploit the outcome of Algorithm \ref{alg:aff} to train the proposed deep learning architectures in absence of supervision.

\subsection{X-Net: Weighted Translation Network}\label{subsec:crossed}
\begin{figure}[b!]
\centering
\includegraphics[width=0.50\columnwidth]{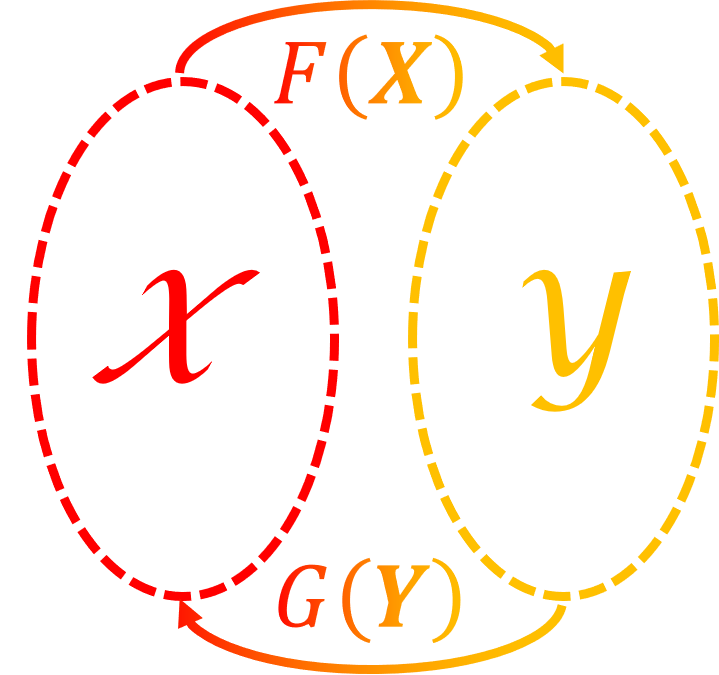}
\caption{First proposed framework, where two domains and two transformations which can translate data across them.}
%\vspace{-0.5cm}
\label{fig:domains1}
\end{figure}
\begin{figure*}[t!]
\centering
\includegraphics[width=1.9\columnwidth]{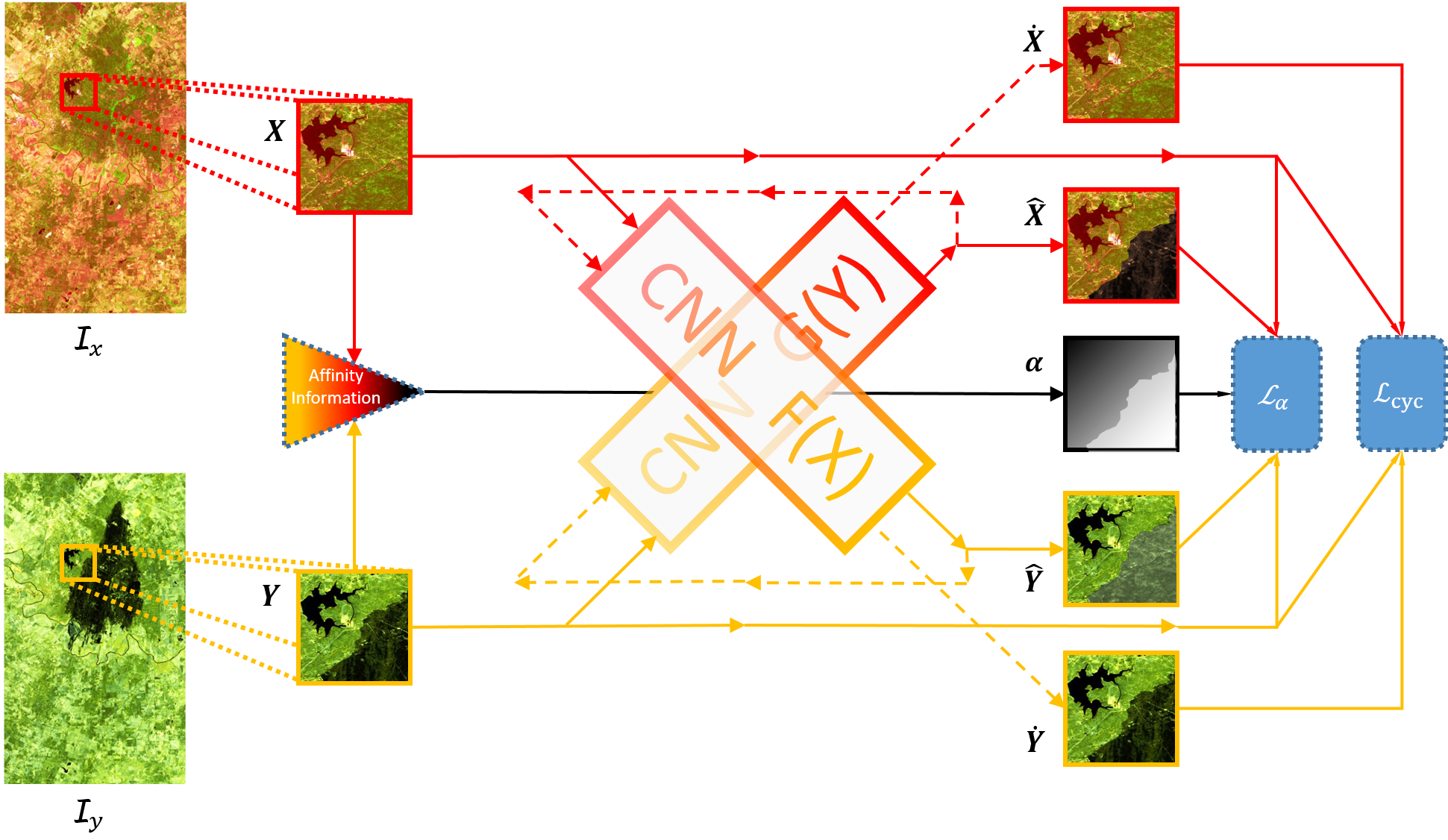}
\caption{Data flow of the X-Net. Two CNNs transform data from the domain of $\boldsymbol{X}$ to the domain of $\boldsymbol{Y}$ and vice versa. Solid lines going through them indicate data transferred from one domain to the other, dashed lines indicate data re-transformed back to their original domain.}
%\vspace{-0.5cm}
\label{fig:xnet}
\end{figure*}
The main goal of our approach is to map data across two domains.
As Fig.\ \ref{fig:domains1} illustrates, this means to train a function $F(\boldsymbol{X})\!:\!\mathcal{X}^{h\times w}\!\to\!\mathcal{Y}^{h\times w}$ to transform data between the domains of $\boldsymbol{X}$ and $\boldsymbol{Y}$, and a second function $G(\boldsymbol{Y}):\mathcal{X}^{h\times w}\to\mathcal{Y}^{h\times w}$ to do the opposite.
The two mapping functions can be implemented as convolutional neural networks (CNNs).
Hence, the training can be carried out by the minimisation of an objective function with respect to the set $\vartheta$ of parameters of the two networks.
The objective function, commonly referred to as the loss function $\mathcal{L}(\vartheta)$, is defined \textit{ad hoc} and usually consists of a weighted sum of loss terms, where each relates to a specific objective or property that we want from the solution.
For this particular framework, we introduce three loss terms.
Note that from now on we refer to training patches of much larger size than the patch size $k$ of \Cref{subsec:affinity} used to compute the affinity-based prior.

% In the loss terms we will need to compute distances between patches, where input patches are compared with translated ones.
% %The comparison cannot be done pixel by pixel, since single pixels in the translated patch are potentially a function of all pixels in the input, depending on the receptive field. 
% We therefore define a general weighted distance between two equal-sized $h\times w$ patches $\boldsymbol{A}$ and $\boldsymbol{B}$ as $\delta(\boldsymbol{A},\boldsymbol{B}|\boldsymbol{\pi})$, where $\boldsymbol{\pi}$ is a vector of weights, each associated with a pixel $i\in\{1,\dots,n\}$ of the patches, with $n=h\cdot w$. In this work we use the mean squared $L_2$ norm as a particular choice of $\delta(\cdot)$, which is appropriate since the pixel measurements $\boldsymbol{a}_i\in\boldsymbol{A}$ and $\boldsymbol{b}_i\in\boldsymbol{B}$ in our datasets are vectors. This means that
% \begin{equation}\label{eq:delta}
% \delta(\boldsymbol{A},\boldsymbol{B}|\boldsymbol{\pi}) = \frac{1}{n}\displaystyle\sum_{i=1}^{n}\pi_i\lVert\boldsymbol{a}_i-\boldsymbol{b}_i\rVert_2^2\,.
% \end{equation}
% When no weights are applied ($\boldsymbol{\pi}=\boldsymbol{1}$, where $\boldsymbol{1}$ denotes a vector of ones), the patch distance is written as $\delta(\boldsymbol{A},\boldsymbol{B}|\boldsymbol{1})=\delta(\boldsymbol{A},\boldsymbol{B})$.
%
\subsubsection{Weighted translation loss}
For a pair of patches $\{\boldsymbol{X},\boldsymbol{Y}\}$, we want in general the domain translation to satisfy:
\begin{equation}\label{eq:approxhat}
\begin{split}
    \hat{\boldsymbol{Y}} = F(\boldsymbol{X}) \simeq \boldsymbol{Y}\,, \\
    \hat{\boldsymbol{X}} = G(\boldsymbol{Y}) \simeq \boldsymbol{X}\,,
\end{split}
\end{equation}
where $\hat{\boldsymbol{Y}} = F(\boldsymbol{X})$ and $\hat{\boldsymbol{X}} = G(\boldsymbol{Y})$ stand for the data transformed from one domain into the other.
However, pixels that are likely to be changed shall not fulfill the same requirements, i.e., condition (\ref{eq:approxhat}) should be satisfied in unchanged areas but should not be enforced in changed ones in order not to hinder the capability of the proposed method to discriminate changes. More formally, if $H_0$ and $H_1$ indicate the ``no-change" and ``change" hypotheses, respectively, then in a least mean-square error (MSE) framework, it would be desired that the network parameters ideally minimized the following MSE conditioned to "no-change":
\begin{equation}\label{eq:idealloss}
    \mathcal{L}_{H_0}(\vartheta)=\mathbb{E}_{\boldsymbol{X},\boldsymbol{Y}}\left[\left.\delta(\boldsymbol{X},\hat{\boldsymbol{X}})\right|H_0\right]+\mathbb{E}_{\boldsymbol{X},\boldsymbol{Y}}\left[\left.\delta(\boldsymbol{Y},\hat{\boldsymbol{Y}})\right|H_0\right],
\end{equation}
where $\delta(\boldsymbol{A},\boldsymbol{B})$ indicates the squared $L_2$ distance between two equal-sized $h\times w$ patches $\boldsymbol{A}$ and $\boldsymbol{B}$, i.e.:
\begin{equation}
    \delta(\boldsymbol{A},\boldsymbol{B})= \frac{1}{h\cdot w}\displaystyle\sum_{i=1}^{h\cdot w}\lVert\boldsymbol{a}_i-\boldsymbol{b}_i\rVert_2^2\,.
\end{equation}
Here, $\boldsymbol{a}_i$ and $\boldsymbol{b}_i$ denote the vectors associated with the $i$-th pixel in patches $\boldsymbol{A}$ and $\boldsymbol{B}$, respectively ($i=1,2,\ldots,h\cdot w$). Estimating the expectations in~\eqref{eq:idealloss} is straightforward using a training set for $H_0$, as it has been done in \cite{touati2020anomaly}. However, a training set is unavailable in the fully unsupervised scenario that is considered here.\\ 
We prove in the Appendix that, under mild conditional independence assumptions, the conditional loss $\mathcal{L}_{H_0}(\vartheta)$ can be equivalently rewritten as:
\begin{equation}\label{eq:idealloss2}
    \mathcal{L}_{H_0}(\vartheta)=\mathbb{E}_{\boldsymbol{X},\boldsymbol{Y}}\left[\delta(\boldsymbol{X},\hat{\boldsymbol{X}}|\boldsymbol\Psi)\right]+\mathbb{E}_{\boldsymbol{X},\boldsymbol{Y}}\left[\delta(\boldsymbol{Y},\hat{\boldsymbol{Y}}|\boldsymbol\Phi)\right],
\end{equation}
where $\delta(\boldsymbol{A},\boldsymbol{B}|\boldsymbol{W})$ indicates a squared $L_2$ distance weighted on a vector of weights $\boldsymbol{W}=[W_1,W_2,\ldots,W_{h\cdot w}]^T$, i.e.: 
\begin{equation}\label{eq:delta}
\delta(\boldsymbol{A},\boldsymbol{B}|\boldsymbol{W}) = \frac{1}{h\cdot w}\displaystyle\sum_{i=1}^{h\cdot w}W_i\lVert\boldsymbol{a}_i-\boldsymbol{b}_i\rVert_2^2\,,
\end{equation}
and where $\boldsymbol\Psi$ and $\boldsymbol\Phi$ are $(h\cdot w)$-dimensional weight vectors whose components are defined in terms of the joint probability distributions of $\boldsymbol{X}$ and $\boldsymbol{Y}$, given $H_0$ and $H_1$. Accordingly, the $H_0$-conditional MSE in (\ref{eq:idealloss}) is equivalent to an unconditional but suitably weighted MSE. In particular, it is also proven in the appendix that the $i$-th component of $\boldsymbol\Psi$ takes values in the interval: 
\begin{equation}
    0<\Psi_i\leq\frac{1}{P(H_0)},
\end{equation}
where $P(H_0)$ is the prior probability of ``no-change." This prior is strictly positive since we assumed at the beginning of this section that the changes affected a limited part of the image. According to a reasoning based on likelihood ratio testing (see Appendix), the lower end $\Psi_i\simeq0$ suggests that the $i$-th pixel of the patch is likely changed, and the upper end $\Psi_i\simeq1/P(H_0)$ suggests that it is likely unchanged ($i=1,2,\ldots,h\cdot w$). The same statement holds for the components of $\boldsymbol\Phi$ as well. This is consistent with the aforementioned interpretation of the equivalence between the $H_0$-conditional non-weighted MSE in (\ref{eq:idealloss}) and the unconditional weighted MSE in (\ref{eq:idealloss2}) because the $i$-th pixel does not contribute to the loss in (\ref{eq:idealloss2}) when it is likely changed. Vice versa, it gives its maximum contribution when it is likely unchanged.

Without training samples, estimating the expectations in (\ref{eq:idealloss2}) is as difficult as estimating those in (\ref{eq:idealloss}) because the weight vectors depend on the joint conditional distributions of $\boldsymbol{X}$ and $\boldsymbol{Y}$.
Therefore, in the proposed method, we leverage on the reformulation as a weighted unconditional MSE in (\ref{eq:idealloss2}) to define an approximation of the $H_0$-conditional MSE in (\ref{eq:idealloss}) by making use of the affinity prior defined in (\ref{eq:alpha}).
As discussed in Section~\ref{subsec:affinity}, every pixel pair $\{\boldsymbol{x}_i,\boldsymbol{y}_i\}$ will be associated with a precomputed prior, $\alpha_i$, that measures through affinity reasoning its chances of being changed.
We exploit this information to approximate (\ref{eq:idealloss2}) as follows:
\begin{equation}\label{eq:weighted}
    \mathcal{L}_{\alpha}\!\left(\vartheta\right) =  \mathbb{E}_{\boldsymbol{X},\boldsymbol{Y}}\!\left[\delta(\hat{\boldsymbol{X}},\boldsymbol{X}|\boldsymbol{\Pi})\right] 
    + \mathbb{E}_{\boldsymbol{X},\boldsymbol{Y}}\!\left[\delta(\hat{\boldsymbol{Y}},\boldsymbol{Y}|\boldsymbol{\Pi})\right]\,,
\end{equation}
%
%where the contribution to the loss of pixel $n$ is weighted by its prior $\Pi\left(\alpha_{n}\right)$, and $\boldsymbol{\alpha}$ holds the $\alpha_n$ associated with the training patches $\boldsymbol{X}$ and $\boldsymbol{Y}$.
where $\boldsymbol{\Pi}=[\Pi(\alpha_1),\dots,\Pi(\alpha_{h\cdot w})]^T$, and $\Pi(\alpha):[0,1]\to[0,1]$ is a monotonically decreasing function that maps $\alpha_i$, measuring the chances of change, into $\Pi_i$, that is used to weigh the contribution of the $i$-th pixel to the loss function ($i=1,2,\ldots,h\cdot w$). Specifically, $\Pi(\alpha_i)$ is supposed to be close to zero when the $i$-th pixel is likely changed (i.e., when $\alpha_i\simeq1$) and close to one when it is likely unchanged (i.e., $\alpha_i\simeq0$). Methodologically, the weighted translation loss $\mathcal{L}_{\alpha}(\vartheta)$ in (\ref{eq:weighted}) is meant as an approximation of (\ref{eq:idealloss2}) -- and thus of the desired conditional non-weighted MSE in (\ref{eq:idealloss}) --, up to a positive multiplicative constant equal to $1/P(H_0)$.
% The key idea is that this approximation can be computed even in a fully unsupervised scenario. In particular, we use the precomputed priors obtained from \Cref{subsec:affinity} to drive the learning process and penalise the contribution of pixels most likely to be affected by changes.
We use the simple $\Pi(\alpha) = 1-\alpha$, but other choices can be considered.

\subsubsection{Cycle-consistency loss}

In their seminal work on CycleGANs~\cite{zhu2017unpaired}, Zhu \textit{et.\ al} pointed out that domain translations should respect the principle of cycle-consistency: Ideally, if $F(\boldsymbol{X})$ and $G(\boldsymbol{Y})$ are perfectly tuned, it must hold true that
\begin{equation}
\begin{split}
    \boldsymbol{\dot{X}} = G(\hat{\boldsymbol{Y}}) = G\left(F(\boldsymbol{X})\right) \simeq \boldsymbol{X}\,, \\
    \boldsymbol{\dot{Y}} = F(\hat{\boldsymbol{X}}) = F\left(G(\boldsymbol{Y})\right) \simeq \boldsymbol{Y}\,,
\end{split}
\end{equation}
where $\boldsymbol{\dot{X}} = G(\hat{\boldsymbol{Y}})$ and $\boldsymbol{\dot{Y}} = F(\hat{\boldsymbol{X}})$ indicate the data re-transformed back to the original domains.
Consequently, the cycle-consistency loss term is defined as:
\begin{equation}
    \mathcal{L}_{\mathrm{cyc}}\left(\vartheta\right) = \mathbb{E}_{\boldsymbol{X}}\left[\delta(\dot{\boldsymbol{X}},\boldsymbol{X})\right] + \mathbb{E}_{\boldsymbol{Y}}\left[\delta(\dot{\boldsymbol{Y}},\boldsymbol{Y})\right]\,.
\end{equation}
%where $\dot{\boldsymbol{x}}\in\dot{\boldsymbol{X}}$ and $\dot{\boldsymbol{y}}\in\dot{\boldsymbol{Y}}$ are cyclically translated pixels corresponding to $\boldsymbol{x}$ and $\boldsymbol{y}$.
Note that training with the cycle-consistency principle does not require paired data.

\subsubsection{Total Loss Function}

The third and last term of the loss function is a weight decay regularisation term, which reduces overfitting by controlling the magnitude of the network parameters $\vartheta$. The total loss function becomes
\begin{equation}
    \mathcal{L}(\vartheta) = \left\{ w_{\mathrm{cyc}} \mathcal{L}_{\mathrm{cyc}}(\vartheta) + w_{\alpha} \mathcal{L}_{\alpha}(\vartheta) + w_{\vartheta} \left\lVert\vartheta\right\rVert^2_2\right\}.
\end{equation}
Optimisation is carried out by seeking its global minimum with respect to $\vartheta$. The weights $w_{\mathrm{cyc}}$, $w_{\alpha}$ and $w_{\vartheta}$ are set to balance the impact of the terms.

Fig.\ \ref{fig:xnet} shows the scheme of the X-Net: One CNN plays the role of $F(\boldsymbol{X})$, the other represents $G(\boldsymbol{Y})$.
Solid lines going through them indicate data transferred from one domain to the other, dashed lines indicate data re-transformed back to their original domain.
The patches from $\boldsymbol{X}$ and $\boldsymbol{Y}$ are used both as input and targets for the CNNs.
Recall that the patch prior $\boldsymbol{\alpha}$ is computed in advance, as explained in \Cref{subsec:affinity}.
For an easier representation, $\boldsymbol{\alpha}$ is deliberately depicted in Fig.\ \ref{fig:xnet} as computed on the fly.

\subsection{ACE-Net: Adversarial Cyclic Encoder Network}
\begin{figure}[ht!]
\centering
\includegraphics[width=0.85\columnwidth]{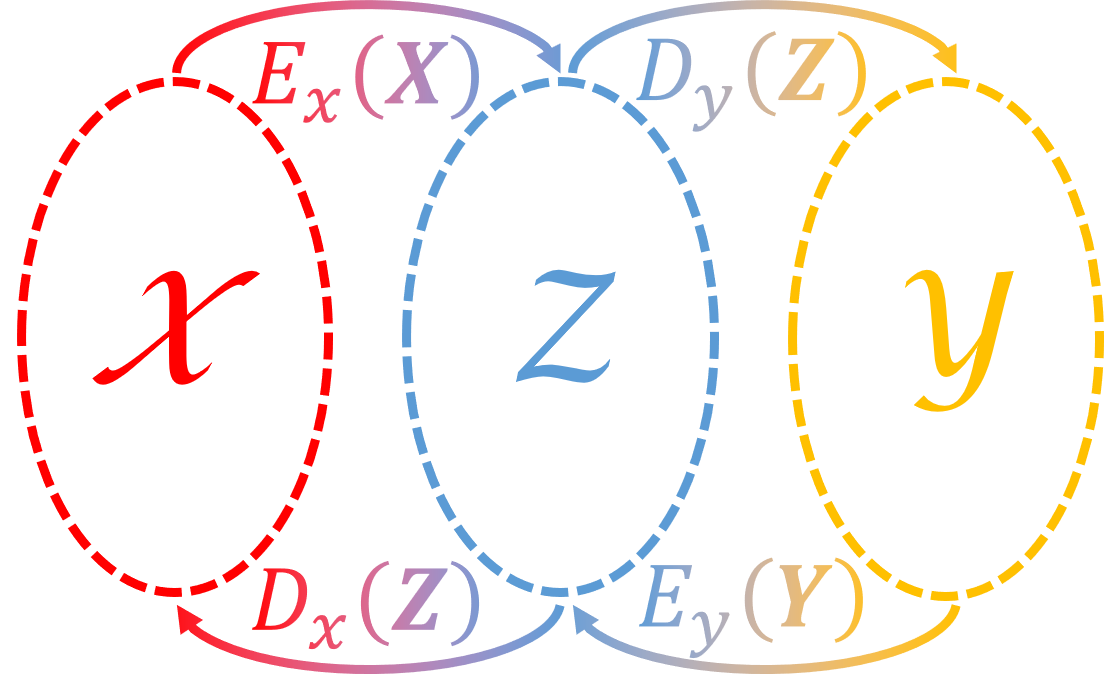}
\caption{Second proposed framework: a latent space $\mathcal{Z}$ is introduced between domains $\mathcal{X}$ and $\mathcal{Y}$, and four regression functions mapping data across them. In this case, $F(\boldsymbol{X})=D_\mathcal{Y}\left(E_\mathcal{X}\left(\boldsymbol{X}\right)\right)$ and $G(\boldsymbol{Y}) = D_\mathcal{X}\left(E_\mathcal{Y}\left(\boldsymbol{Y}\right)\right)$.}
%\vspace{-0.5cm}
\label{fig:domains2}
\end{figure}
\begin{figure*}[ht!]
\centering
\includegraphics[width=1.9\columnwidth]{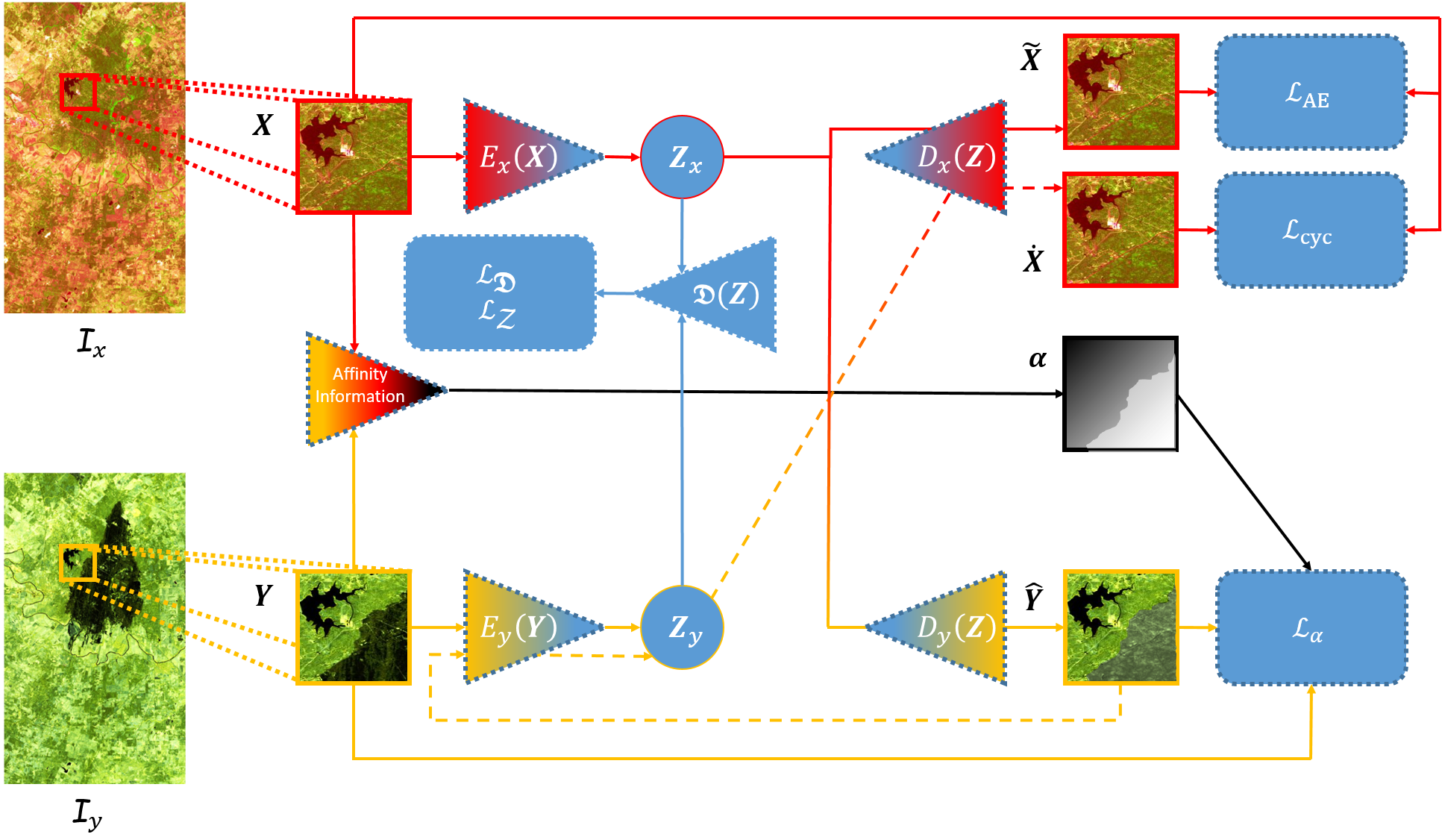}
\caption{Data flow of the ACE-Net. The encoders $E_\mathcal{X}\left(\boldsymbol{X}\right)$ and $E_\mathcal{Y}\left(\boldsymbol{Y}\right)$ transform incompatible data into two code spaces, which are aligned by adversarial training against the discriminator $\mathfrak{D}\left(\boldsymbol{Z}\right)$. The decoders $D_\mathcal{X}\left(\boldsymbol{Z}\right)$ and $D_\mathcal{Y}\left(\boldsymbol{Z}\right)$ are taught to map data from the latent space back into the original spaces. For simplicity, only the loss terms related to $\boldsymbol{X}$ and their corresponding data flows are depicted. Dash lines refer to data which have been transformed already once, have gone through the framework again and have been transformed back into their original domain.}
%\vspace{-0.5cm}
\label{fig:ACEnet}
\end{figure*}

Inspired by Murez \emph{et al.} \cite{murez2018image}, we expand the X-Net framework by introducing a latent space $\mathcal{Z}$ between domain $\mathcal{X}$ and domain $\mathcal{Y}$.
Differently from the X-Net, this architecture consists of five CNNs. The first four networks are image regression functions (see Fig.\ \ref{fig:domains2}): Encoders $E_\mathcal{X}\left(\boldsymbol{X}\right):\mathcal{X}^{h\times w}$ and $E_\mathcal{Y}\left(\boldsymbol{Y}\right):\mathcal{Y}^{h\times w}$ transform data from the original domains into the new common space and a representation referred to as the code: $\boldsymbol{Z}\in\mathcal{Z}^{h\times w}$. Note that the spatial dimensions of $\boldsymbol{Z}$, $h$ and $w$, are equal to those of $\boldsymbol{X}$ and $\boldsymbol{Y}$. This is an empirical choice, as this is seen to produce best image translation and change detection performance. Bottlenecking (dimensionality reduction) at the code layer is not needed for regularisation, as with conventional autoencoders, due to the constraints imposed by loss functions associated with cross-domain mapping. The decoders $D_\mathcal{X}\left(\boldsymbol{Z}\right):\mathcal{Z}^{h\times w}\to\mathcal{X}^{h\times w}$ and $D_\mathcal{Y}\left(\boldsymbol{Z}\right):\mathcal{Z}^{h\times w}\to\mathcal{Y}^{h\times w}$ map latent space data back into their original domains. The fifth network is a discriminator, which is described later.

Despite the added complexity, is simple to notice an analogy between the two schemes,
namely: $F(\boldsymbol{X})=D_\mathcal{Y}\left(E_\mathcal{X}\left(\boldsymbol{X}\right)\right)$ and $G(\boldsymbol{Y}) = D_\mathcal{X}\left(E_\mathcal{Y}\left(\boldsymbol{Y}\right)\right)$.
Therefore, we can include the same loss terms that the X-Net uses: weighted translation loss and cycle-consistency loss, in addition to the weight decay regularisation term.
In this case,
\begin{equation}
\begin{split}
    \hat{\boldsymbol{X}} &= G(\boldsymbol{Y}) = D_\mathcal{X}\left(E_\mathcal{Y}\left(\boldsymbol{Y}\right)\right)\,, \\
    \hat{\boldsymbol{Y}} &= F(\boldsymbol{X}) = D_\mathcal{Y}\left(E_\mathcal{X}\left(\boldsymbol{X}\right)\right)\,, \\
    \boldsymbol{\dot{X}} &= G(\hat{\boldsymbol{Y}}) = D_\mathcal{X}\left(E_\mathcal{Y}\left(D_\mathcal{Y}\left(E_\mathcal{X}\left(\boldsymbol{X}\right)\right)\right)\right)\,, \\
    \boldsymbol{\dot{Y}} &= F(\hat{\boldsymbol{X}}) = D_\mathcal{Y}\left(E_\mathcal{X}\left(D_\mathcal{X}\left(E_\mathcal{Y}\left(\boldsymbol{Y}\right)\right)\right)\right)\,.
\end{split}
\end{equation}
Nonetheless, the ACE-Net framework allows to define two additional loss terms.

\subsubsection{Reconstruction Loss}

The composite functions $D_{\mathcal X}(E_{\mathcal X}(\boldsymbol{X}))$ and $D_{\mathcal Y}(E_{\mathcal Y}(\boldsymbol{Y}))$ constitute autoencoders, whose goal is to reproduce their input as faithfully as possible in output. This means that the reconstructed images $\Tilde{\boldsymbol{X}}$ and $\Tilde{\boldsymbol{Y}}$ must satisfy:
\begin{equation}
\begin{split}
    \Tilde{\boldsymbol{X}} &= D_\mathcal{X}\left(E_\mathcal{X}\left(\boldsymbol{X}\right)\right) \simeq \boldsymbol{X}\,, \\
    \Tilde{\boldsymbol{Y}} &= D_\mathcal{Y}\left(E_\mathcal{Y}\left(\boldsymbol{Y}\right)\right) \simeq \boldsymbol{Y}\,. 
\end{split}
\end{equation}
Consequently, we introduce the reconstruction loss term:
\begin{equation}\label{eq:recon}
    \mathcal{L}_{\mathrm{AE}}\left(\vartheta_{\mathrm{AE}}\right) = \mathbb{E}_{\boldsymbol{X}}\left[\delta(\tilde{\boldsymbol{X}},\boldsymbol{X})\right] + \mathbb{E}_{\boldsymbol{Y}}\left[\delta(\tilde{\boldsymbol{Y}},\boldsymbol{Y})\right]\,,
\end{equation}
where $\vartheta_{\mathrm{AE}}$ denotes all parameters in the autoencoders, consisting of $E_\mathcal{X}(\boldsymbol{X})$, $D_\mathcal{Y}(\boldsymbol{Z})$, $E_\mathcal{Y}(\boldsymbol{Y})$ and $D_\mathcal{X}(\boldsymbol{Z})$.
\subsubsection{Adversarial Code Alignment Losses}

Even after implementing the cycle-consistency loss and the weighted translation loss, there is no guarantee that the latent domain is the same for both AEs. Although the code layers might align in distribution, there is still a risk that class signatures do not correspond due to mode swapping or other perturbations in feature space. To ensure that they align both in distribution and in feature space location of classes, we apply adversarial training and feed a discriminator with a stack of the two codes.
The discriminator $\mathfrak{D}(\boldsymbol{Z}):\mathcal{Z}^{h\times w}\to[0,1]$ is rewarded if it is able to distinguish the codes, whereas the generators (i.e.\ the encoders) are penalised when the discriminator succeeds.
Let successful discrimination be defined as: $\mathfrak{D}(E_\mathcal{X}(\boldsymbol{X}))=1$ and $\mathfrak{D}(E_\mathcal{Y}(\boldsymbol{Y}))=0$. Thus, the last two loss terms become:
\begin{equation}
    \!\!\mathcal{L}_{\mathfrak D}\!\left(\vartheta_{\mathfrak D}\right) =  \mathbb{E}_{\boldsymbol{X}}\!\!\left[\left(\mathfrak{D}\!\left(E_\mathcal{X}\!\left(\boldsymbol{X}\right)\right)\!-\!1\right)^2\right] + \mathbb{E}_{\boldsymbol{Y}}\!\!\left[\mathfrak{D}\!\left(E_\mathcal{Y}\!\left(\boldsymbol{Y}\right)\right)^2\right]
\label{eq:discr}
\end{equation}
%\begin{equation}
%\begin{split}
%    \mathcal{L}_{\mathfrak D}\left(\vartheta_{\mathfrak D}\right) = & \mathbb{E}_{\boldsymbol{X}}\left[\left(\mathfrak{D}\!\left(E_\mathcal{X}\!\left(\boldsymbol{X}\right)\right)-1\right)^2\right] + \\
%    & \mathbb{E}_{\boldsymbol{Y}}\left[\mathfrak{D}\!\left(E_\mathcal{Y}\!\left(\boldsymbol{Y}\right)\right)^2\right]\,,
%\end{split}
%\label{eq:code}
%\end{equation}
\begin{equation}
    \!\!\mathcal{L}_{\mathcal Z}\!\left(\vartheta_{E}\right) =  \mathbb{E}_{\boldsymbol{X}}\!\!\left[\mathfrak{D}\!\left(E_\mathcal{X}\!\left(\boldsymbol{X}\right)\right)^2\right] + 
    \mathbb{E}_{\boldsymbol{Y}}\!\!\left[\left(\mathfrak{D}\!\left(E_\mathcal{Y}\!\left(\boldsymbol{Y}\right)\right)\!-\!1\right)^2\right]
\label{eq:code}
\end{equation}
%\begin{equation}
%\begin{split}
%    \mathcal{L}_{\boldsymbol Z}\left(\vartheta_{E}\right) = & \mathbb{E}_{\boldsymbol{X}}\left[\mathfrak{D}\!\left(E_\mathcal{X}\!\left(\boldsymbol{X}\right)\right)^2\right] + \\
%    & \mathbb{E}_{\boldsymbol{Y}}\left[\left(\mathfrak{D}\!\left(E_\mathcal{Y}\!\left(\boldsymbol{Y}\right)\right)-1\right)^2\right]\,,
%\end{split}
%\label{eq:code}
%\end{equation}
where the discrimination loss $\mathcal{L}_{\mathfrak D}$ is used to adjust the parameters $\vartheta_{\mathfrak D}$ of the discriminator. The code layer is used as generator, and the code loss $\mathcal{L}_{\mathcal Z}$ is used to train the parameters $\vartheta_{E}$ of the encoders $E_\mathcal{X}(X)$ and $E_\mathcal{Y}(Y)$ that generate the codes.
The adversarial scheme is evident from \eqref{eq:discr} and \eqref{eq:code}, the two generators and the discriminator aim at the opposite goal and, therefore, have opposite loss terms.
%The labels are set arbitrarily, but one can swap them as long as Eq.\ \ref{eq:discr} and \ref{eq:code} are swapped as well.
As in \cite{zhu2017unpaired}, we choose an adversarial objective function based on mean squared errors rather than a logarithmic one.
Note that two discriminators could also have been placed after the decoders to distinguish transformed \textit{fake} data from the reconstructed ones, as in \cite{gong2019coupling}.
However, to train two additional networks and find a good balance between all the involved parties is not trivial and require the correct design of each and every network in the architecture, on top of which fine-tuning of all the involved weights must be carried out.
In conclusion, we decided to have a less complex framework with just one discriminator for the code space.

\subsubsection{Total loss function}

The total loss function $\mathcal{L}(\vartheta)$ in this case is composed of six terms:
\begin{equation}
\begin{split}
    \mathcal{L}(\vartheta) = & w_{\mathrm{adv}}\left[\mathcal{L}_{\mathcal Z}\left(\vartheta_{E}\right)+\mathcal{L}_{\mathfrak D}\left(\vartheta_{\mathfrak D}\right)\right] + \\
    & w_{\mathrm{AE}} \mathcal{L}_{\mathrm{AE}}\left(\vartheta_{\mathrm{AE}}\right) + w_{\mathrm{cyc}} \mathcal{L}_{\mathrm{cyc}}(\vartheta_{\mathrm{AE}}) + \\
    & w_{\alpha} \mathcal{L}_{\alpha}(\vartheta_{\mathrm{AE}}) + w_{\vartheta} \left\lVert\vartheta\right\rVert^2_2 \, .
\end{split}
\end{equation}
The weights balancing the adversarial losses ($w_{\mathrm{adv}}$), the reconstruction loss ($w_{\mathrm{AE}}$), the cycle-consistency loss ($w_{\mathrm{cyc}}$), the weighted translation loss ($w_{\alpha}$), and the weight regularisation ($w_{\vartheta}$) must be tuned.

Fig.\ \ref{fig:ACEnet} show the schematics of the ACE-Net.
For simplicity, the arrows represent the data flow involving only the loss terms related to $\boldsymbol{X}$.
$\boldsymbol{Y}$ in this image is used only to produce its code and as a target for translation from $\boldsymbol{X}$.
The flow diagram for loss terms related to $\boldsymbol{Y}$ would be symmetric.
Solid arrows represent images going through the encoder-decoder pairs only once (namely $\Tilde{\boldsymbol{X}}$ and $\hat{\boldsymbol{Y}}$), dashed arrows are the second half of the cycle leading to $\dot{\boldsymbol{X}}$.
The discriminator $\mathfrak{D}\left(\boldsymbol{Z}\right)$ takes as input $E_\mathcal{X}\left(\boldsymbol{X}\right)$ and $E_\mathcal{Y}\left(\boldsymbol{Y}\right)$ and tries to tell them apart.

\subsection{Change extraction}

At this stage of the proposed methodology, any homogeneous change detection technique could be used to highlight changes. Among these, we must choose the most appropriate according to the characteristics of the data. However, the translated images go through severely nonlinear transformations, and defining an analytical model describing their statistics is not trivial. Moreover, the main objective of this work is to propose two translation methods, whose contribution might be concealed by a more complex homogeneous change detection approach. Therefore, image subtraction is the most appropriate operation: its requirement is that the original images and the translated ones are in the same domain, which is the final goal of the translation networks.

Once the X-Net and the ACE-Net are trained and the transformed images $\hat{\boldsymbol{X}}$ and $\hat{\boldsymbol{Y}}$ obtained, the elements of two distance images $d^\mathcal{X}$ and $d^\mathcal{Y}$ can be computed as the vector norms of the pixel-wise subtractions
\begin{equation*}
d^{\mathcal{X}}_i = \lVert\hat{\boldsymbol{x}}_i-\boldsymbol{x}_i\rVert_2\quad \mathrm{and}\quad
d^{\mathcal{Y}}_i = \lVert\hat{\boldsymbol{y}}_i-\boldsymbol{y}_i\rVert_2
\end{equation*}
for all pixels $i\in\{1,\dots,N\}$, where $\boldsymbol{x}_i$, $\boldsymbol{y}_i$, $\hat{\boldsymbol{x}}_i$ and $\hat{\boldsymbol{y}}_i$ represent, respectively, pixels of $\boldsymbol{X}$, $\boldsymbol{Y}$, $\hat{\boldsymbol{X}}$ and $\hat{\boldsymbol{Y}}$.
\iffalse
\begin{equation}
\begin{aligned}
  d_m^{\boldsymbol{X}} = \left\lVert\hat{\boldsymbol{x}}_m-\boldsymbol{x}_m\right\rVert_2\,, \\
  d_m^{\boldsymbol{Y}} = \left\lVert\hat{\boldsymbol{y}}_m-\boldsymbol{y}_m\right\rVert_2
\end{aligned}
\, m \in \{1,\dots,M\}
\end{equation}
%
\fi
These difference images are normalised and combined together so that changes are highlighted, whereas false alarms that are present in only one of the two distance images are suppressed.
Outliers might affect the two normalisations, so the distances in $d^{\mathcal{X}}$ and $d^{\mathcal{Y}}$ beyond three standard deviations of the mean values are clipped. We combine the normalised distance images with a simple average and obtain the final difference image $d$.
The latter is then filtered and thresholded to achieve a binary segmentation, which provides the final goal of a CD method: the change map.

Concerning filtering, the method proposed in~\cite{krahenbuhl2011efficient} is used.
It exploits spatial context to filter $d$ with a fully connected conditional random field model.
It defines pairwise edge potentials between all pairs of pixels in the image by a linear combination of Gaussian kernels in an arbitrary feature space.
The main downside of the iterative optimisation of the random field is that it requires the propagation of all the potentials across the image.
However, this highly efficient algorithm reduces the computational complexity from quadratic to linear in the number of pixels by approximating the random field with a mean field whose iterative update can be computed using Gaussian filtering in the feature space.
The number of iterations and the kernel width of the Gaussian kernels are the only hyperparameters manually set, and we opted to tune them according to \cite{luppino2019unsupervised}: $5$ iterations and a kernel width of $0.1$.

Finally, it is fundamental to threshold the filtered difference image correctly: a low threshold yields unnecessary false alarms. Vice versa, a high threshold increases the number of missed changes.
%Determining this threshold heuristically is not convenient, because it requires a visual inspection and an arbitrary evaluation.
Methods such as~\cite{otsu1979threshold,kapur1985new,shanbhag1994utilization,yen1995new} are able to set the threshold automatically. %, therefore they are preferable.
Among these, we selected the well known Otsu's method~\cite{otsu1979threshold}.

%The most important aspect of the compared architectures is their ability to transform the data and, consequently, the quality of the obtained difference image $d$, whereas the image processing applied on the latter should not be relevant for the analysis.
%Therefore, although \cite{liu2016deep} and \cite{niu2018conditional} deploy different filtering and thresholding techniques, the methods selected in this work are used on all the difference images for a fair comparison of the final change maps.

\section{Experimental results}\label{sec:results}

First, the three datasets used in this work are presented in \Cref{subsec:data}.
\Cref{subsec:net_sett} provides the details of our experimental setup.
Then, the proposed prior computation is compared against its previous version in \Cref{subsec:pre_comp}.
For simplicity, we refer to the latter as prior computation (PC) and to the former as improved PC (IPC).
The improvements are demonstrated by qualitative comparisons and further reflected in reductions of the computation time.
Finally, in \Cref{subsec:res} the performance of the proposed networks is compared against the one obtained with several methods from the heterogeneous CD literature.
Along with the mean elapsed times, this section reports the area under the curve (AUC), the overall accuracy (OA), the $F1$ score and Cohen’s Kappa Coefficient $\kappa$~\cite{cohen1960coefficient}. 

The experiments were performed on a machine running Ubuntu 14 with a 8-core CPU @ $2.7$ GHz.
Moreover, $64$ GB of RAM and an NVIDIA GeForce GTX TITAN X (Maxwell) allowed to reduce considerably the training times through parallel computation.
The methods were all implemented in Python using TensorFlow 1.4.0.

\subsection{Datasets}\label{subsec:data}

\subsubsection{Forest fire in Texas}
\begin{figure}[hb!]

\begin{subfigure}[t]{0.30\columnwidth}
\includegraphics[width=\linewidth,keepaspectratio]{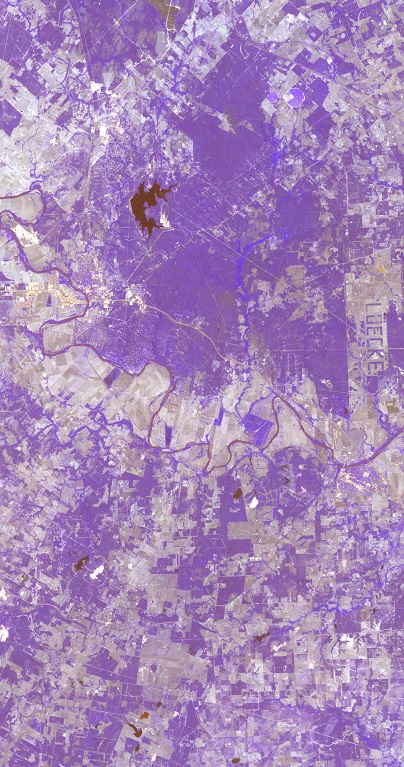}
\caption{Landsat 5 ($t_1$)}
\label{fig:L5}
\end{subfigure}
\hfill
\begin{subfigure}[t]{0.30\columnwidth}
\includegraphics[width=\linewidth,keepaspectratio]{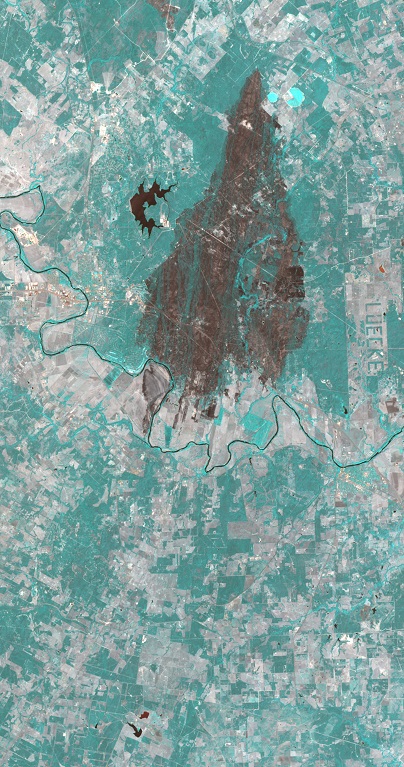}
\caption{EO-1 ALI ($t_2$)}
\label{fig:ALI}
\end{subfigure}
\hfill
\begin{subfigure}[t]{0.30\columnwidth}
\includegraphics[width=\linewidth,keepaspectratio]{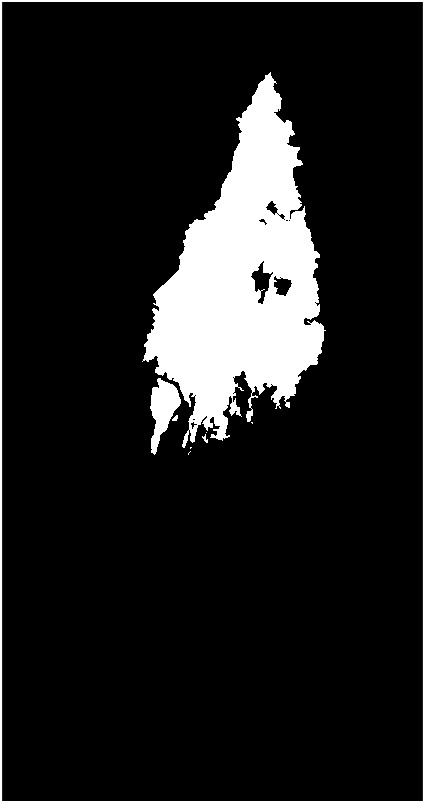}
\caption{Ground Truth}
\label{fig:GT1}
\end{subfigure}
\caption{Forest fire in Texas. Landsat 5 ($t1$), (b) EO-1 ALI ($t2$), (c) ground truth.}
\label{fig:dataset1}
\end{figure}

Bastrop County in Texas was struck by a forest fire during September-October, 2011. The Landsat 5 TM and the Earth Observing-1 Advanced Land Imager (EO-1 ALI) acquired two multispectral optical images before and after the event.
The resulting co-registered and cropped images of size $1520 \times 800$ are displayed in false colour in Fig.\ \ref{fig:L5} and Fig. \ref{fig:ALI}\footnote{Distributed by LP DAAC, http://lpdaac.usgs.gov \label{foot1}}.
Some of the spectral bands of the instruments ($7$ and $10$ in total, respectively) overlap, so the signatures of the land covers involved are partly similar.
Volpi \textit{et al.}~\cite{volpi2015spectral} provided the ground truth shown in Fig.\ \ref{fig:GT1}.

\subsubsection{Flood in California}
\begin{figure}[ht!]
\begin{subfigure}[t]{0.30\columnwidth}
\includegraphics[width=\linewidth,keepaspectratio]{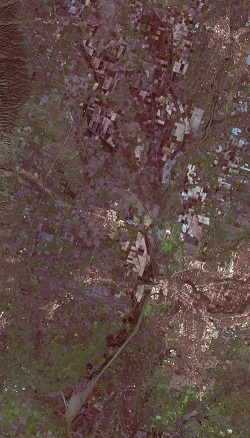}
\caption{Landsat 8 ($t_1$)}
\label{fig2:L8}
\end{subfigure}
\hspace*{\fill}%
\begin{subfigure}[t]{0.30\columnwidth}
\includegraphics[width=\linewidth,keepaspectratio]{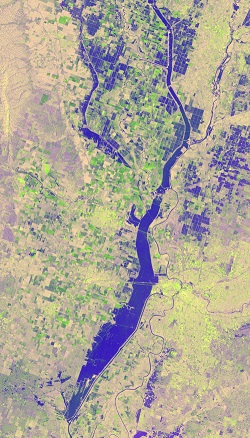}
\caption{Sentinel-1A ($t_2$)}
\label{fig2:S1A}
\end{subfigure}
\hspace*{\fill}%
\begin{subfigure}[t]{0.30\columnwidth}
\includegraphics[width=\linewidth,keepaspectratio]{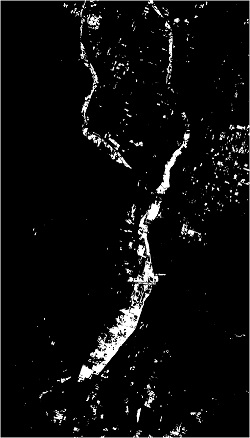}
\caption{Ground Truth}
\label{fig:GT2}
\end{subfigure}
\caption{Flood in California. (a) Landsat 8 ($t_1$), (b) Sentinel-1A ($t_2$), (c) ground truth.}
\label{fig2:dataset2}
\end{figure}

Fig.\ \ref{fig2:L8} displays the RGB channels of a Landsat 8 acquisition$^1$ covering Sacramento County, Yuba County and Sutter County, California, on 5 January 2017.
The OLI and TIRS sensors on Landsat 8 together acquire data in $11$ channels, from deep blue up to thermal infrared.
The same area was affected by a flood, as can be seen in Fig.\ \ref{fig2:S1A}.
This is a Sentinel-1A\footnote{Data processed by ESA, http://www.copernicus.eu/} acquisition, recorded in polarisations VV and VH on 18 February 2017.
The ratio between the two intensities is included both as the blue component of the false colour composite in \ \ref{fig2:S1A} and as the third channel provided as input to the networks.
The ground truth in Fig.\ \ref{fig:GT2} is provided by Luppino \textit{et al.}~\cite{luppino2019unsupervised}.
Originally of $3500 \times 2000$ pixels, these images were resampled to $850 \times 500$ pixels to reduce the computation time.

\subsubsection{Constructions in China}

\begin{figure}[hb!]
\begin{subfigure}[t]{0.30\columnwidth}
\includegraphics[width=\linewidth,keepaspectratio]{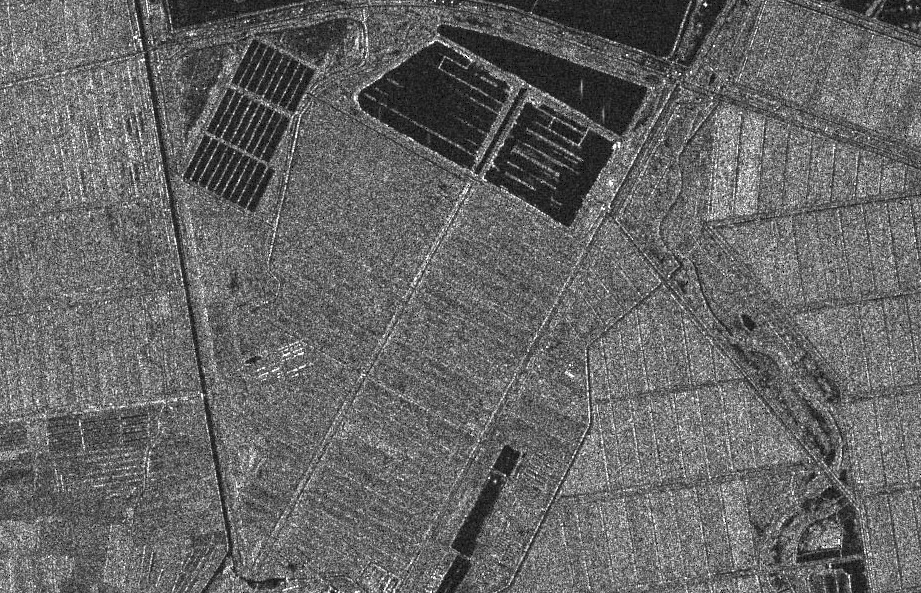}
\caption{SAR image at $t_1$}
\label{fig3:RS2}
\end{subfigure}
\hspace*{\fill}%
\begin{subfigure}[t]{0.30\columnwidth}
\includegraphics[width=\linewidth,keepaspectratio]{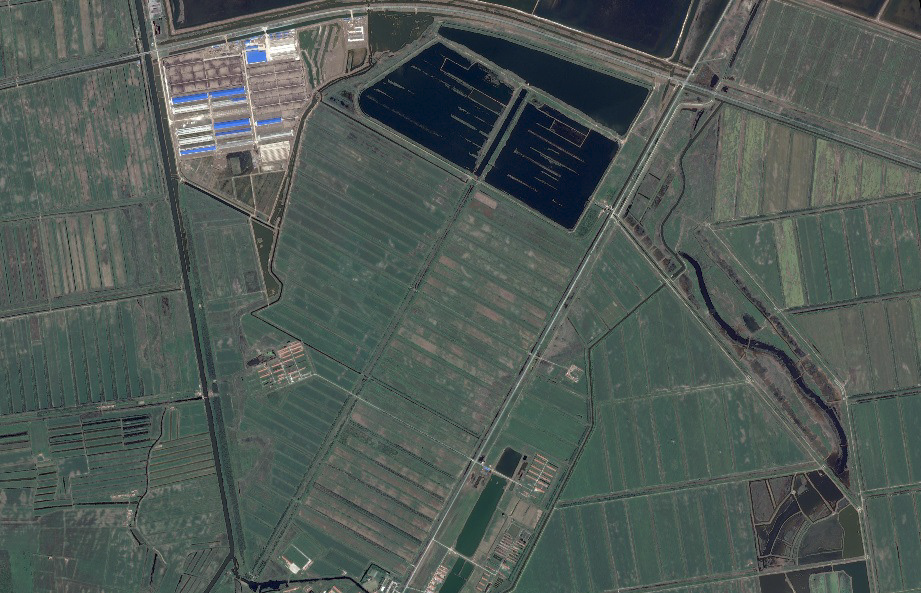}
\caption{Optical RGB image at $t_2$}
\label{fig3:QBL7}
\end{subfigure}
\hspace*{\fill}%
\begin{subfigure}[t]{0.30\columnwidth}
\includegraphics[width=\linewidth,keepaspectratio]{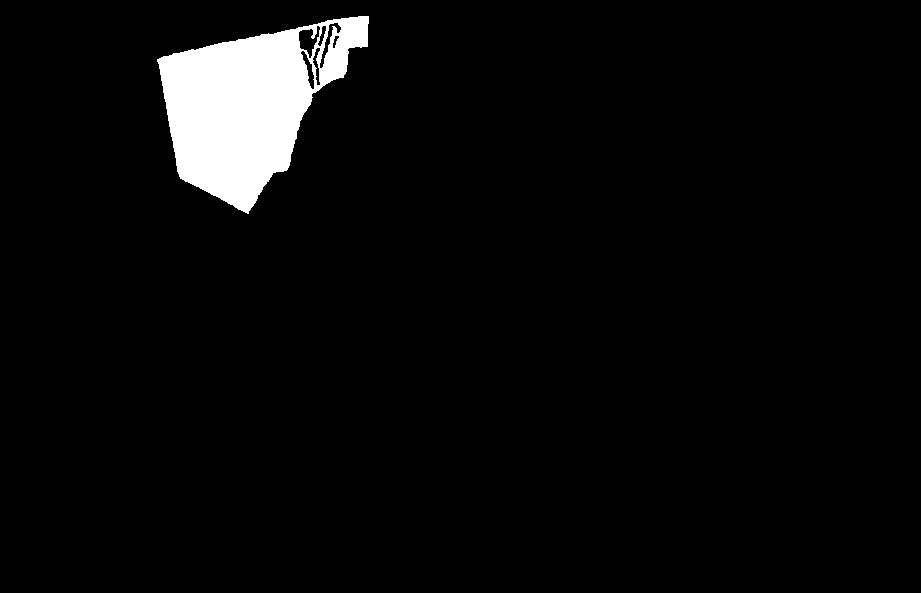}
\caption{Ground Truth}
\label{fig3:GT3}
\end{subfigure}
\caption{Constructions in China. (a) RADARSAT-2 ($t_1$), (b) Quickbird / Landsat 7 ($t_2$), (c) ground truth.}
\label{fig3:dataset3}
\end{figure}

The SAR image in Fig.\ \ref{fig3:RS2} and the coregistered optical image in Fig.\ \ref{fig3:QBL7} were acquired in June 2008 and in September 2012 respectively over the Shuguang village next to Dongying City, China.
Both images have $593 \times 921$ pixels with a spatial resolution of 8 meters, and the ground truth in Fig.\ \ref{fig3:GT3} highlights the edification of buildings which took the place of some farmlands.

\subsection{Experimental setup}\label{subsec:net_sett}

\subsubsection{X-Net and ACE-Net}

For the design of the proposed methods, we opted for CNNs with fully convolutional layers.
One of the advantages is their flexibility with respect to the input size.
At first, one can use batches of small patches extracted from the original images for the training, but once this stage is over, the banks of filters can be applied directly to the whole dataset at once.

Since the goal is to transform each pixel from one domain to another and regularisation of the autoencoders is efficiently handled by other network constraints, there is no need to have a bottleneck in the code layer of the ACE-Net, that is, to reduce the size of the input height and width to compress the data.
Hence, $3 \times 3$ filters were applied without stride on the input patches, whose borders were padded with zeros.
In the X-Net, both networks have four layers:
The first three consist of $100$, $50$, and $20$ filters; The last layer matches the number of channels of the translated data, with $|\mathcal{Y}|$ filters for $F(\boldsymbol{X})$ and $|\mathcal{X}|$ filters for $G(\boldsymbol{Y})$.
The encoders of the ACE-Net have three layers of $100$, $50$, and $20$ filters, and these numbers are reversed for the decoders.
The ACE-Net discriminator is the only network which, after three convolutional layers with $64$, $32$, and $16$ filters, deploys a fully-connected layer with one output neuron.

Concerning the activation functions, a leaky ReLU~\cite{maas2013rectifier} was chosen with the slope for negative arguments set equal to $\beta=0.3$.
The last layer of each network represents an exception: The sigmoid was selected for the discriminator, which must provide outputs between $0$ and $1$, whereas for every other network the hyperbolic tangent was chosen because our data was normalised between $-1$ and $1$.
With this range of data values the training was sped up as expected~\cite{lecun2012efficient}.
Batch normalisation~\cite{ioffe2015batch} turned out to be unnecessary and was discarded, as it did not improve the optimisation and it actually slowed down our experiments.

After each layer, dropout is applied with a dropout rate of $20\%$ during the training phase to enhance the robustness of the framework against overfitting and input noise~\cite{srivastava2014dropout}.
Also, data augmentation helps increasing the size of the training sample by introducing some more variety in the data: Before feeding the patches to the network, these were randomly flipped and rotated.

The weights in $\boldsymbol{\vartheta}$ were initialised with a truncated normal distribution according to~\cite{glorot2010understanding} and the biases were initialised as zeros.
For every epoch of the training $10$ batches were used, each containing $10$ patches of size $100 \times 100$.
The Adam optimizer~\cite{reddi2018on} minimised the loss function for $240$ epochs at a learning rate of $10^{-5}$.
The weights of the loss functions in the ACE-Net are five: $w_{\mathrm{adv}}=1$; $w_{\mathrm{AE}}=0.2$; $w_{\mathrm{cyc}}=2$; $w_{\alpha}=3$; and $w_{\vartheta}=0.001$. 
The X-Net uses only three of these, namely $w_{\mathrm{cyc}}$, $w_{\alpha}$ and $w_{\vartheta}$, and the same values were used for these.

After several training epochs, a preliminary evaluation of the difference image $d$ is computed and scaled to fall into the range $[0,1]$, and the prior is updated as $\boldsymbol{\Pi} = 1 - d$.
In this way, pixels associated with a large $d$ entry are penalised by a small weight, whereas the opposite happens to pixels more likely to be unchanged.
The $\boldsymbol{\Pi}$ is updated at two milestones placed at one third and two thirds of the total epochs, namely at epoch $80$ and epoch $160$.
This form of self-supervision paradigm has already proven robust in other tasks such as deep clustering~\cite{caron2018deep} and deep image recovery~\cite{ulyanov2018deep}.

\subsubsection{SCCN and CAN}

We implemented two methods as state-of-the-art competitors, namely SCCN~\cite{liu2016deep} and the conditional adversarial network in~\cite{niu2018conditional}, which is from now on referred to as CAN.
A brief description of these methods can be found in the last paragraph of \Cref{sec:sccn} and \Cref{sec:cgan}, respectively.

The most important aspect of the compared architectures is their ability to transform the data and, consequently, the quality of the obtained difference image $d$, whereas the postprocessing applied to $d$ is not considered relevant in the present comparison.
Therefore, although~\cite{liu2016deep} and~\cite{niu2018conditional} deploy different filtering and thresholding techniques, the methods selected in this work are used on all the difference images for a fair comparison of the final change maps.
The implementations of the SCCN and the CAN were as faithful as possible based on the details shared in~\cite{liu2016deep} and~\cite{niu2018conditional}.
%
\iffalse
Notwithstanding, some aspects of the SCCN were not explained in detail.
In~\cite{liu2016deep}, a factor $\lambda$ is used both as the weight of a regularising term in the loss function, and as the threshold used to binarise the difference image in order to iteratively flag/unflag changed pixels.
Given the two code images of $20$ channels with values between $0$ and $1$, the norms of the pixel-wise difference bewteen them can go from $0$ to $\sqrt{20}$, which is approximately $4.5$.
However, $\lambda$ was set equal to $0.2$, which seems reasonable as a weight but clearly too low as a threshold.
Since it is not clear whether the norms should be normalised (or the threshold multiplied) by the maximum norm $\sqrt{20}$, we opted for using the Otsu's method to find the optimal threshold for every iteration.
\fi
%
However, to make the SCCN work we had to replace a fixed parameter described in the paper with the output of Otsu's method to find an optimal threshold for the difference image in the iterative refinement of the change map.
We also had to interpret the description in~\cite{liu2016deep}: To avoid trivial solutions, we implemented their pretraining phase with decoders having one coupling layer (convolutional layer with filters of $1\times1$) and $250$ epochs.
This was empirically found to be the minimum amount of epochs needed to consistently obtain a meaningful representation of the data in the code space to be used as starting point for the training procedure.
Also, in~\cite{liu2016deep} Liu \textit{et al.}selected a rigorous stopping criterion for the latter, but it was hardly reached during our experiments, so a maximum number of epochs was set to $500$.

\subsubsection{Comparisons with other methods}

In order to better frame our architectures within the state-of-the-art of heterogeneous CD, we also present a comparison on the widely used benchmark dataset of the constructions in China.
There are several versions of this dataset in terms of image sizes and ground truth, so we focused on the methods from the literature which used the same version, to ensure that all considered results are fully comparable. 

Beside SCCN and CAN, we report for this dataset the results obtained by several methods. The mixed-norm-based (MNB) method by Touati \textit{et al.}~\cite{touati2019reliable}, the coupling translation network (CPTN) by Gong \textit{et al.}~\cite{gong2019coupling}, and the coupled dictionary learning (ICDL) method by Gong \textit{et al.}~\cite{gong2016coupled} are unsupervised.
Instead, the post-classification comparison (PCC)~\cite{chuvieco2016fundamentals}, the conditional copulas (CC) method by Mercier \textit{et al.}~\cite{mercier2008conditional}, and the anomaly feature learning (AFL) method by Touati \textit{et al.}~\cite{touati2020anomaly} are supervised approaches. For the experimental setup and implementation details of these methods applied to this specific dataset, we refer to their original papers.

Although these methods are evaluated on the same dataset, the supervised ones make use of training samples (e.g., on the "no-change" class or on the thematic classes in the scene). In terms of change detection performance, this is a clear advantage over unsupervised method, however, it comes with the cost of manual annotation based on experts' knowledge or data collection on location.
Therefore, the results must be interpreted fairly, since unsupervised methods do not make use of this kind of input but on the other hand they do not require any user prompt.

Finally, we stress another distinction: SCCN, CAN, CPTN and AFL deploy deep neural networks, so they present a similar methodological framework with respect to the proposed architectures, whereas PCC, CC, MNB, and ICDL rely on more traditional machine learning and pattern recognition techniques.

\begin{figure*}[ht!]
\begin{center}
\begin{subfigure}[t]{0.22\textwidth}
\includegraphics[width=\linewidth,keepaspectratio]{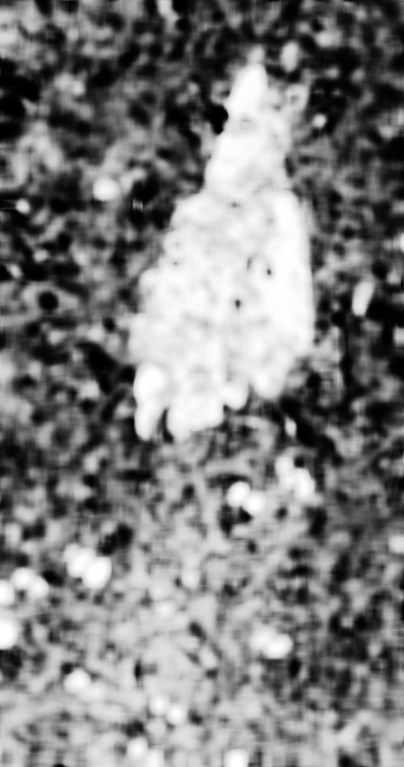}
\caption{PC, $\Delta=1$}
\label{fig:old_no_tx}
\end{subfigure}
\hspace{0.02\textwidth}%
\begin{subfigure}[t]{0.22\textwidth}
\includegraphics[width=\linewidth,keepaspectratio]{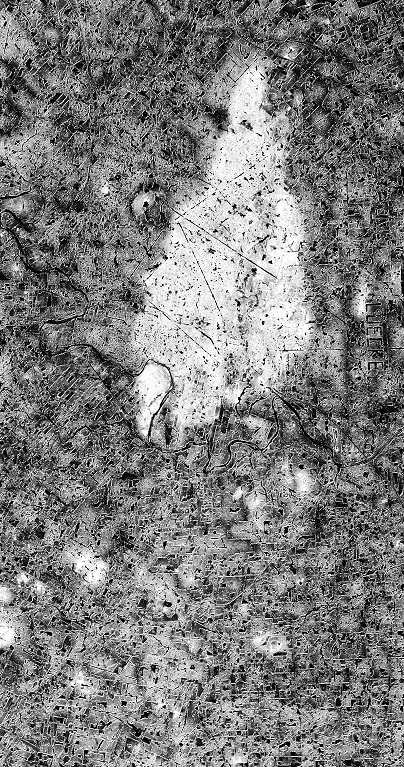}
\caption{IPC, $\Delta=1$}
\label{fig:new_no_tx}
\end{subfigure}
\hspace{0.02\textwidth}%
\begin{subfigure}[t]{0.22\textwidth}
\includegraphics[width=\linewidth,keepaspectratio]{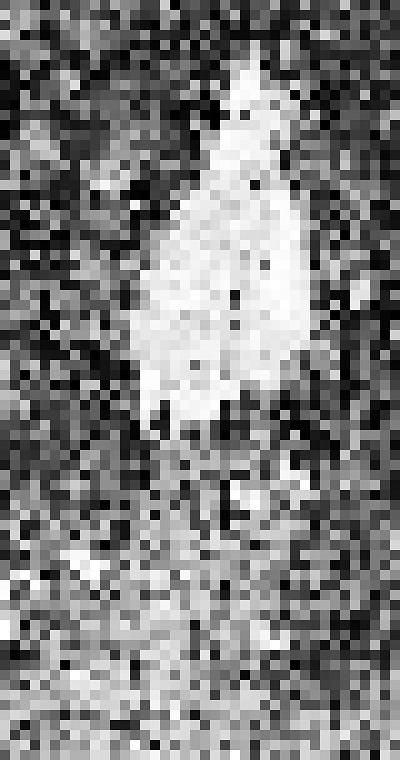}
\caption{PC, $\Delta=20$}
\label{fig:old_stride_tx}
\end{subfigure}
\hspace{0.02\textwidth}%
\begin{subfigure}[t]{0.22\textwidth}
\includegraphics[width=\linewidth,keepaspectratio]{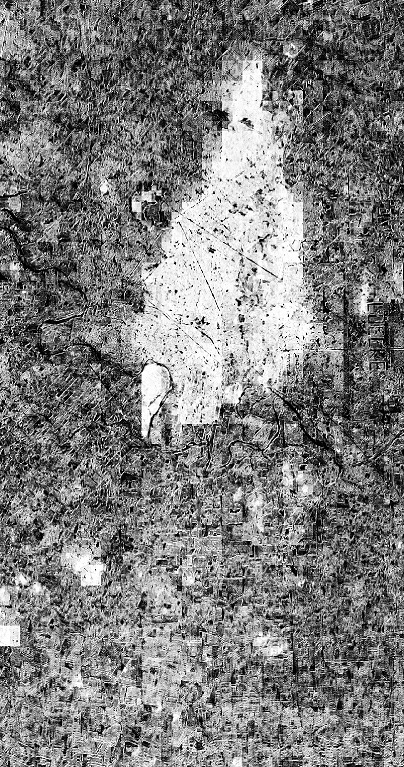}
\caption{IPC, $\Delta=20$}
\label{fig:new_stride_tx}
\end{subfigure}
\begin{subfigure}[t]{0.22\textwidth}
\includegraphics[width=\linewidth,keepaspectratio]{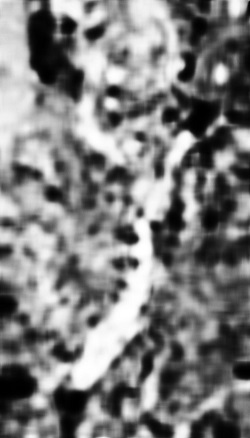}
\caption{PC, $\Delta=1$}
\label{fig:old_no_cal}
\end{subfigure}
\hspace{0.02\textwidth}%
\begin{subfigure}[t]{0.22\textwidth}
\includegraphics[width=\linewidth,keepaspectratio]{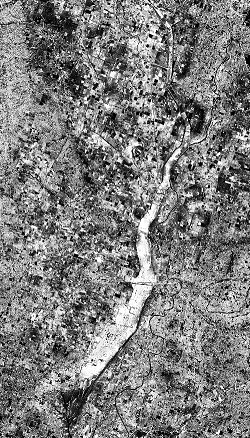}
\caption{IPC, $\Delta=1$}
\label{fig:new_no_cal}
\end{subfigure}
\hspace{0.02\textwidth}%
\begin{subfigure}[t]{0.22\textwidth}
\includegraphics[width=\linewidth,keepaspectratio]{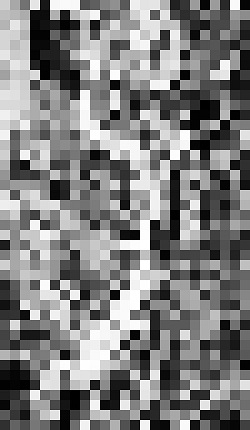}
\caption{PC, $\Delta=20$}
\label{fig:old_stride_cal}
\end{subfigure}
\hspace{0.02\textwidth}%
\begin{subfigure}[t]{0.22\textwidth}
\includegraphics[width=\linewidth,keepaspectratio]{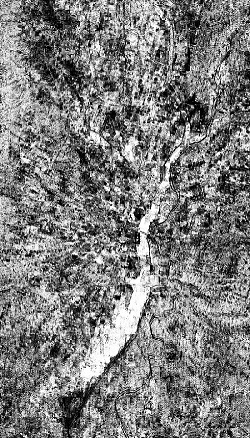}
\caption{IPC, $\Delta=20$}
\label{fig:new_stride_cal}
\end{subfigure}
\begin{subfigure}[t]{0.22\textwidth}
\includegraphics[width=\linewidth,keepaspectratio]{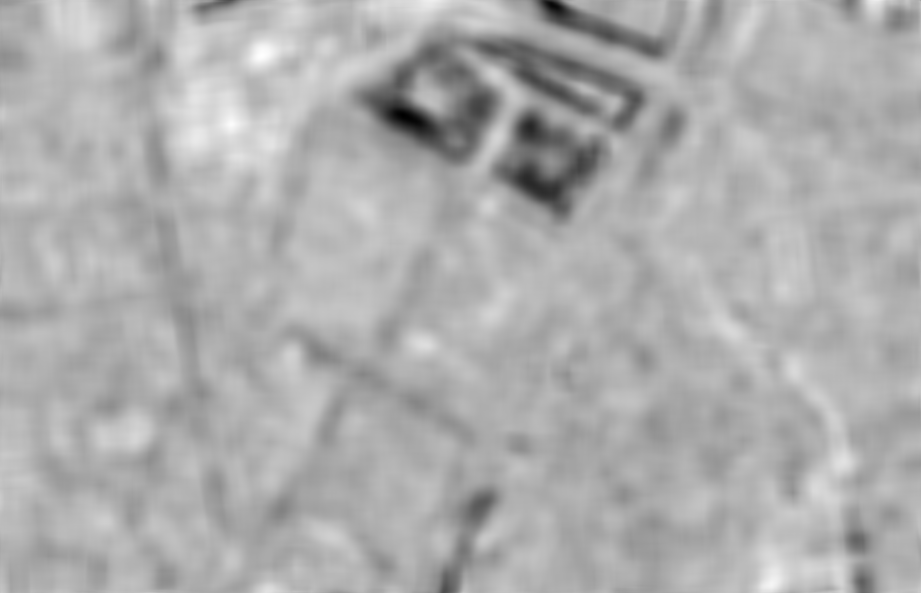}
\caption{PC, $\Delta=1$}
\label{fig:old_no_ch}
\end{subfigure}
\hspace{0.02\textwidth}%
\begin{subfigure}[t]{0.22\textwidth}
\includegraphics[width=\linewidth,keepaspectratio]{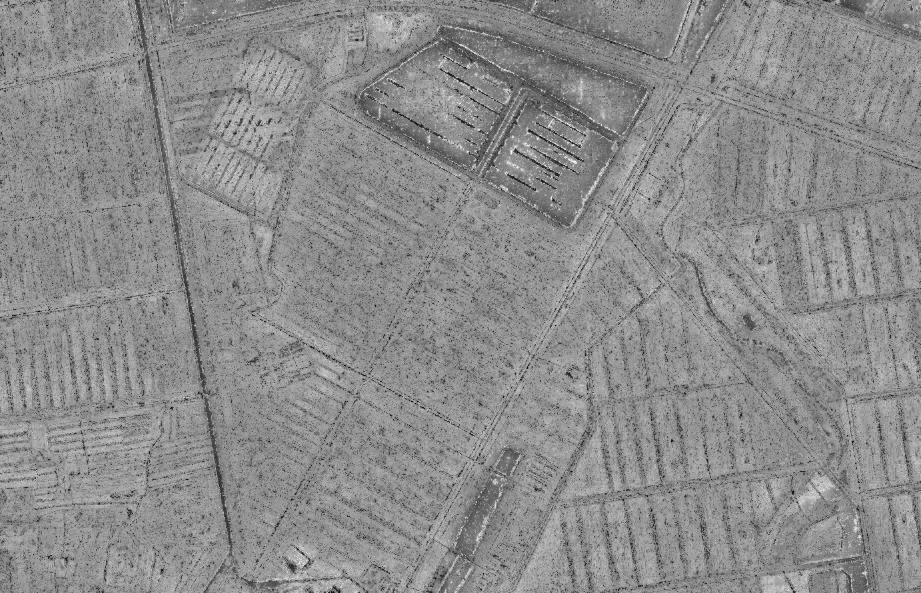}
\caption{IPC, $\Delta=1$}
\label{fig:new_no_ch}
\end{subfigure}
\hspace{0.02\textwidth}%
\begin{subfigure}[t]{0.22\textwidth}
\includegraphics[width=\linewidth,keepaspectratio]{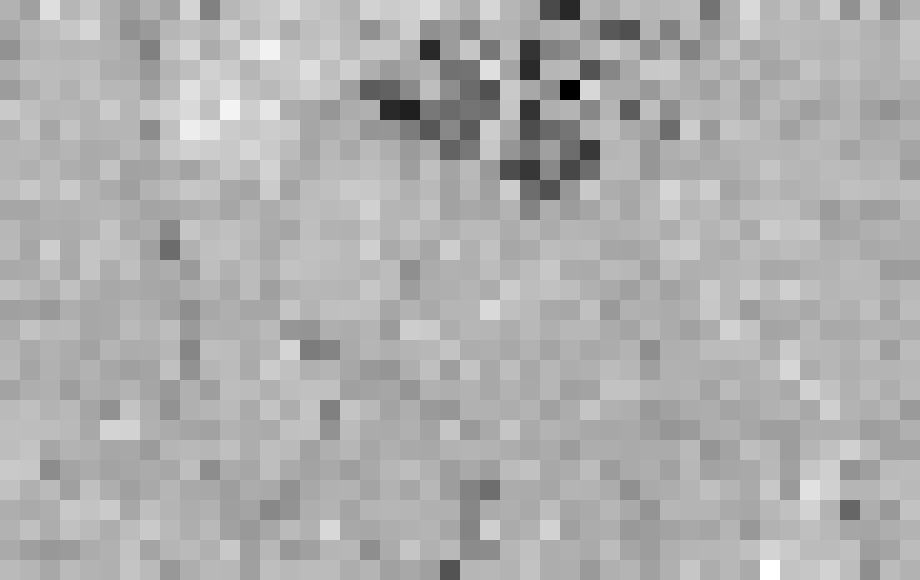}
\caption{PC, $\Delta=20$}
\label{fig:old_stride_ch}
\end{subfigure}
\hspace{0.02\textwidth}%
\begin{subfigure}[t]{0.22\textwidth}
\includegraphics[width=\linewidth,keepaspectratio]{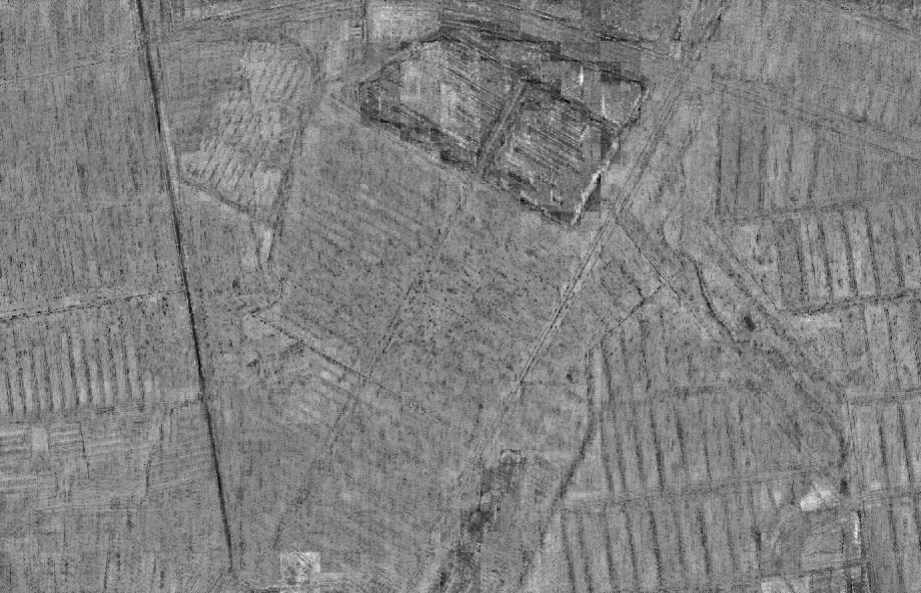}
\caption{IPC, $\Delta=20$}
\label{fig:new_stride_ch}
\end{subfigure}
\end{center}
\caption{Results on the three datasets for the PC and the IPC, for $\Delta=1$ and for $\Delta=20$.}
\label{fig:affm}
\end{figure*}

\subsection{PC vs IPC}\label{subsec:pre_comp}

The effects of the proposed modifications to the affinity matrix analysis are evaluated by a visual comparison of the results obtained by both the PC and the IPC.
Based on~\cite{luppino2019unsupervised}, a patch size of $k=20$ was selected for all the experiments.
Fig.\ \ref{fig:affm} shows the outcomes for the three datasets in the two most extreme cases, namely with strides of $\Delta=1$ and $\Delta=k$.
In the first column, one can notice how the PC provides more blurry results where the areas highlighted by their $\alpha$ values have soft edges.
In contrast, the images in the second column were obtained with the IPC and they unarguably represent a more precise result with sharp edges and smaller segments of highlighted pixels.
The third column shows the strong impact that a large $\Delta$ has on the outcomes of PC.
The PC method's assignment of one value to an entire patch leads to the tiled pattern mentioned in \Cref{subsec:affinity}.
Instead, the IPC is not as affected by the stride applied to the patch shifts, as shown in the fourth column of Fig.\ \ref{fig:affm}.

\begin{table}[b!]
\centering
\caption{Approximate $\lvert \mathcal{P} \rvert$ and computation time of the two methods applied to the three datasets for $\Delta=1$ and $\Delta=k$.}
\label{tab:Aff}

\begin{tabular}{ c c | c c c }
\toprule
\multicolumn{2}{c}{\,} & $\lvert \mathcal{P} \rvert$ & PC & IPC \\
\midrule
\multirow{2}{*}{Texas} & $\Delta = 1$ & $1.2\times10^6$ & 45 min & 76 min\\
& $\Delta = 20$ & $ 3\times10^3$ & 2:37 min& 6 min \\
\hline
\multirow{2}{*}{California} & $\Delta = 1$ & $ 4\times10^5$ & 15 min & 24 min\\
& $\Delta = 20$ & $ 1\times10^3$ & 0:37 min & 1:45 min \\
\hline
\multirow{2}{*}{China} & $\Delta = 1$ &  $5.2\times10^5$ &  35 min  &  62 min  \\
& $\Delta = 20$ & $ 1\times10^3$ & 0:40 min & 2:20 min \\
\bottomrule
\end{tabular}

\end{table}

Table \ref{tab:Aff} reports an approximate total number of patches $\lvert \mathcal{P} \rvert$ and the computation time spent by the two methods on the three datasets for the two considered cases.
As it can be seen, the major drawback of setting $\Delta=1$ is the large value of $|\mathcal{P}|$.
Recall that we propose to apply the IPC three times: with $k_{small} = 10$ and $k = 20$ to the images at the original sizes, and with $k=20$ to the images resampled at half the sizes.

\begin{table*}[ht!]
\centering
\caption{Mean and standard deviation of the evaluation metrics for the four methods applied to the Texas dataset. Best results are in bold.}
\label{tab:metrics_tx}
% \rowcolors{1}{lightgray}{lightgray}
\begin{tabular}{c | c  c  c  c c }
\toprule
\, & AUC & OA & F1 & $\kappa$ & $t$ \\
\midrule
$\boldsymbol{\alpha}$ & 0.956 & 0.922 & 0.695 & 0.652 & 42 min\\
\toprule
ACE-Net & 0.968 (0.12) & 0.951 (0.008) & 0.747 (0.047) & 0.720 (0.051) & 42 + 13 min \\
\hline
X-Net & \textbf{0.968 (0.007)} & \textbf{0.961 (0.006)} & \textbf{0.785 (0.049)} & \textbf{0.767 (0.028)} & 42 + 7 min\\
\hline
CAN~\cite{niu2018conditional}& 0.951 (0.009) & 0.925 (0.006) & 0.587 (0.048) & 0.548 (0.050) & 69 min \\
\hline
SCCN~\cite{liu2016deep} & 0.893 (0.14) & 0.880 (0.010) & 0.613 (0.309) & 0.551 (0.362) & 16 min \\
\bottomrule
\end{tabular}

\end{table*}

\begin{table*}[ht!]
\centering
\caption{Mean and standard deviation of the evaluation metrics for the four methods applied to the California dataset. Best results are in bold.}
\label{tab:metrics_cal}
% \rowcolors{1}{lightgray}{lightgray}
\begin{tabular}{c | c  c  c  c c }
\toprule
\, & AUC & OA & F1 & $\kappa$ & $t$ \\
\midrule
$\boldsymbol{\alpha}$ & 0.803 & 0.788 & 0.281 & 0.204 & 13 min\\
\toprule
ACE-Net & 0.881 (0.008) & \textbf{0.915 (0.007)} & 0.459 (0.027) & 0.415 (0.030) & 13 + 12 min \\
\hline
X-Net & 0.892 (0.006) & 0.911 (0.004) & 0.447 (0.019) & 0.402 (0.021) & 13 + 6 min \\
\hline
CAN~\cite{niu2018conditional}& 0.857 (0.008) & 0.904 (0.005) & 0.424 (0.020) & 0.365 (0.023) & 21 min \\
\hline
SCCN~\cite{liu2016deep} & \textbf{0.920 (0.002)} & 0.903 (0.007) & \textbf{0.500 (0.015)} & \textbf{0.454 (0.017)} & 15 min\\
\bottomrule
\end{tabular}

\end{table*}

Finally, for the training of the ACE-Net and the X-Net we opted for $k=20$ and $\Delta = 5$, for which the proposed approach took approximately $42$ min, $13$ min, and $19$ min for the Texas, California, and China datasets, respectively. 

\subsection{Results} \label{subsec:res}

For the first two datasets, the two proposed techniques and the previous SCCN and CAN methods were applied. Each of the four architectures was initialised randomly and trained for $100$ independent runs.
The average (standard deviation) of the evaluation metrics are reported in Table \ref{tab:metrics_tx} and Table \ref{tab:metrics_cal}, together with the average training times.
%These plots represent the behaviour of $\kappa$ \iffalse the selected metrics \fi for the compared methods: a box covers the values from the $25^\text{th}$ percentile to the $75^\text{th}$ with an orange line showing the median, while whiskers indicate the span between the $5^\text{th}$ and the $95^\text{th}$ percentile.
%Outliers beyond the whiskers are marked as circles.
As a reference, the results achieved by directly filtering and thresholding the prior $\boldsymbol{\alpha}$ are also included.
Recall that the X-Net and the ACE-Net require the computation of $\alpha$ as pre-training step: in practice, its computational time must be accounted for and added to the training time of the two architectures.
The X-Net is the simplest framework, and this explains its fast training procedure.
The ACE-Net and the SCCN have similar complexities, so they require similar times.
By contrast, the CAN paper~\cite{niu2018conditional} defines one training epochs as using all $5\times 5$ non-overlapping patches in the images, and the computational load of training grows accordingly with image size.
One may suggest to train the networks on a subsample of patches randomly picked at every epoch, but there may be a trade-off between speed and performance.
\iffalse
\begin{figure}[b!]
\begin{center}
\includegraphics[width=\columnwidth,keepaspectratio]{Kappa_SOTA_Texas.png}
\end{center}
\caption{Boxplots of \iffalse the AUC (a) and \fi the $\kappa$ coefficient for the four methods applied to the Texas dataset. The red horizontal line shows the $\kappa$ achieved with the affinity matrices comparison.}
\label{fig:plots_tx}
\end{figure}

In Fig.\ \ref{fig:plots_tx}, the results of the four methods on the Texas dataset are compared.\fi
Focusing on Table \ref{tab:metrics_tx}, The X-Net and the CAN show stable and consistent performance.
%However, only the former achieves better results than the filtered and segmented IPC, which produces $\kappa=0.65$.
The ACE-Net and the SCCN sometimes reach higher values of the evaluation metrics than the X-net, but the average is lower and the variance is high.
The performance obtained starting from $\boldsymbol{\alpha}$ suggests that technically the IPC algorithm can be used to perform heterogeneous CD autonomously, yet its application as a prior for the X-Net and the ACE-Net yields the best results.
%When compared to the IPC reference, the ACE-Net exceeds it performance in $75\%$ of the test runs, and the SCCN only in $50\%$.
A different scenario was found for the California dataset in Table \ref{tab:metrics_cal}.
%The methods perform similarly and their metrics reach consistently above the reference $\kappa=0.2$, which is the reference value produced by the IPC.
The ACE-Net outperforms the X-Net and the CAN in terms of average $\kappa$, but has more variability.
The SCCN performs best on this dataset as measured by its $\kappa$, which reaches significantly higher values than the other algorithms, and with a low variability when compared to SCCN behaviour for the Texas dataset.
However, upon closer inspection the transformations applied by this method on this dataset are not as intended and the performance is degenerate, which will be explained in \Cref{sec:limits}.
In this case, the computation of $\boldsymbol{\alpha}$ as heterogeneous CD does not seem as reliable, but it still boosts the performance of the ACE-Net and the X-Net.

\iffalse
\begin{figure}[b!]
\begin{center}
\includegraphics[width=\columnwidth,keepaspectratio]{Kappa_SOTA_California.png}
\end{center}
\caption{Boxplots of \iffalse the AUC (a) and \fi the $\kappa$ coefficient for the four methods applied to the California dataset. The red horizontal line shows the $\kappa$ achieved with the affinity matrices comparison.}
\label{fig:plots_cal}
\end{figure}
\fi

Finally, the results obtained by the state-of-the-art methods on the China dataset, along with the ones obtained by ACE-Net and X-Net, are reported in Table \ref{tab:metrics_ch}.
In addition, we note that the multidimensional scaling (MDS) method by Touati \textit{et al.}~obtained an OA of $0.967$ in~\cite{touati2018change}.
Again, the result obtained by filtering and thresholding the prior $\boldsymbol{\alpha}$ is included as reference, and the comments about the performance of $\boldsymbol{\alpha}$ for the California dataset apply also here.
The first part of Table \ref{tab:metrics_ch} consists of supervised methods, which are purposely separated from the rest: although they are evaluated on the same dataset, they require supervision and user prompt for sample selection, which makes the comparison with the other methods unfair.
We also remark that our architectures are applied with the same hyperparameters for all the datasets, whereas the hyperparameters used in~\cite{niu2018conditional,gong2019coupling} for the other methods were tuned on a case-by-case basis.
Both the ACE-Net and the X-Net outperform the other unsupervised methods, with the latter reaching higher values. These results are discussed further in \Cref{sec:limits}.

\begin{table}[hb!]
\centering
\caption{Evaluation metrics for the methods applied to the China dataset. The results indicated with $^\dag$ and $\ddag$ are reported by~\cite{gong2019coupling} and~\cite{niu2018conditional} respectively. Best results are in bold.}
\label{tab:metrics_ch}
% \rowcolors{1}{lightgray}{lightgray}
\begin{tabular}{c | c  c  c  c c }
\toprule
\, & AUC & OA & F1 & $\kappa$ \\
\midrule
$\boldsymbol{\alpha}$ & 0.848 & 0.699 & 0.248 & 0.171\\
\toprule
AFL~\cite{touati2020anomaly} & - & \textbf{0.980} &	\textbf{0.732} & \textbf{0.722}\\
\hline
$\ddag$CC~\cite{mercier2008conditional} & 0.938 & 0.951 & 0.523 & 0.444 \\
\hline
$\ddag$PCC & - & 0.821 & 0.335 & 0.257 \\
\toprule
X-Net & \textbf{0.987} & \textbf{0.984} & \textbf{0.731} & \textbf{0.696} \\
\hline
ACE-Net & 0.980 & 0.982 & 0.726 & 0.689\\
\hline
$\dag$SCCN~\cite{liu2016deep} & 0.959 & 0.976 & 0.728 & 0.679 \\
\hline
CAN~\cite{niu2018conditional}& 0.976 & 0.978 & 0.717 & 0.662 \\
\hline
CPTN~\cite{gong2019coupling} & 0.963 & 0.978 & 0.672 & 0.662\\
\hline
$\dag$ICDL~\cite{gong2016coupled} & 0.921 & 0.951 & 0.469 & 0.444 \\
\hline
MNB~\cite{touati2019reliable} & - & 0.884 & 0.370 & 0.324\\
\hline
\bottomrule
\end{tabular}

\end{table}

Fig.\ \ref{fig:tx_res}, Fig.\ \ref{fig:cal_res}, and Fig.\ \ref{fig:ch_res} show examples of output delivered by each of the four methods on the three datasets.
False colour images of the original and transformed images are composed with a subset of three channels from those available. Translated images are shown for the X-Net and the ACE-Net, followed by the resulting difference image and a confusion map (CM), which allows to visualise the accuracy of the results: TN are depicted in black, TP in white, FN in red, and FP in green. For the CAN and SCCN algorithms, the translated images are replaced with the equivalent images used by these methods to compute the difference image. For the CAN algorithm, these are a generated image $\hat{\boldsymbol{Y}}$ and a approximated image $\tilde{\boldsymbol{Y}}$ in the $\mathcal{Y}$ domain. For the SCCN algorithm, these are code images $\boldsymbol{Z}_\mathcal{X}$ and $\boldsymbol{Z}_\mathcal{Y}$ from a common latent space.

\section{Discussion}\label{sec:limits}

\subsection{Comparison among X-Net, ACE-Net, SCCN, and CAN}
Stability and consistency are the advantages of the X-Net and CAN algorithms.
They both provide good results on the selected datasets, with the former performing better.
The X-Net has other positive aspects, for example the simplicity of its architecture composed of only two CNNs of few layers each, yielding a total number of $|\vartheta|\sim 1.3\times10^5$ parameters, and fast convergence during training thanks to a limited number of terms in the loss function.

The same cannot be said about the CAN.
The framework counts three fully connected networks with $|\vartheta|\sim 3.1\times10^5$, and the use of all possible $5 \times 5$ patches as input makes its training epochs time consuming, especially for bigger datasets like the Texas one.
In addition, it shows a high tendency to miss some of the changes due to unwanted alignment of changed areas in the generated and the approximated images.
This can be noticed by the high amount of FN in Fig.\ \ref{fig:cgan_cm_tx} and Fig.\ \ref{fig:cgan_cm_cal}.

The ACE-Net has a large amount of parameters ($|\vartheta|\sim 2.8\times10^5$), and together with its complex loss function they guarantee the flexibility that allows to achieve the best overall performance on the three datasets. 
However, the complexity is also the main drawback of this architecture, because it implies a difficult and possibly slow convergence, which also results in higher variability in performance.
In conclusion, it has the potential to outperform the other methods, but a costly optimisation of its parameters might be necessary.

The SCCN requires a thorough analysis.
First of all, this network is very simple: it consists of two symmetric networks with four layers and the total amount of parameters is just $|\vartheta|\sim 6\times10^3$.
Its parameters space is thus limited when compared to its contenders.
This may explain why the method often fails to converge and provides very poor results on the first dataset (see Table \ref{tab:metrics_tx}).
The very good results displayed in Table \ref{tab:metrics_cal} instead are explained by a visual inspection of the image translations it performs on the California dataset.
After preliminary training of the two encoders, the one transforming $\boldsymbol{Y}$ is frozen, while the other is taught to align the codes of those pixels which are flagged as unchanged.
However, it can be seen in Fig.\ \ref{fig:sccn_xa_cal} that the encoder is not able to capture more than the background average colour of Fig.\ \ref{fig:sccn_ya_cal}, which can be characterized as degenerate behaviour.
Basically, the difference image in Fig.\ \ref{fig:sccn_d_cal} is highlighting the water bodies of the SAR image in Fig.\ \ref{fig2:S1A}, and this coincidentally results in high accuracy when detecting the flood.
The same situation was faced when freezing the other encoder, and this issue was encountered similarly on the China dataset, as it can be seen in Fig.\ \ref{fig:sccn_xa_ch}.
Note that high number of training epochs ($500$) in our customized implementation of the SCCN was beneficial for the Texas dataset, since it managed to converge more often to a meaningful solution, but it did not make much of a difference on the other two datasets, for which the method consistently brings the loss function to a local minimum that corresponds to a degenerate result within the first hundred of epochs, and then not being able to improve it further.

\subsection{Discussion of the results on the benchmark dataset}

The experiments on the China dataset offer a broader view in relation to the state-of-the-art.
In general, one can appreciate that there is a trade-off between performance and interpretability: CC, PCC, ICDL, and MNB formalize well-defined intuitions -- with different degrees of complexity --, and their results can be easily interpreted, but they do not achieve the same performance as the other methods.
PCC is an intuitive and very simple approach, but is quite ineffective when the two separate classifications of the single images are not very accurate because it accumulates their errors.
The concept of ICDL of using sparse dictionaries to map data into a common space is also intuitive, but its performance is less accurate in the application to this dataset as compared to the other considered methods. The same can be said for CC and MNB, whose approaches can be broadly interpreted as examples of feature engineering aimed at heterogeneous CD.

Another notable detail is the performance gap between the aforementioned CC, PCC, ICDL, and MNB, which deploy more traditional machine learning algorithms, and the deep learning methods, namely AFL, X-Net, ACE-Net, SCCN, CAN, and CPTN.
Focusing on the $\kappa$ coefficient and on the considered benchmark dataset, the algorithms in the former group do not obtain higher values than $0.44$, whereas the algorithms in the latter reach $0.66$ and above.

More in particular, CPTN and ACE-Net are similar architectures with two AEs sharing their code spaces and adversarial losses. However, CPTN places two discriminators at the output of the AEs to set apart real reconstructed images from fake transformed images, whereas ACE-Net has only one discriminator acting in the code space.
The results in \Cref{tab:metrics_ch} suggest higher effectiveness of the proposed ACE-Net configuration in the application to the China dataset.

Finally, we recall that AFL also deploys AEs, but it uses a training set made of patches selected from unchanged areas to enforce the alignment of the code spaces.
Still, the results obtained by the proposed unsupervised methodologies are in line with the ones of this supervised approach. This further suggests the effectiveness of the developed affinity prior in capturing information on unchanged areas in an unsupervised manner and of the proposed X-Net and ACE-Net architectures in taking benefit from this information.

\subsection{Ablation study}

In order to compare the contribution of each component of our networks, an ablation study was carried out on the \textit{California} dataset. Alongside, we evaluated the impact of the proposed prior. Moreover, we investigated whether the overall performance can benefit from adding adversarial learning on the image contents, i.e.\ by adding two more discriminators, one for each input space, where the translated data are compared to the original data from the same domain.

\begin{table*}[t!]
\centering
\caption{Ablation study on the California dataset: mean values (standard deviations) of metrics obtained before thresholding (AUC) and after thresholding (OA, F1, $\kappa$) with the two methodologies applied with different configurations. Best results are in bold.}
\label{tab:abl_tran}
% \rowcolors{1}{lightgray}{lightgray}
\begin{tabular}{ c c | c c c c}
\toprule
\multicolumn{2}{c}{\,} & AUC & OA & F1 & $\kappa$ \\
\midrule
\multirow{5}{*}{X-Net} & Discr Output & $0.864 (0.017)$ & $0.891 (0.013)$ & $0.412 (0.035)$ & $0.359 (0.040)$ \\
& No Alpha & $0.876 (0.004)$ & $0.908 (0.003)$ & $0.439 (0.012)$ & $0.392 (0.013)$ \\
& No Cycle & $0.883 (0.008)$ & $0.910 (0.005)$ & $0.433 (0.025)$ & $0.387 (0.027)$ \\
& No Milestones & $0.873 (0.006)$ & $0.897 (0.005)$ & $0.423 (0.015)$ & $0.372 (0.017)$ \\
& Proposed & $\textbf{0.892 (0.006)}$ & $\textbf{0.911 (0.004)}$ & $\textbf{0.447 (0.019)}$ & $\textbf{0.402 (0.021)}$ \\
\hline
\multirow{7}{*}{ACE-Net} & Discr Output & $0.870 (0.041)$ & $0.884 (0.079)$ & $0.419 (0.059)$ & $0.367 (0.070)$ \\
& No Alpha & $0.865 (0.030)$ & $0.907 (0.018)$ & $0.434 (0.041)$ & $0.387 (0.047)$ \\
& No Cycle & $0.881 (0.11)$ & $0.908 (0.006)$ & $0.429 (0.028)$ & $0.382 (0.032)$ \\
& No Milestones & $0.866 (0.030)$ & $0.892 (0.009)$ & $0.402 (0.041)$ & $0.350 (0.045)$ \\
& Proposed  & $\textbf{0.881 (0.008)}$ & $\textbf{0.915 (0.007)}$ & $\textbf{0.459 (0.027)}$ & $\textbf{0.415 (0.030)}$ \\
& No Discr & $0.872 (0.008)$ & $0.912 (0.005)$ & $0.455 (0.025)$ & $0.411 (0.027)$ \\
& No Recon & $0.875 (0.016)$ & $0.912 (0.006)$ & $0.447 (0.029)$ & $0.403 (0.032)$ \\
\bottomrule
\end{tabular}

\end{table*}

\iffalse
\begin{figure}[h!]
    \begin{center}
        \includegraphics[width=\linewidth,keepaspectratio]{Kappa_OmittedTran_California.png}
    \caption{Ablation study on the California dataset. The two methodologies are applied with different configurations. Best viewed in colours.}
    \label{fig:abl_tran}
    \end{center}
\end{figure}
\fi

%The results in Fig.\ \ref{fig:abl_tran} were obtained with the X-Net and the ACE-Net applied with different configurations:
The results in Table \ref{tab:abl_tran} were obtained with the X-Net and the ACE-Net applied with different configurations:
\begin{itemize}
    \item \textbf{Discr Output:} with two added discriminators
    \item \textbf{No Alpha:} with a randomly initialised prior.
    \item \textbf{No Cycle:} without cycle-consistency.
    \item \textbf{No Milestone:} without the two milestone updates.
    \item \textbf{Proposed:} as proposed.
    \item \textbf{No Discr:} without the discriminator for the code contents (ACE-Net only).
    \item \textbf{No Recon:} without the reconstruction loss (ACE-Net only).
\end{itemize}
As it can be noticed, discarding any element from the total loss function of the two methods is not beneficial.
Similarly, the gain of using the proposed prior rather than a random one is considerable.
Also, the update of the prior at the two milestones is shown to improve the performance.
Instead, the same cannot be said about adding the two discriminators for adversarial learning on the image contents.

\begin{figure*}[ht!]
\begin{center}
\begin{subfigure}[t]{0.175\textwidth}
\includegraphics[width=\linewidth,keepaspectratio,trim={0 2.5cm 0 0},clip]{Figures/Exp/Texas/x.jpg}
\caption{$\text{Input image: }\boldsymbol{X}$}
\label{fig:x_tx}
\end{subfigure}
\hfill
\begin{subfigure}[t]{0.175\textwidth}
\includegraphics[width=\linewidth,keepaspectratio,trim={0 2.5cm 0 0},clip]{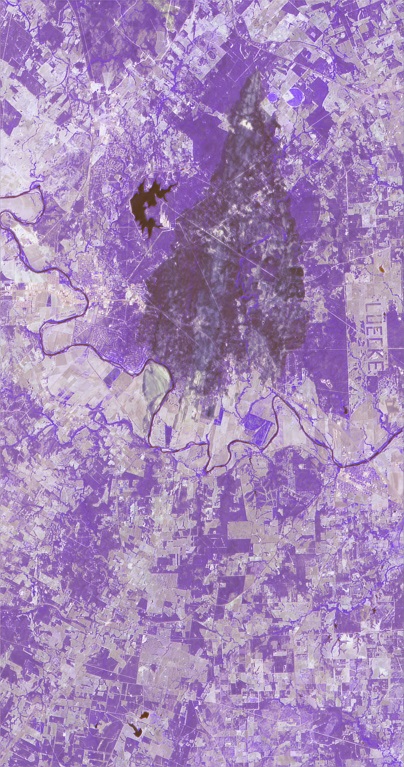}
\caption{$\text{ACE-Net transl.: }\boldsymbol{\hat{X}}$}
\label{fig:ace_xh_tx}
\end{subfigure}
\hfill
\begin{subfigure}[t]{0.175\textwidth}
\includegraphics[width=\linewidth,keepaspectratio,trim={0 2.5cm 0 0},clip]{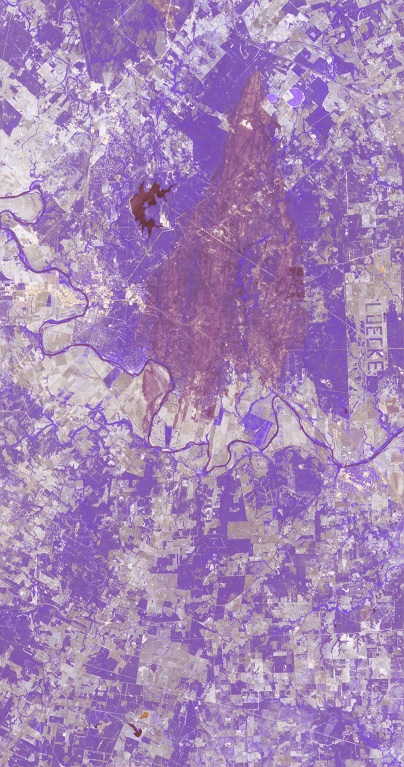}
\caption{$\text{X-Net translation: }\boldsymbol{\hat{X}}$}
\label{fig:x_xh_tx}
\end{subfigure}
\hfill
\begin{subfigure}[t]{0.175\textwidth}
\includegraphics[width=\linewidth,keepaspectratio,trim={0 2.5cm 0 0},clip]{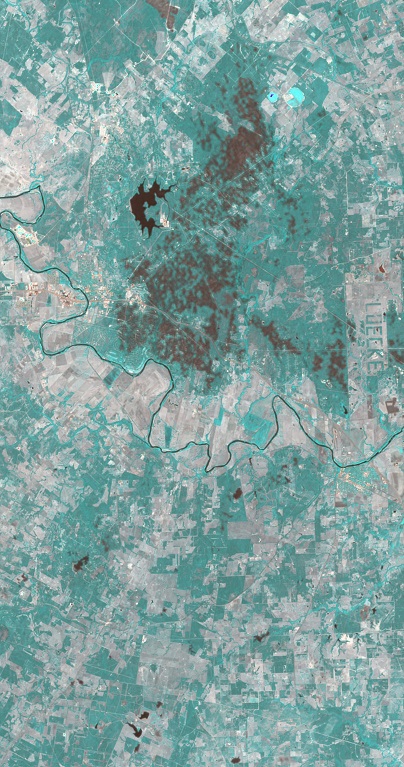}
\caption{$\text{CAN generation: }\boldsymbol{\hat{Y}}$}
\label{fig:cgan_xh_tx}
\end{subfigure}
\hfill
\begin{subfigure}[t]{0.175\textwidth}
\includegraphics[width=\linewidth,keepaspectratio,trim={0 2.5cm 0 0},clip]{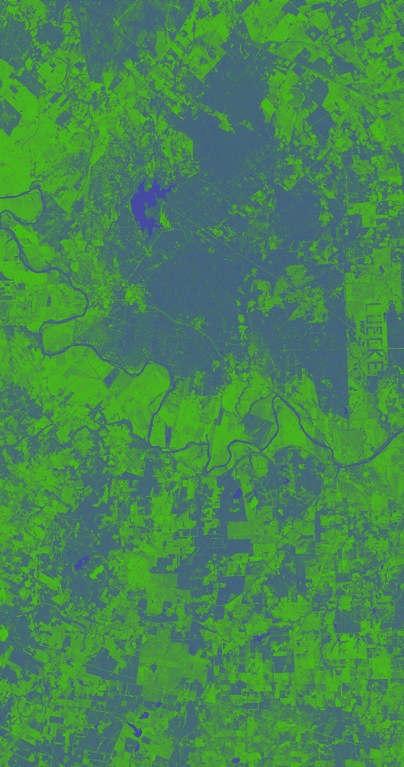}
\caption{$\text{SCCN code: }\!\boldsymbol{Z}_\mathcal{X}$}
\label{fig:sccn_xa_tx}
\end{subfigure}
\newline

\begin{subfigure}[t]{0.175\textwidth}
\includegraphics[width=\linewidth,keepaspectratio,trim={0 2.5cm 0 0},clip]{Figures/Exp/Texas/y.jpg}
\caption{$\text{Input image: }\boldsymbol{Y}$}
\label{fig:y_tx}
\end{subfigure}
\hfill
\begin{subfigure}[t]{0.175\textwidth}
\includegraphics[width=\linewidth,keepaspectratio,trim={0 2.5cm 0 0},clip]{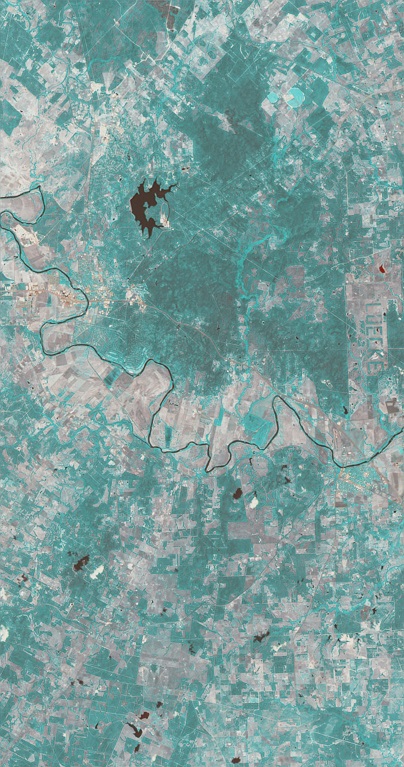}
\caption{$\text{ACE-Net transl.: }\boldsymbol{\hat{Y}}$}
\label{fig:ace_yh_tx}
\end{subfigure}
\hfill
\begin{subfigure}[t]{0.175\textwidth}
\includegraphics[width=\linewidth,keepaspectratio,trim={0 2.5cm 0 0},clip]{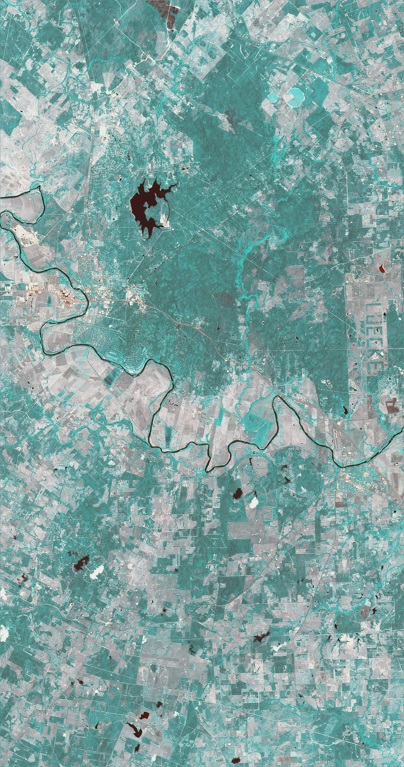}
\caption{$\text{X-Net translation: }\boldsymbol{\hat{Y}}$}
\label{fig:x_yh_tx}
\end{subfigure}
\hfill
\begin{subfigure}[t]{0.175\textwidth}
\includegraphics[width=\linewidth,keepaspectratio,trim={0 2.5cm 0 0},clip]{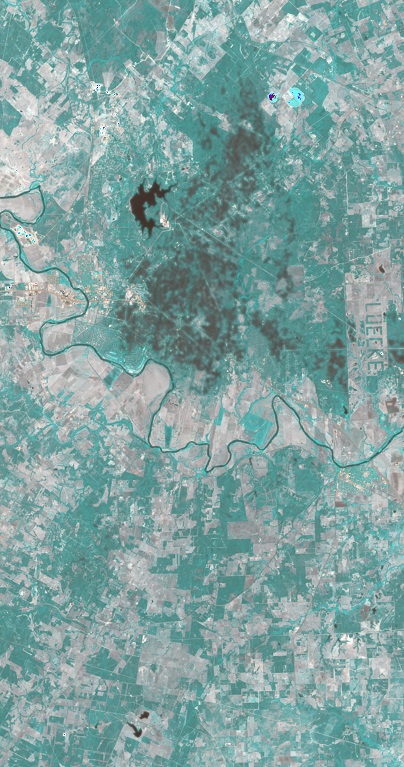}
\caption{$\text{CAN approximation: }\boldsymbol{\tilde{Y}}$}
\label{fig:cgan_yh_tx}
\end{subfigure}
\hfill
\begin{subfigure}[t]{0.175\textwidth}
\includegraphics[width=\linewidth,keepaspectratio,trim={0 2.5cm 0 0},clip]{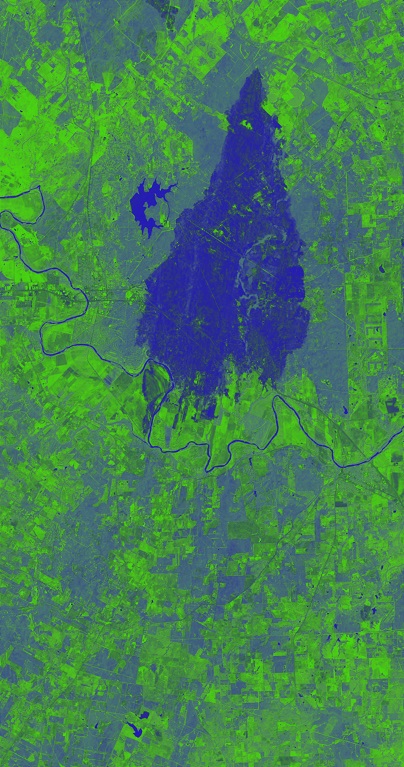}
\caption{$\!\text{SCCN code: }\boldsymbol{Z}_\mathcal{Y}$}
\label{fig:sccn_ya_tx}
\end{subfigure}
\newline

\begin{subfigure}[t]{0.175\textwidth}
\includegraphics[width=\linewidth,keepaspectratio,trim={0 2.5cm 0 0},clip]{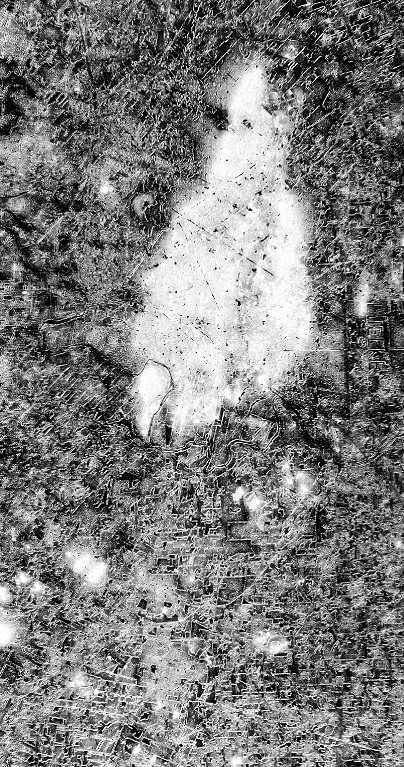}
\caption{$\text{Improved prior: }\boldsymbol{\alpha}$}
\label{fig:alpha_tx}
\end{subfigure}
\hfill
\begin{subfigure}[t]{0.175\textwidth}
\includegraphics[width=\linewidth,keepaspectratio,trim={0 2.5cm 0 0},clip]{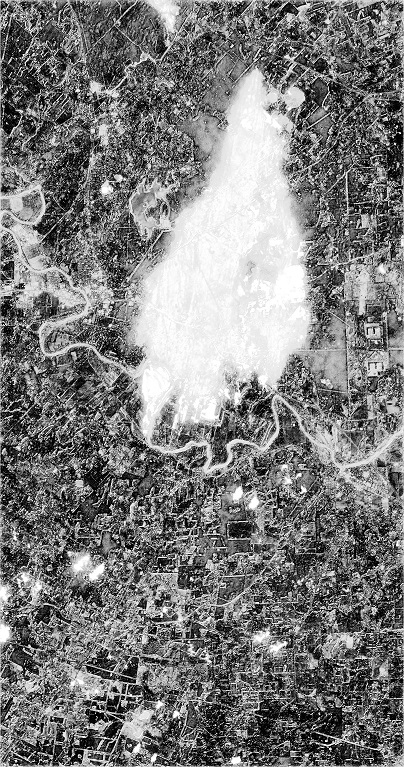}
\caption{$\text{ACE-Net diff. image}$}
%\caption{$\frac{\lVert\boldsymbol{X}-\boldsymbol{\hat{X}}\rVert+\lVert\boldsymbol{Y}-\boldsymbol{\hat{Y}}\rVert}{2}$}
\label{fig:ace_d_tx}
\end{subfigure}
\hfill
\begin{subfigure}[t]{0.175\textwidth}
\includegraphics[width=\linewidth,keepaspectratio,trim={0 2.5cm 0 0},clip]{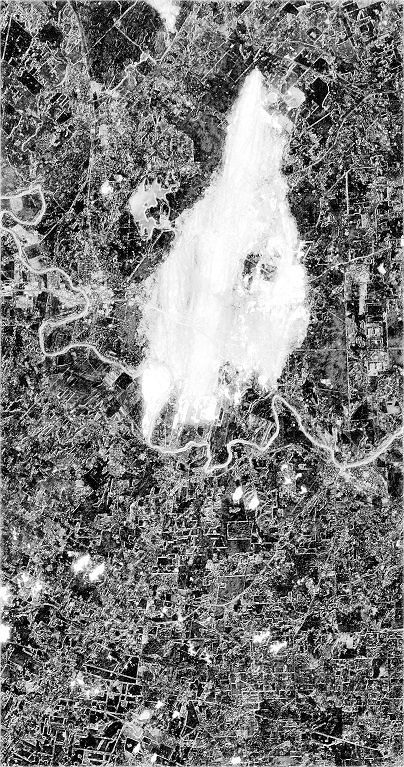}
\caption{$\text{X-Net diff. image}$}
%\caption{$\frac{\lVert\boldsymbol{X}-\boldsymbol{\hat{X}}\rVert+\lVert\boldsymbol{Y}-\boldsymbol{\hat{Y}}\rVert}{2}$}
\label{fig:x_d_tx}
\end{subfigure}
\hfill
\begin{subfigure}[t]{0.175\textwidth}
\includegraphics[width=\linewidth,keepaspectratio,trim={0 2.5cm 0 0},clip]{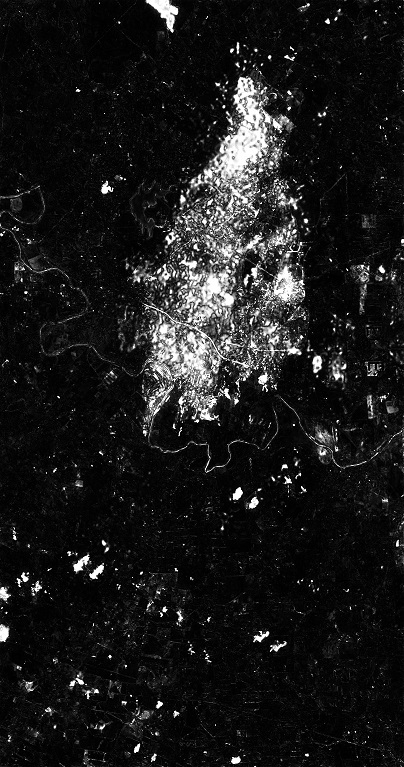}
\caption{$\text{CAN difference image}$}
%\caption{$\lVert\boldsymbol{\tilde{Y}}-\boldsymbol{\hat{Y}}\rVert$}
\label{fig:cgan_d_tx}
\end{subfigure}
\hfill
\begin{subfigure}[t]{0.175\textwidth}
\includegraphics[width=\linewidth,keepaspectratio,trim={0 2.5cm 0 0},clip]{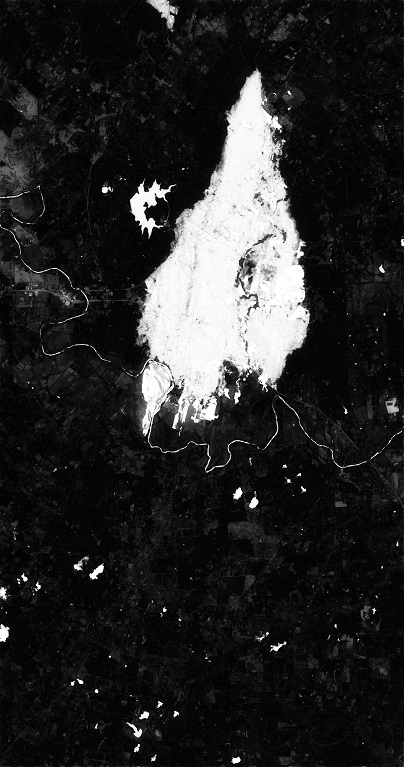}
\caption{$\text{SCCN difference image}$}
%\caption{$\lVert\boldsymbol{Z}_\mathcal{X}-\boldsymbol{Z}_\mathcal{Y}\rVert$}
\label{fig:sccn_d_tx}
\end{subfigure}
\newline

\begin{subfigure}[t]{0.175\textwidth}
\includegraphics[width=\linewidth,keepaspectratio,trim={0 2.45cm 0 0},clip]{Figures/Exp/Texas/Ground_truth.jpg}
\caption{Ground truth}
\label{fig:gt_tx}
\end{subfigure}
\hfill
\begin{subfigure}[t]{0.175\textwidth}
\includegraphics[width=\linewidth,keepaspectratio,trim={0 2.5cm 0 0},clip]{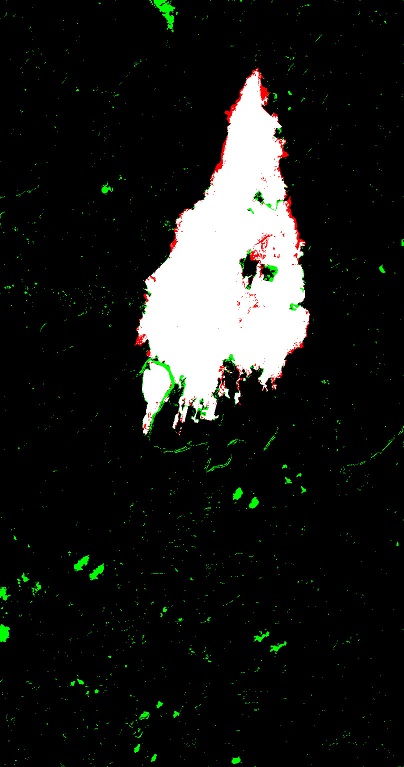}
\caption{ACE-Net CM}
\label{fig:ace_cm_tx}
\end{subfigure}
\hfill
\begin{subfigure}[t]{0.175\textwidth}
\includegraphics[width=\linewidth,keepaspectratio,trim={0 2.5cm 0 0},clip]{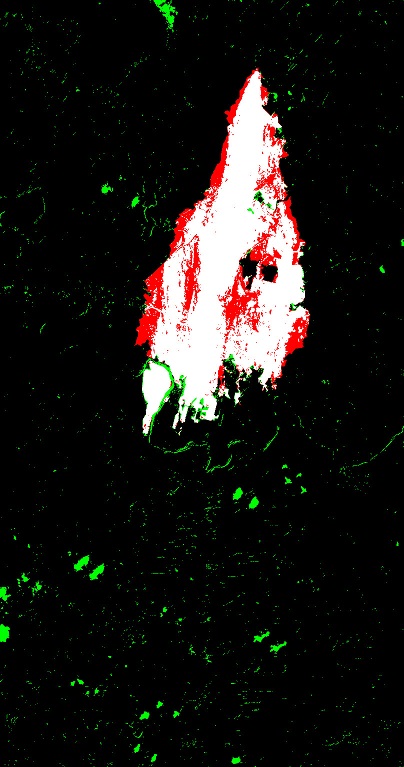}
\caption{X-Net CM}
\label{fig:x_cm_tx}
\end{subfigure}
\hfill
\begin{subfigure}[t]{0.175\textwidth}
\includegraphics[width=\linewidth,keepaspectratio,trim={0 2.5cm 0 0},clip]{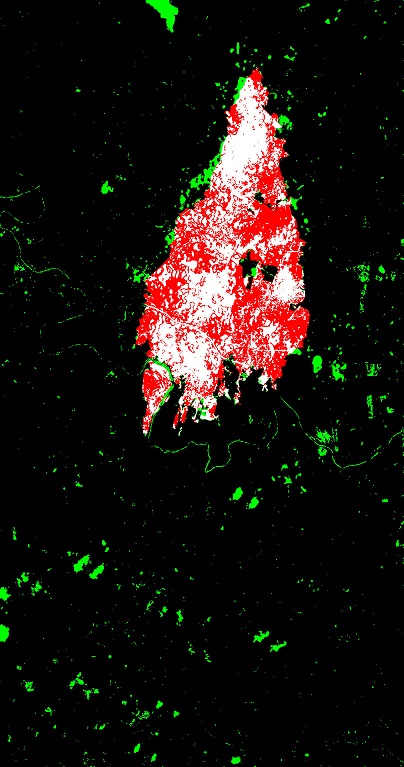}
\caption{CAN CM}
\label{fig:cgan_cm_tx}
\end{subfigure}
\hfill
\begin{subfigure}[t]{0.175\textwidth}
\includegraphics[width=\linewidth,keepaspectratio,trim={0 2.5cm 0 0},clip]{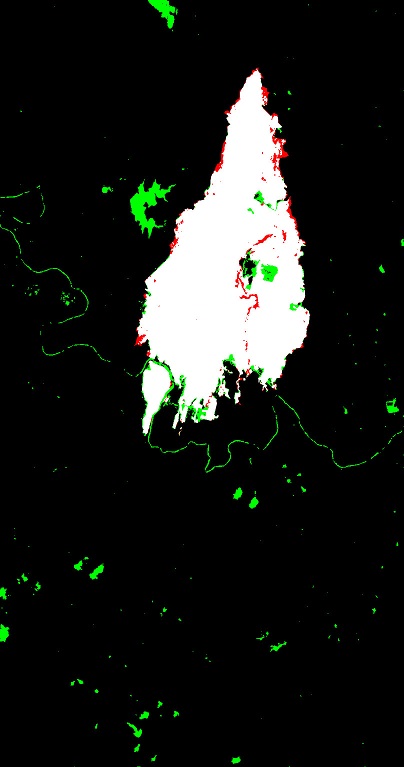}
\caption{SCCN CM}
\label{fig:sccn_cm_tx}
\end{subfigure}
\end{center}
\caption{Texas dataset. \textbf{First column}: Input images $\boldsymbol{X}$ (a) and  $\boldsymbol{Y}$ (f), IPC output $\boldsymbol{\alpha}$ (k), and ground truth (p); \textbf{Second column}: Transformed images $\boldsymbol{\hat{X}}$ (b) and $\boldsymbol{\hat{Y}}$ (g) obtained with the ACE-Net, their difference image (l) and resulting confusion map (CM) (q); \textbf{Third column}: Transformed images $\boldsymbol{\hat{X}}$ (c) and $\boldsymbol{\hat{Y}}$ (h) obtained with the X-Net, their difference image (m) and resulting CM (r); \textbf{Fourth column}: Generated SAR image $\hat{\boldsymbol{Y}}$ (d) and approximated image $\boldsymbol{\tilde{Y}}$ (i) obtained with CAN, their image difference (n), and resulting CM (s); \textbf{Fifth column}: Code images $\boldsymbol{Z}_\mathcal{X}$ (e) and $\boldsymbol{Z}_\mathcal{Y}$ (j) obtained with SCCN, their image difference (o), and resulting confusion CM (t).}
\label{fig:tx_res}
\end{figure*}

\begin{figure*}[ht!]
\begin{center}
\begin{subfigure}[t]{0.155\textwidth}
\includegraphics[width=\linewidth,keepaspectratio]{Figures/Exp/Cal/x.jpg}
\caption{Input image: $\boldsymbol{X}$}
\label{fig:x_cal}
\end{subfigure}
\hfill
\begin{subfigure}[t]{0.155\textwidth}
\includegraphics[width=\linewidth,keepaspectratio]{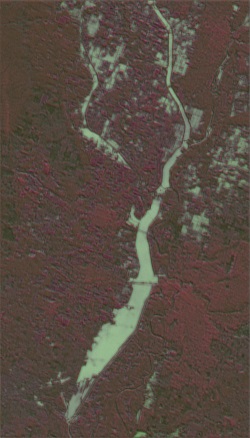}
\caption{ACE-Net transl.: $\boldsymbol{\hat{X}}$}
\label{fig:ace_xh_cal}
\end{subfigure}
\hfill
\begin{subfigure}[t]{0.155\textwidth}
\includegraphics[width=\linewidth,keepaspectratio]{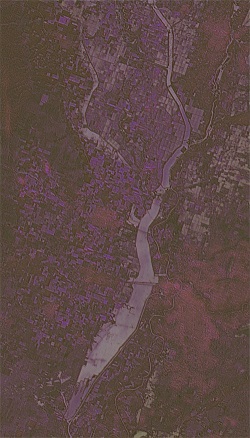}
\caption{X-Net translation: $\boldsymbol{\hat{X}}$}
\label{fig:x_xh_cal}
\end{subfigure}
\hfill
\begin{subfigure}[t]{0.155\textwidth}
\includegraphics[width=\linewidth,keepaspectratio]{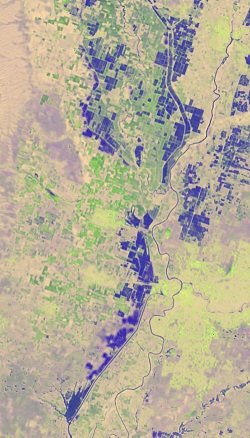}
\caption{CAN generation: $\boldsymbol{\hat{Y}}$}
\label{fig:cgan_xh_cal}
\end{subfigure}
\hfill
\begin{subfigure}[t]{0.155\textwidth}
\includegraphics[width=\linewidth,keepaspectratio]{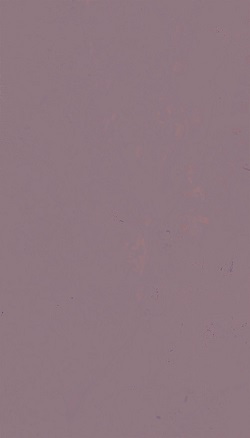}
\caption{SCCN code: $\boldsymbol{Z}_\mathcal{X}$}
\label{fig:sccn_xa_cal}
\end{subfigure}
\newline

\begin{subfigure}[t]{0.155\textwidth}
\includegraphics[width=\linewidth,keepaspectratio]{Figures/Exp/Cal/y.jpg}
\caption{Input image: $\boldsymbol{Y}$}
\label{fig:y_cal}
\end{subfigure}
\hfill
\begin{subfigure}[t]{0.155\textwidth}
\includegraphics[width=\linewidth,keepaspectratio]{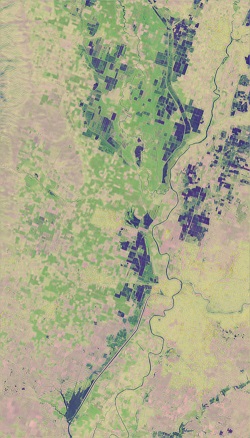}
\caption{ACE-Net trans.: $\boldsymbol{\hat{Y}}$}
\label{fig:ace_yh_cal}
\end{subfigure}
\hfill
\begin{subfigure}[t]{0.155\textwidth}
\includegraphics[width=\linewidth,keepaspectratio]{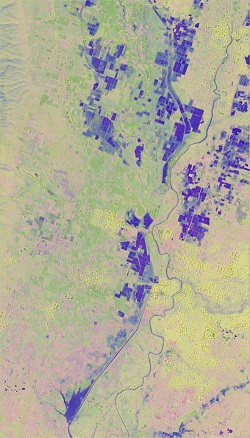}
\caption{X-Net translation: $\boldsymbol{\hat{Y}}$}
\label{fig:x_yh_cal}
\end{subfigure}
\hfill
\begin{subfigure}[t]{0.155\textwidth}
\includegraphics[width=\linewidth,keepaspectratio]{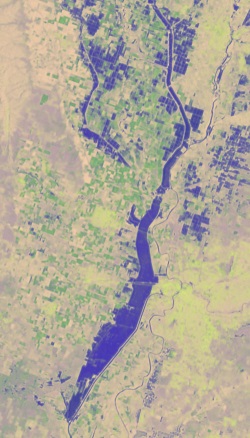}
\caption{CAN approx.: $\hat{\boldsymbol{X}}$}
\label{fig:cgan_yh_cal}
\end{subfigure}
\hfill
\begin{subfigure}[t]{0.155\textwidth}
\includegraphics[width=\linewidth,keepaspectratio]{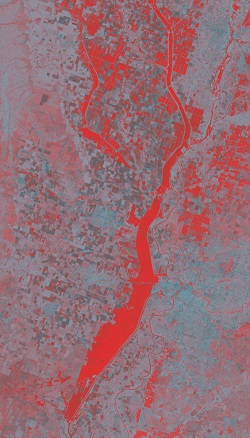}
\caption{SCCN code: $\boldsymbol{Z}_\mathcal{Y}$}
\label{fig:sccn_ya_cal}
\end{subfigure}
\newline

\begin{subfigure}[t]{0.155\textwidth}
\includegraphics[width=\linewidth,keepaspectratio]{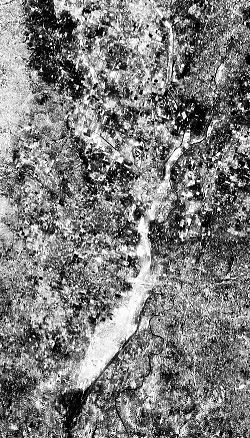}
\caption{Improved prior: $\boldsymbol{\alpha}$}
\label{fig:alpha_cal}
\end{subfigure}
\hfill
\begin{subfigure}[t]{0.155\textwidth}
\includegraphics[width=\linewidth,keepaspectratio]{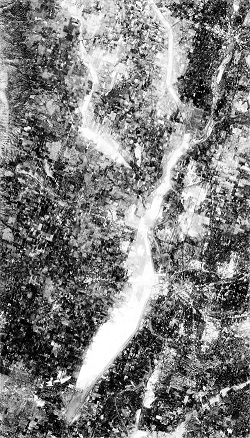}
\caption{ACE-Net diff. image}
%\caption{$\frac{\lVert\boldsymbol{X}-\boldsymbol{\hat{X}}\rVert+\lVert\boldsymbol{Y}-\boldsymbol{\hat{Y}}\rVert}{2}$}
\label{fig:ace_d_cal}
\end{subfigure}
\hfill
\begin{subfigure}[t]{0.155\textwidth}
\includegraphics[width=\linewidth,keepaspectratio]{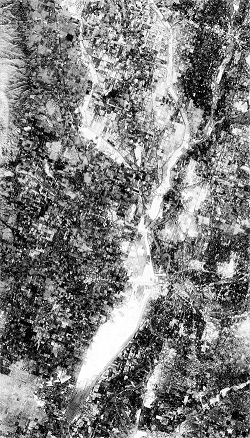}
\caption{X-Net diff. image}
%\caption{$\frac{\lVert\boldsymbol{X}-\boldsymbol{\hat{X}}\rVert+\lVert\boldsymbol{Y}-\boldsymbol{\hat{Y}}\rVert}{2}$}
\label{fig:x_d_cal}
\end{subfigure}
\hfill
\begin{subfigure}[t]{0.155\textwidth}
\includegraphics[width=\linewidth,keepaspectratio]{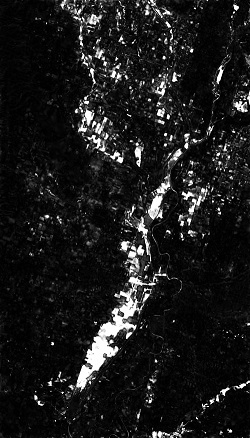}
\caption{CAN diff. image}
%\caption{$\lVert\boldsymbol{\tilde{Y}}-\boldsymbol{\hat{Y}}\rVert$}
\label{fig:cgan_d_cal}
\end{subfigure}
\hfill
\begin{subfigure}[t]{0.155\textwidth}
\includegraphics[width=\linewidth,keepaspectratio]{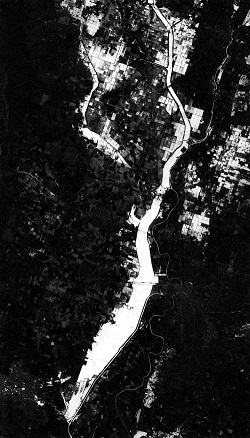}
\caption{SCCN diff. image}
%\caption{$\lVert\boldsymbol{H_x}\!(\boldsymbol{X})\!-\!\boldsymbol{H_y}\!(\boldsymbol{Y})\rVert$}
\label{fig:sccn_d_cal}
\end{subfigure}
\newline

\begin{subfigure}[t]{0.155\textwidth}
\includegraphics[width=\linewidth,keepaspectratio]{Figures/Exp/Cal/Ground_Truth.jpg}
\caption{Ground truth}
\label{fig:gt_cal}
\end{subfigure}
\hfill
\begin{subfigure}[t]{0.155\textwidth}
\includegraphics[width=\linewidth,keepaspectratio]{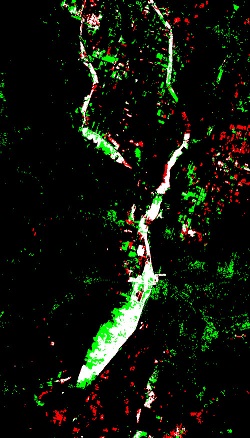}
\caption{ACE-Net CM}
\label{fig:ace_cm_cal}
\end{subfigure}
\hfill
\begin{subfigure}[t]{0.155\textwidth}
\includegraphics[width=\linewidth,keepaspectratio]{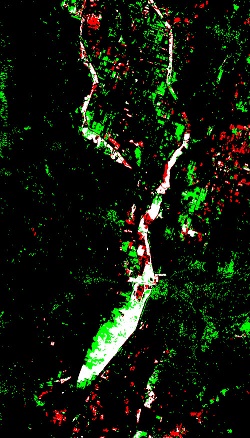}
\caption{X-Net CM}
\label{fig:x_cm_cal}
\end{subfigure}
\hfill
\begin{subfigure}[t]{0.155\textwidth}
\includegraphics[width=\linewidth,keepaspectratio]{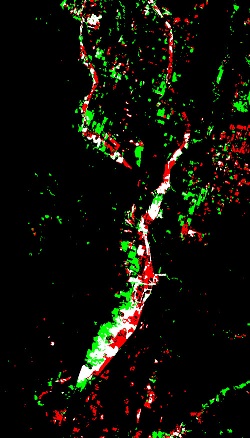}
\caption{CAN CM}
\label{fig:cgan_cm_cal}
\end{subfigure}
\hfill
\begin{subfigure}[t]{0.155\textwidth}
\includegraphics[width=\linewidth,keepaspectratio]{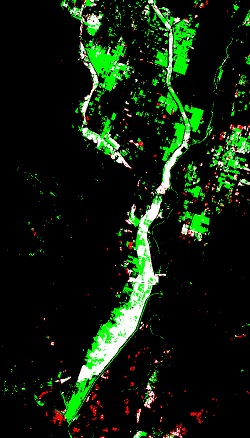}
\caption{SCCN CM}
\label{fig:sccn_cm_cal}
\end{subfigure}

\hfill

\end{center}
\caption{California dataset. \textbf{First column}: Input images $\boldsymbol{X}$ (a) and  $\boldsymbol{Y}$ (f), IPC output $\boldsymbol{\alpha}$ (k), and ground truth (p); \textbf{Second column}: Transformed images $\boldsymbol{\hat{X}}$ (b) and $\boldsymbol{\hat{Y}}$ (g) obtained with the ACE-Net, their difference image (l) and resulting confusion map (CM) (q); \textbf{Third column}: Transformed images $\boldsymbol{\hat{X}}$ (c) and $\boldsymbol{\hat{Y}}$ (h) obtained with the X-Net, their difference image (m) and resulting CM (r); \textbf{Fourth column}: Generated SAR image $\hat{\boldsymbol{Y}}$ (d) and approximated image $\boldsymbol{\tilde{Y}}$ (i) obtained with CAN, their image difference (n), and resulting CM (s); \textbf{Fifth column}: Code images $\boldsymbol{Z}_\mathcal{X}$ (e) and $\boldsymbol{Z}_\mathcal{Y}$ (j) obtained with SCCN, their image difference (o), and resulting confusion CM (t).}
\label{fig:cal_res}
\end{figure*}

\begin{figure*}[ht!]
\begin{center}
\begin{subfigure}[t]{0.175\textwidth}
\includegraphics[width=\linewidth,keepaspectratio,trim={0 2.5cm 0 0},clip]{Figures/Exp/China/t1.png}
\caption{$\text{Input image: }\boldsymbol{X}$}
\label{fig:x_ch}
\end{subfigure}
\hfill
\begin{subfigure}[t]{0.175\textwidth}
\includegraphics[width=\linewidth,keepaspectratio,trim={0 2.5cm 0 0},clip]{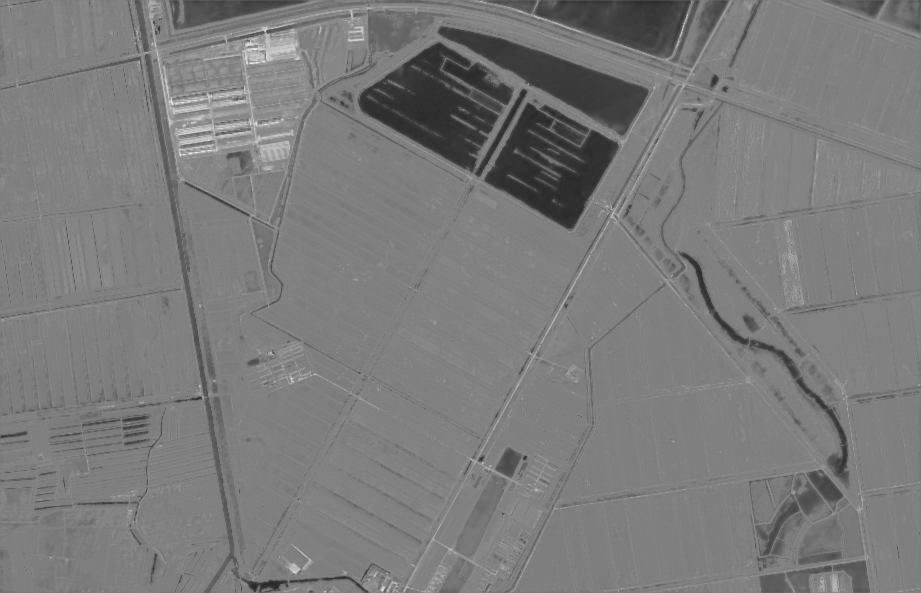}
\caption{$\text{ACE-Net transl.: }\boldsymbol{\hat{X}}$}
\label{fig:ace_xh_ch}
\end{subfigure}
\hfill
\begin{subfigure}[t]{0.175\textwidth}
\includegraphics[width=\linewidth,keepaspectratio,trim={0 2.5cm 0 0},clip]{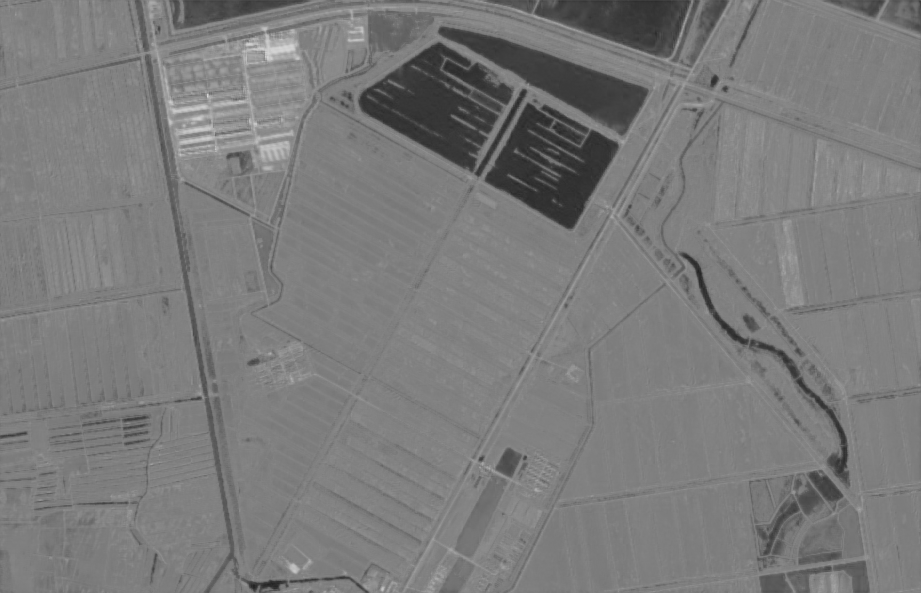}
\caption{$\text{X-Net translation: }\boldsymbol{\hat{X}}$}
\label{fig:x_xh_ch}
\end{subfigure}
\hfill
\begin{subfigure}[t]{0.175\textwidth}
\includegraphics[width=\linewidth,keepaspectratio,trim={0 2.5cm 0 0},clip]{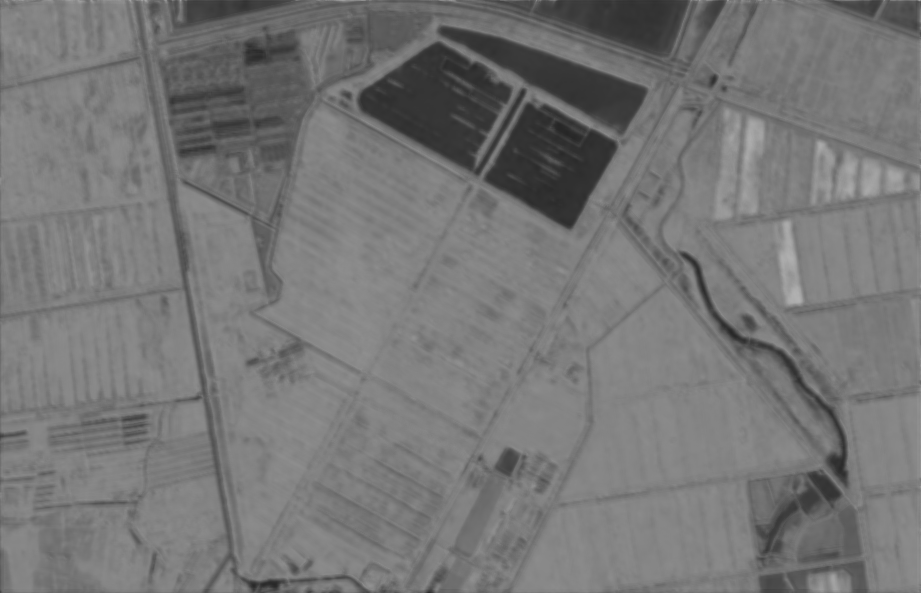}
\caption{$\text{CAN generation: }\boldsymbol{\hat{Y}}$}
\label{fig:cgan_xh_ch}
\end{subfigure}
\hfill
\begin{subfigure}[t]{0.175\textwidth}
\includegraphics[width=\linewidth,keepaspectratio,trim={0 2.5cm 0 0},clip]{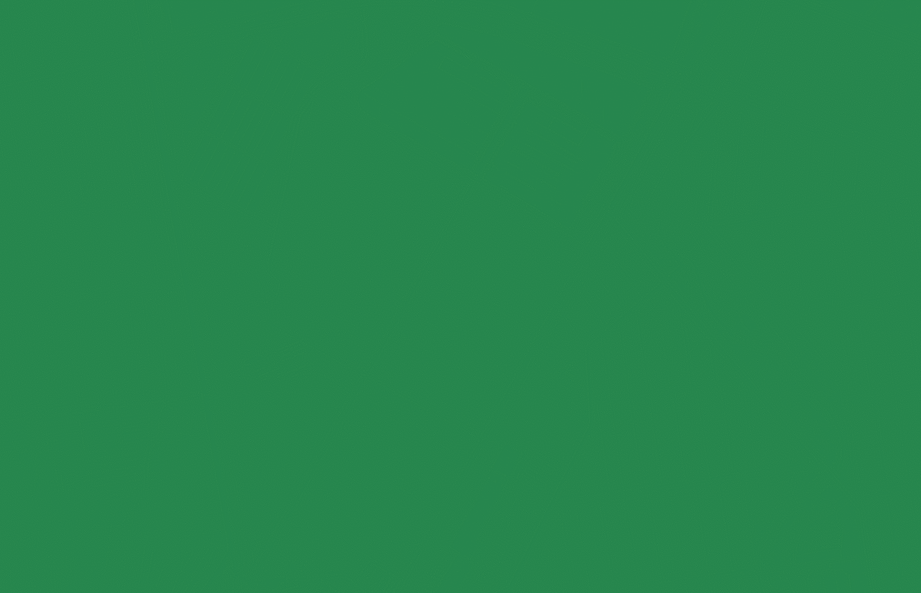}
\caption{$\text{SCCN code: }\!\boldsymbol{Z}_\mathcal{X}$}
\label{fig:sccn_xa_ch}
\end{subfigure}
\newline

\begin{subfigure}[t]{0.175\textwidth}
\includegraphics[width=\linewidth,keepaspectratio,trim={0 2.5cm 0 0},clip]{Figures/Exp/China/t2.png}
\caption{$\text{Input image: }\boldsymbol{Y}$}
\label{fig:y_ch}
\end{subfigure}
\hfill
\begin{subfigure}[t]{0.175\textwidth}
\includegraphics[width=\linewidth,keepaspectratio,trim={0 2.5cm 0 0},clip]{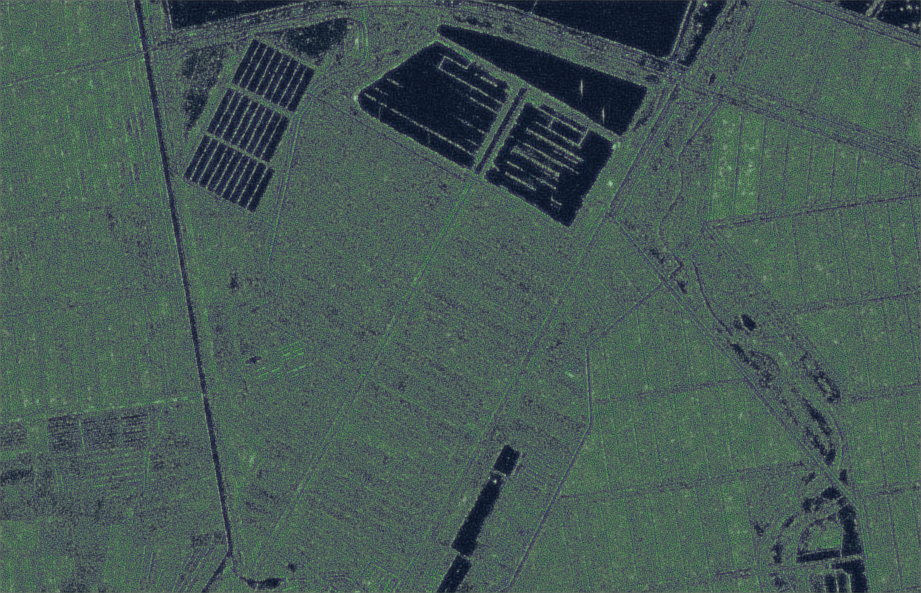}
\caption{$\text{ACE-Net transl.: }\boldsymbol{\hat{Y}}$}
\label{fig:ace_yh_ch}
\end{subfigure}
\hfill
\begin{subfigure}[t]{0.175\textwidth}
\includegraphics[width=\linewidth,keepaspectratio,trim={0 2.5cm 0 0},clip]{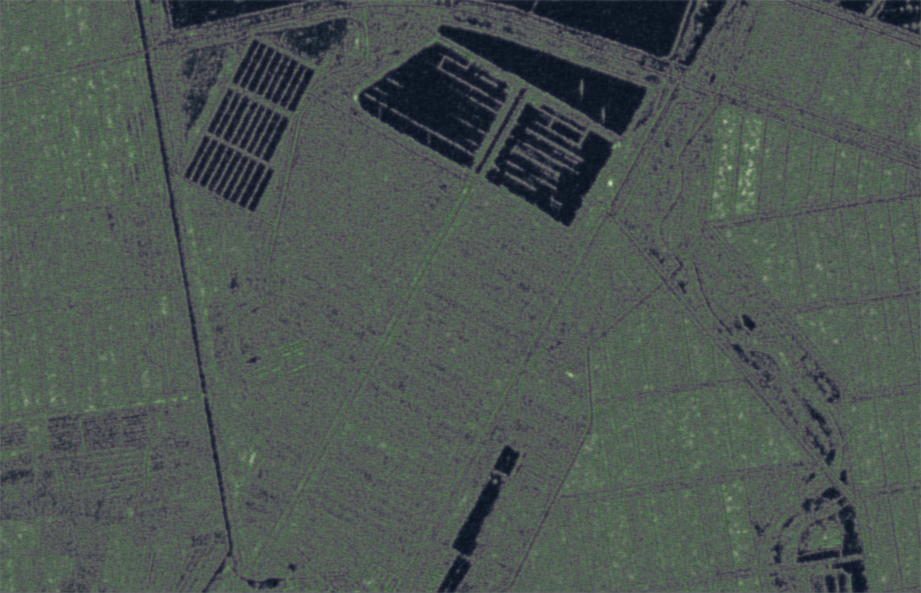}
\caption{$\text{X-Net translation: }\boldsymbol{\hat{Y}}$}
\label{fig:x_yh_ch}
\end{subfigure}
\hfill
\begin{subfigure}[t]{0.175\textwidth}
\includegraphics[width=\linewidth,keepaspectratio,trim={0 2.5cm 0 0},clip]{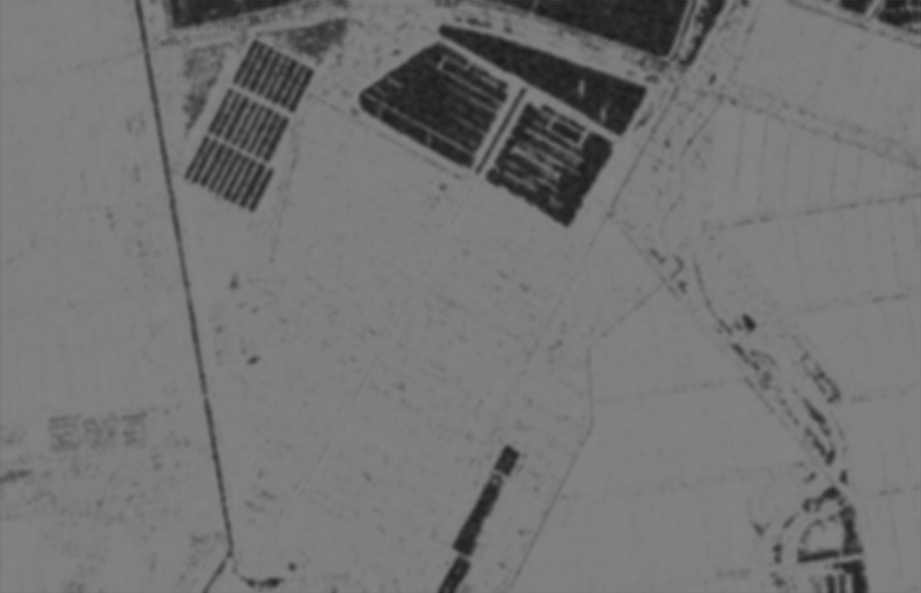}
\caption{$\text{CAN approximation: }\boldsymbol{\tilde{Y}}$}
\label{fig:cgan_yh_ch}
\end{subfigure}
\hfill
\begin{subfigure}[t]{0.175\textwidth}
\includegraphics[width=\linewidth,keepaspectratio,trim={0 2.5cm 0 0},clip]{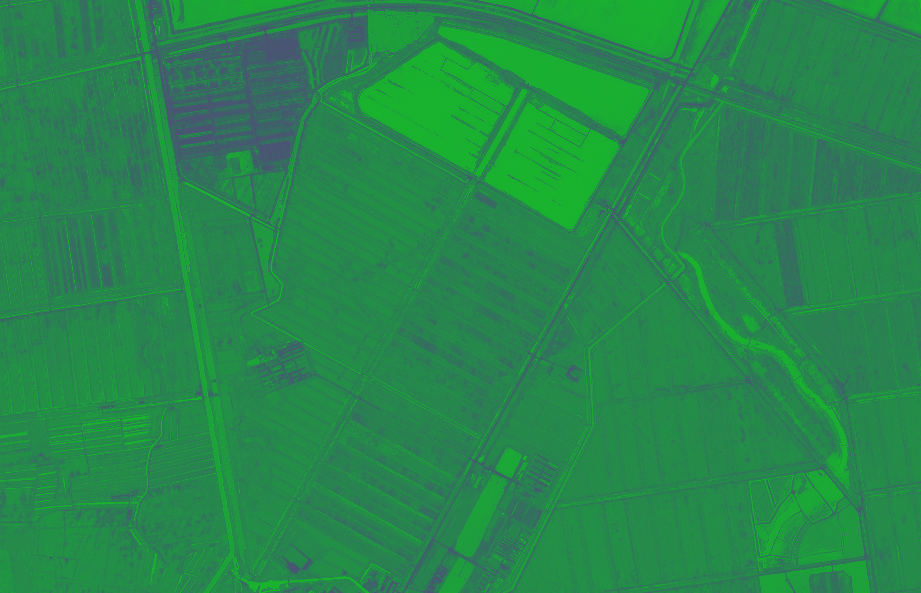}
\caption{$\!\text{SCCN code: }\boldsymbol{Z}_\mathcal{Y}$}
\label{fig:sccn_ya_ch}
\end{subfigure}
\newline

\begin{subfigure}[t]{0.175\textwidth}
\includegraphics[width=\linewidth,keepaspectratio,trim={0 2.5cm 0 0},clip]{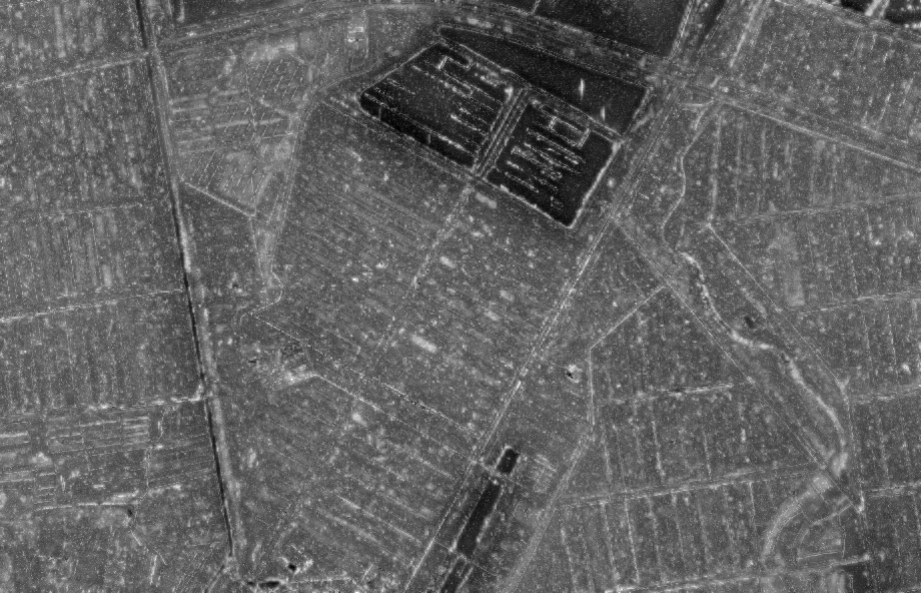}
\caption{$\text{Improved prior: }\boldsymbol{\alpha}$}
\label{fig:alpha_ch}
\end{subfigure}
\hfill
\begin{subfigure}[t]{0.175\textwidth}
\includegraphics[width=\linewidth,keepaspectratio,trim={0 2.5cm 0 0},clip]{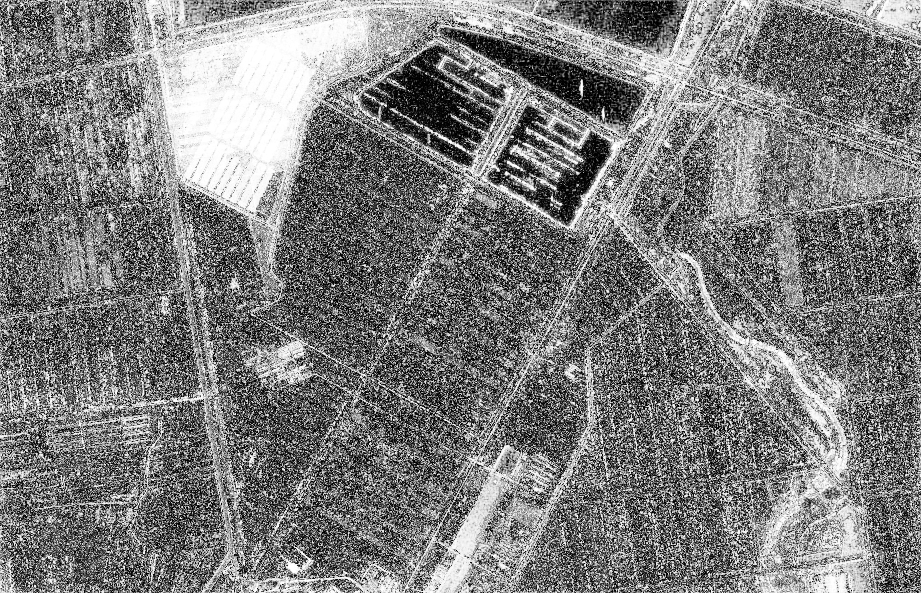}
\caption{$\text{ACE-Net diff. image}$}
%\caption{$\frac{\lVert\boldsymbol{X}-\boldsymbol{\hat{X}}\rVert+\lVert\boldsymbol{Y}-\boldsymbol{\hat{Y}}\rVert}{2}$}
\label{fig:ace_d_ch}
\end{subfigure}
\hfill
\begin{subfigure}[t]{0.175\textwidth}
\includegraphics[width=\linewidth,keepaspectratio,trim={0 2.5cm 0 0},clip]{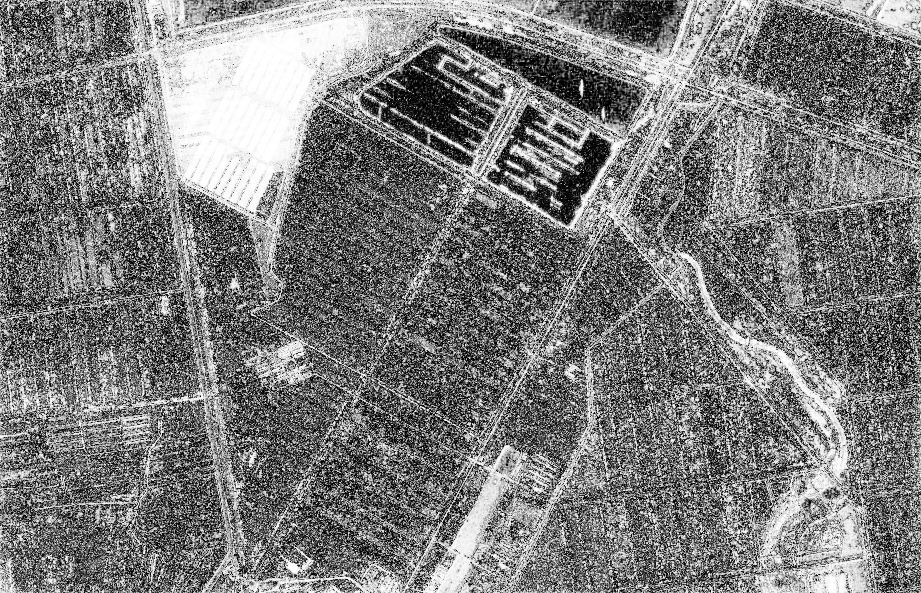}
\caption{$\text{X-Net diff. image}$}
%\caption{$\frac{\lVert\boldsymbol{X}-\boldsymbol{\hat{X}}\rVert+\lVert\boldsymbol{Y}-\boldsymbol{\hat{Y}}\rVert}{2}$}
\label{fig:x_d_ch}
\end{subfigure}
\hfill
\begin{subfigure}[t]{0.175\textwidth}
\includegraphics[width=\linewidth,keepaspectratio,trim={0 2.5cm 0 0},clip]{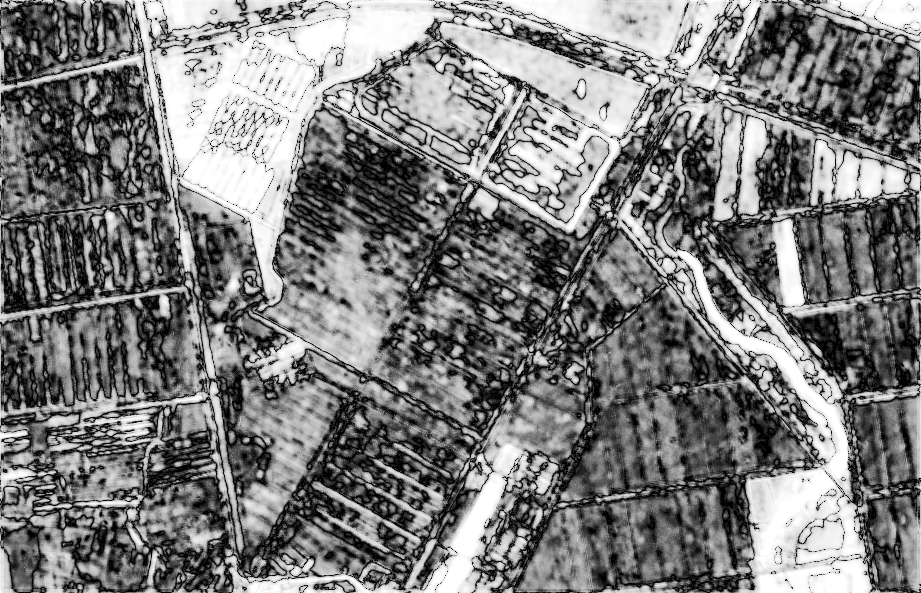}
\caption{$\text{CAN difference image}$}
%\caption{$\lVert\boldsymbol{\tilde{Y}}-\boldsymbol{\hat{Y}}\rVert$}
\label{fig:cgan_d_ch}
\end{subfigure}
\hfill
\begin{subfigure}[t]{0.175\textwidth}
\includegraphics[width=\linewidth,keepaspectratio,trim={0 2.5cm 0 0},clip]{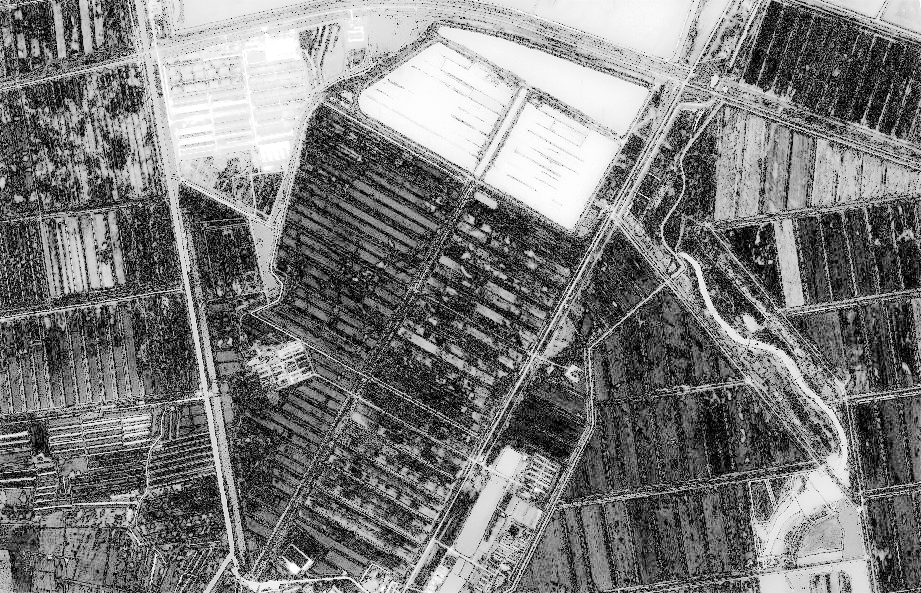}
\caption{$\text{SCCN difference image}$}
%\caption{$\lVert\boldsymbol{Z}_\mathcal{X}-\boldsymbol{Z}_\mathcal{Y}\rVert$}
\label{fig:sccn_d_ch}
\end{subfigure}
\newline

\begin{subfigure}[t]{0.175\textwidth}
\includegraphics[width=\linewidth,keepaspectratio,trim={0 2.45cm 0 0},clip]{Figures/Exp/China/ROI.png}
\caption{Ground truth}
\label{fig:gt_ch}
\end{subfigure}
\hfill
\begin{subfigure}[t]{0.175\textwidth}
\includegraphics[width=\linewidth,keepaspectratio,trim={0 2.5cm 0 0},clip]{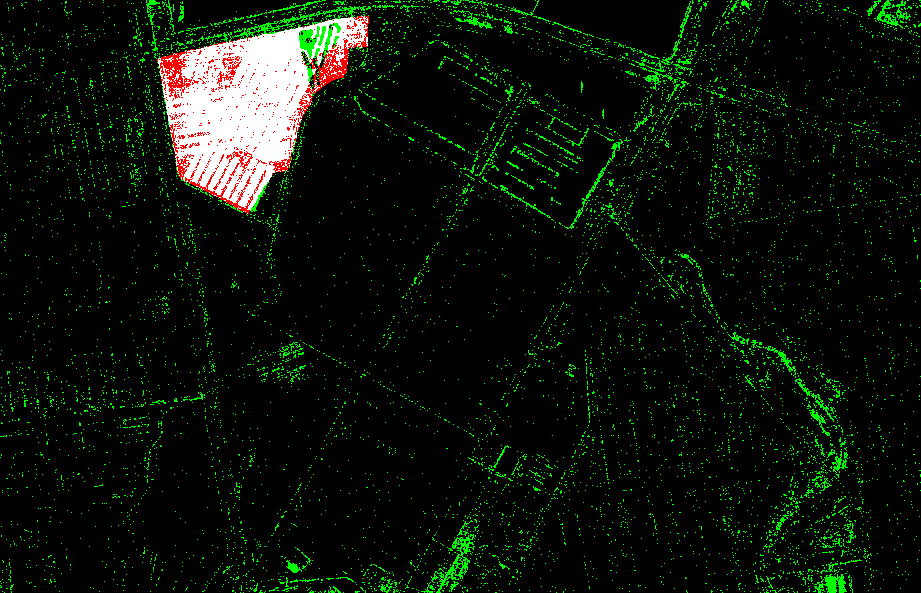}
\caption{ACE-Net CM}
\label{fig:ace_cm_ch}
\end{subfigure}
\hfill
\begin{subfigure}[t]{0.175\textwidth}
\includegraphics[width=\linewidth,keepaspectratio,trim={0 2.5cm 0 0},clip]{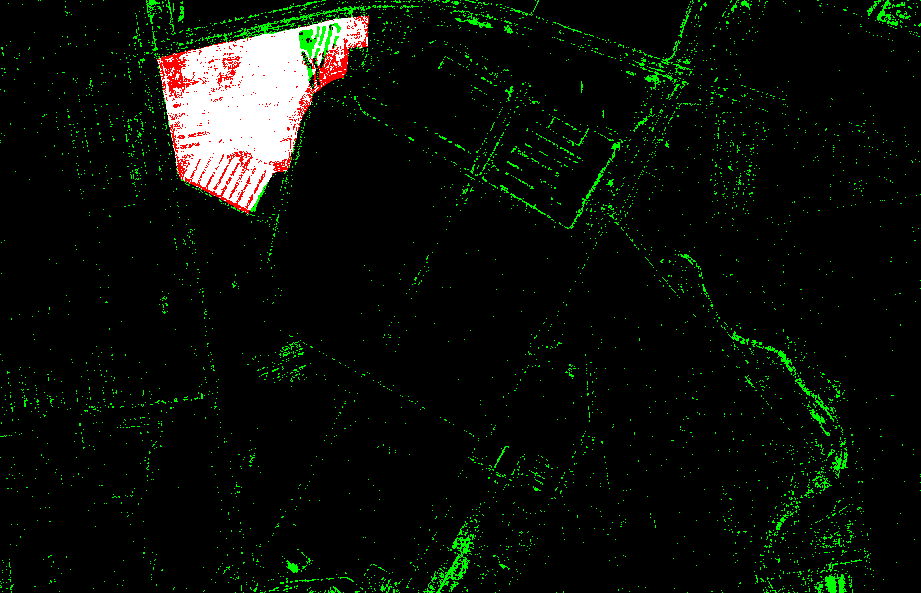}
\caption{X-Net CM}
\label{fig:x_cm_ch}
\end{subfigure}
\hfill
\begin{subfigure}[t]{0.175\textwidth}
\includegraphics[width=\linewidth,keepaspectratio,trim={0 2.5cm 0 0},clip]{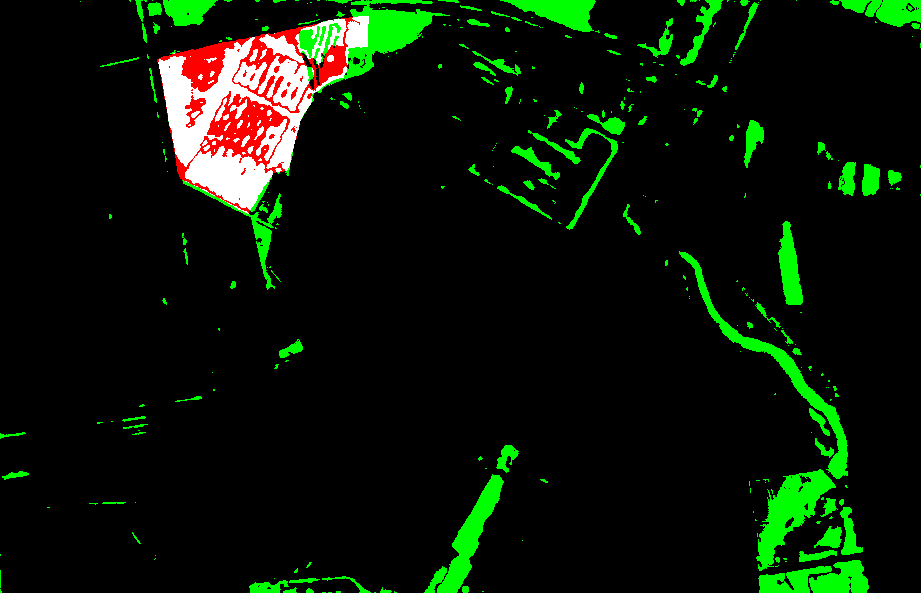}
\caption{CAN CM}
\label{fig:cgan_cm_ch}
\end{subfigure}
\hfill
\begin{subfigure}[t]{0.175\textwidth}
\includegraphics[width=\linewidth,keepaspectratio,trim={0 2.5cm 0 0},clip]{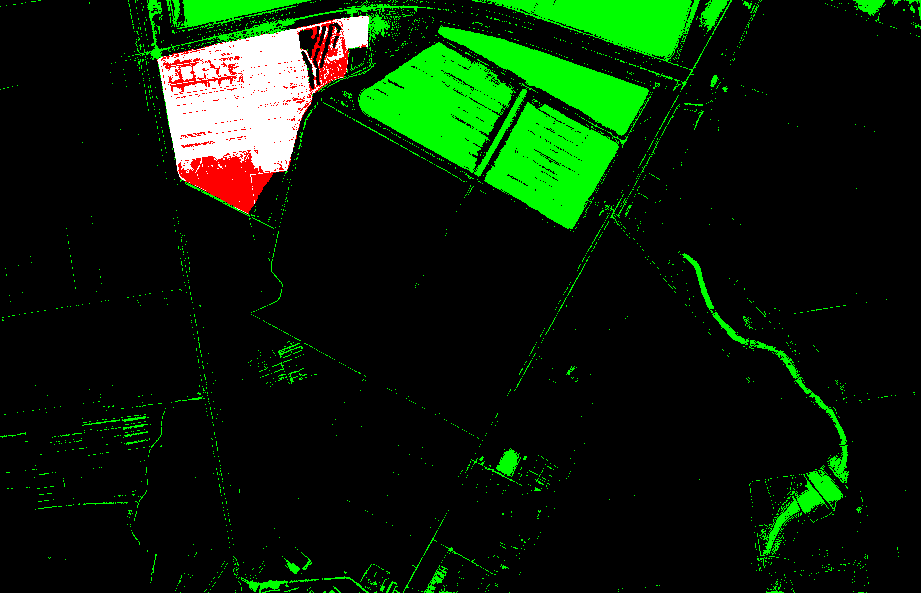}
\caption{SCCN CM}
\label{fig:sccn_cm_ch}
\end{subfigure}
\end{center}
\caption{China dataset. \textbf{First column}: Input images $\boldsymbol{X}$ (a) and  $\boldsymbol{Y}$ (f), IPC output $\boldsymbol{\alpha}$ (k), and ground truth (p); \textbf{Second column}: Transformed images $\boldsymbol{\hat{X}}$ (b) and $\boldsymbol{\hat{Y}}$ (g) obtained with the ACE-Net, their difference image (l) and resulting confusion map (CM) (q); \textbf{Third column}: Transformed images $\boldsymbol{\hat{X}}$ (c) and $\boldsymbol{\hat{Y}}$ (h) obtained with the X-Net, their difference image (m) and resulting CM (r); \textbf{Fourth column}: Generated SAR image $\hat{\boldsymbol{Y}}$ (d) and approximated image $\boldsymbol{\tilde{Y}}$ (i) obtained with CAN, their image difference (n), and resulting CM (s); \textbf{Fifth column}: Code images $\boldsymbol{Z}_\mathcal{X}$ (e) and $\boldsymbol{Z}_\mathcal{Y}$ (j) obtained with SCCN, their image difference (o), and resulting confusion CM (t).}
\label{fig:ch_res}
\end{figure*}

\section{Conclusions}\label{sec:concl}

In this work we proposed two deep convolutional neural network architectures for heterogeneous change detection: the X-Net and the ACE-Net.
In particular, we used an affinity-based change prior learnt from the input data to obtain an unsupervised algorithm.
This prior was used to drive the training process of our architectures, and the experimental results proved the effectiveness of our framework.
Both outperformed consistently state-of-the-art methods, and each has its own advantages: the X-Net proved to produce very stable and consistent performance and reliable transformations of the data; the ACE-Net showed to be able to achieve the best results, at the cost of higher complexity and a more diligent training.

\section{Acknowledgement}

The project and the first author was funded by the Research Council of Norway under research grant no.\ 251327. We gratefully acknowledge the support of NVIDIA Corporation by the donation of the GPU used for this research. The authors thank Devis Tuia for valuable discussions.

\section*{Appendix}\label{sec:spp}
In the following, we prove the equivalence between the $H_0$-conditional MSE in (\ref{eq:idealloss}) and the unconditional weighted MSE in (\ref{eq:idealloss2}). Let us focus on the first term of the loss in (\ref{eq:idealloss}), i.e.:
\begin{equation}\label{eq:app1}
    \begin{split}
        \mathcal{L}_1(\vartheta)&=\mathbb{E}_{\boldsymbol{X},\boldsymbol{Y}}\left[\left.\delta(\boldsymbol{X},\hat{\boldsymbol{X}})\:\right|H_0\right]=\\
        &=\mathbb{E}_{\boldsymbol{X},\boldsymbol{Y}}\left[\left.\delta(\boldsymbol{X},G(\boldsymbol{Y}))\:\right|H_0\right].
    \end{split}
\end{equation}

Plugging the expression of the $L_2$ squared distance into (\ref{eq:app1}) leads to:
\begin{equation}
    \mathcal{L}_1(\vartheta)=\frac{1}{n}\sum_{i=1}^n\mathbb{E}_{\boldsymbol{X},\boldsymbol{Y}}\left[\left.||G_i(\boldsymbol{Y})-\boldsymbol{x}_i||_2^2\:\right|H_0\right],
\end{equation}
where $n=h\cdot w$ and $G_i(\boldsymbol{Y})$ is the vector corresponding in $G(\boldsymbol{Y})$ to the $i$-th pixel of the patch ($i=1,2,\ldots,n$). We assume that the sample pairs $(\boldsymbol{x}_i,\boldsymbol{y}_i)$ associated with the pixels in the patch are mutually independent when conditioned to the ``no-change" hypothesis $H_0$, and that all $\boldsymbol{x}_i$ and $\boldsymbol{y}_i$ vectors are continuous random vectors ($i=1,2,\ldots,n$). The former is a rather classical conditional independence assumption, which is frequently accepted in change detection studies~\cite{bovolo2015time,mercier2008conditional,akbari2016polarimetric,solarna2018markovian}. The latter is very common, and the reformulation in the case of discrete or mixed variables is straightforward. Accordingly:
\begin{equation}\label{eq:app2}
\begin{split}
    \mathcal{L}_1(\vartheta)&=\frac{1}{n}\sum_{i=1}^n\mathbb{E}_{\boldsymbol{x}_i,\boldsymbol{Y}}\left[\left.||G_i(\boldsymbol{Y})-\boldsymbol{x}_i||_2^2\:\right|H_0\right]=\\
    &=\frac{1}{n}\sum_{i=1}^n\int||G_i(\boldsymbol{Y})-\boldsymbol{x}_i||_2^2\,p(\boldsymbol{x}_i,\boldsymbol{Y}|H_0)\,d\boldsymbol{x}_i d\boldsymbol{Y},
\end{split}
\end{equation}
where $p(\boldsymbol{x}_i,\boldsymbol{Y}|H_0)$ is the joint probability density function (PDF) of $\boldsymbol{x}_i$ and $\boldsymbol{Y}$ conditioned to $H_0$, and the Lebesgue integral is implicitly extended over the whole multidimensional space of all components of $\boldsymbol{x}_i$ and $\boldsymbol{Y}$. Thanks to the law of total probability ($i=1,2,\ldots,n$):
\begin{equation}
    p(\boldsymbol{x}_i,\boldsymbol{Y})=P(H_0)p(\boldsymbol{x}_i,\boldsymbol{Y}|H_0)+P(H_1)p(\boldsymbol{x}_i,\boldsymbol{Y}|H_1),
\end{equation}
where $p(\boldsymbol{x}_i,\boldsymbol{Y})$ is the unconditional joint PDF of $\boldsymbol{x}_i$ and $\boldsymbol{Y}$, $p(\boldsymbol{x}_i,\boldsymbol{Y}|H_1)$ is their joint PDF conditioned to $H_1$, and $P(H_0)$ and $P(H_1)$ are the prior probabilities of the two hypotheses. Straightforward algebraic manipulations allow proving that:
\begin{equation}\label{eq:app5}
    p(\boldsymbol{x}_i,\boldsymbol{Y}|H_0)=\psi(\boldsymbol{x}_i,\boldsymbol{Y})p(\boldsymbol{x}_i,\boldsymbol{Y}),
\end{equation}
where:
\begin{equation}\label{eq:app3}
    \psi(\boldsymbol{x}_i,\boldsymbol{Y})=\frac{1}{P(H_0)+P(H_1)\Lambda(\boldsymbol{x}_i,\boldsymbol{Y})}
\end{equation}
and where:
\begin{equation}
    \Lambda(\boldsymbol{x}_i,\boldsymbol{Y})=\frac{p(\boldsymbol{x}_i,\boldsymbol{Y}|H_1)}{p(\boldsymbol{x}_i,\boldsymbol{Y}|H_0)}
\end{equation}
is the likelihood ratio associated with the two hypotheses~\cite{VanTrees}. Plugging (\ref{eq:app5}) into (\ref{eq:app2}) yields:
\begin{equation}
    \begin{split}
        \mathcal{L}_1(\vartheta)&=\frac{1}{n}\sum_{i=1}^n\int||G_i(\boldsymbol{Y})-\boldsymbol{x}_i||_2^2\,\psi(\boldsymbol{x}_i,\boldsymbol{Y})p(\boldsymbol{x}_i,\boldsymbol{Y})\,d\boldsymbol{x}_i d\boldsymbol{Y}=\\
        &=\frac{1}{n}\sum_{i=1}^n\mathbb{E}_{\boldsymbol{x}_i,\boldsymbol{Y}}\left[||G_i(\boldsymbol{Y})-\boldsymbol{x}_i||_2^2\,\psi(\boldsymbol{x}_i,\boldsymbol{Y})\right]=\\
        &=\frac{1}{n}\sum_{i=1}^n\mathbb{E}_{\boldsymbol{X},\boldsymbol{Y}}\left[||G_i(\boldsymbol{Y})-\boldsymbol{x}_i||_2^2\,\psi(\boldsymbol{x}_i,\boldsymbol{Y})\right],
    \end{split}
\end{equation}
where in the last equality the conditional independence of the $(\boldsymbol{x}_i,\boldsymbol{y}_i)$ pairs ($i=1,2,\ldots,n$) has been used again. Given the notation in (\ref{eq:delta}) for the weighted $L_2$ squared loss, we can conclude that:
\begin{equation}\label{eq:app4}
    \begin{split}
        \mathcal{L}_1(\vartheta)&=\mathbb{E}_{\boldsymbol{X},\boldsymbol{Y}}\left[\frac{1}{n}\sum_{i=1}^n||G_i(\boldsymbol{Y})-\boldsymbol{x}_i||_2^2\,\psi(\boldsymbol{x}_i,\boldsymbol{Y})\right]=\\
        &=\mathbb{E}_{\boldsymbol{X},\boldsymbol{Y}}\left[\delta(\boldsymbol{X},G(\boldsymbol{Y})|\boldsymbol\Psi)\right]=\\
        &=\mathbb{E}_{\boldsymbol{X},\boldsymbol{Y}}\left[\delta(\boldsymbol{X},\hat{\boldsymbol{X}}|\boldsymbol\Psi)\right],
    \end{split}
\end{equation}
where $\boldsymbol\Psi$ is the $n$-dimensional vector collecting all terms $\psi(\boldsymbol{x}_i,\boldsymbol{Y})$ that weigh the $L_2$ loss (i.e., $\Psi_i=\psi(\boldsymbol{x}_i,\boldsymbol{Y})$ for $i=1,2,\ldots,n$). This proves the equivalence, under the aforementioned conditional independence assumption, between the $H_0$-conditional MSE term involving $\boldsymbol{X}$ and $\hat{\boldsymbol{X}}$ in (\ref{eq:idealloss}) and the corresponding term in the unconditional weighted MSE in (\ref{eq:idealloss2}). The same argument can be used to prove the equivalence between the MSE terms involving $\boldsymbol{Y}$ and $\hat{\boldsymbol{Y}}$.

Furthermore, focusing on the $i$-th pixel of the patch ($i=1,2,\ldots,n$), we note that the likelihood ratio $\Lambda(\boldsymbol{x}_i,\boldsymbol{Y})$ takes values in $[0,+\infty)$. Hence, according to (\ref{eq:app3}), $\psi(\boldsymbol{x}_i,\boldsymbol{Y})$ takes values in $(0,1/P(H_0)]$. Specifically, small values of $\psi(\boldsymbol{x}_i,\boldsymbol{Y})$ are obtained in the case of large values of $\Lambda(\boldsymbol{x}_i,\boldsymbol{Y})$ (in the limit case, $\psi(\boldsymbol{x}_i,\boldsymbol{Y})\longrightarrow0^+$ if $\Lambda(\boldsymbol{x}_i,\boldsymbol{Y})\longrightarrow+\infty$). Following the reasoning of a Bayesian likelihood ratio test, these comments suggest that, if $\psi(\boldsymbol{x}_i,\boldsymbol{Y})$ takes a small value, then the $i$-th pixel likely belongs to $H_1$~\cite{VanTrees}. Vice versa, values of $\psi(\boldsymbol{x}_i,\boldsymbol{Y})$ close to the maximum $1/P(H_0)$ are achieved if $\Lambda(\boldsymbol{x}_i,\boldsymbol{Y})$ is small (in particular, $\psi(\boldsymbol{x}_i,\boldsymbol{Y})=1/P(H_0)$ if and only if $\Lambda(\boldsymbol{x}_i,\boldsymbol{Y})=0$) -- a configuration that suggests that the $i$-th pixel likely belongs to $H_0$~\cite{VanTrees}. This confirms the interpretation of small and large values of the components of $\boldsymbol\Psi$ in relation to membership to ``change" or ``no-change," respectively. Similar comments hold with regard to the components of $\boldsymbol\Phi$.

\bibliographystyle{IEEEtran}
\bibliography{references}
\end{document}